\pdfoutput=1  
\documentclass[11pt]{article}

\usepackage{acl}

\usepackage{times}
\usepackage{latexsym}
\usepackage[T1]{fontenc}
\usepackage[utf8]{inputenc}
\usepackage{microtype}
\usepackage{inconsolata}
\usepackage{booktabs}
\usepackage{multirow}
\usepackage{amssymb}
\usepackage{stmaryrd}
\usepackage{array}
\usepackage{newunicodechar}
\newunicodechar{≥}{\ensuremath{\geq}}
\usepackage{amssymb}
\usepackage{newunicodechar}
\newunicodechar{↔}{\ensuremath{\leftrightarrow}}
\newunicodechar{≤}{\ensuremath{\leq}}
\usepackage{amsmath,amssymb}
\usepackage{newunicodechar}
\newunicodechar{∈}{\ensuremath{\in}}
\usepackage{graphicx}
\usepackage{subcaption}
\usepackage{xcolor}
\usepackage[capitalize]{cleveref}
\usepackage{enumitem}
\usepackage{fvextra}
\usepackage{xspace}
\usepackage{tabularx}
\usepackage{ragged2e}
\usepackage{adjustbox}
\newcolumntype{Y}{>{\RaggedRight\arraybackslash}X}
\DefineVerbatimEnvironment{MyVerbatim}{Verbatim}{
  breaklines=true,
  breakanywhere=true,
  fontsize=\scriptsize,
  frame=single,
  framesep=1.5mm
}

\usepackage{makecell}
\usepackage{needspace}

\newif\ifanonymous
\anonymousfalse

\newcommand{\PromptBlock}[4]{%
  \subsubsection{#1}%
  \label{#2}%
  \vspace{0.2em}%
  \ifanonymous
    \Needspace{8\baselineskip}%
    \VerbatimInput[
      breaklines=true,
      breakanywhere=true,
      fontsize=\tiny,
      baselinestretch=0.82,
      frame=single
    ]{#4}%
  \else
    \noindent\textbf{Full prompt:} \href{#3}{GitHub source}.%
  \fi
  \par\medskip
}

\DeclareFontShape{T1}{ptm}{m}{scit}{<->ssub*ptm/m/sc}{}
\DeclareFontShape{T1}{ptm}{b}{scit}{<->ssub*ptm/b/sc}{}

\crefname{section}{\S}{\S\S}
\Crefname{section}{\S}{\S\S}
\usepackage{amsthm, amssymb}
\theoremstyle{definition}

\newtheorem{definition}{Definition}
\newtheorem{theorem}{Theorem}
\newtheorem{corollary}{Corollary}
\newcommand{\name}{\textsc{Verdict}\xspace}
\newcommand{\MaxSMT}{\textsc{MaxSMT}\xspace}
\newcommand{\SMT}{\textsc{SMT}\xspace}
\newcommand{\CF}{\textsc{CF}\xspace}

\newcommand{\ourLLM}{\textsc{LlmMatch}\xspace}

\newcommand{\TParse}{\textsc{TParse}\xspace}
\newcommand{\PParse}{\textsc{PParse}\xspace}
\newcommand{\Atom}{\textsc{Atom}\xspace}
\newcommand{\Solve}{\textsc{Solve}\xspace}
\newcommand{\Verbalize}{\textsc{Verbalize}\xspace}

\newcommand{\satisfied}{\textsc{satisfied}\xspace}
\newcommand{\violated}{\textsc{violated}\xspace}
\newcommand{\deferred}{\textsc{deferred}\xspace}
\newcommand{\eligible}{\textsc{eligible}\xspace}
\newcommand{\ineligible}{\textsc{ineligible}\xspace}
\newcommand{\todo}[1]{\textcolor{red}{\textbf{TODO:} #1}}

\newcommand{\observed}{\textsc{observed}}
\newcommand{\imputed}{\textsc{imputed}}
\newcommand{\DLSC}{DLSC\xspace}
\newcommand{\unresolved}{\textsc{unresolved}}

\title{Accountable AI with Grounded, Faithful, Consistent, Actionable Rationales: \\
A Case Study in Clinical Trial Matching with \name\\ }

\author{%
Zikai Zhou$^{1}$ \quad Yufei Jin$^{2}$\quad Yilin Xu$^{3}$\\
\textbf{Yu-Chiang Wang$^{4}$\quad Chieh-Ju Chao$^{4}$}\quad
\textbf{Monica S. Lam}$^{1}$\vspace{-0.1in} \\
\\
$^{1}$Department of Computer Science, Stanford University, Stanford, CA, USA \\
$^{2}$Samueli Electrical and Computer Engineering, UCLA, Los Angeles, CA, USA \\
$^{3}$Department of Computer Science, Emory University, Atlanta, GA, USA \\
$^{4}$Mayo Clinic, Rochester, MN, USA \\
\texttt{zikai@stanford.edu, lam@stanford.edu}
}

\begin{document}\maketitle

\begin{abstract}

\textbf{Accountability} means a decision can be examined, justified, and contested. LLMs make this hard: fluent output may be ungrounded, incomplete, or unfaithful to the decision process. Achieving accountability requires \textbf{verified rationales} (how was the decision reached), \textbf{assumptions} (what was assumed rather than known), \textbf{policy consistency} (the same treatment for the same facts), and \textbf{pivotal conditions} (what would change the outcome). We introduce \textit{self-faithfulness} as an automatic test of accountability: changing the pivotal conditions should change the decision.

We examine accountable AI through clinical trial matching, a high-stakes task central to evidence-based medicine. Although LLM-based matchers match patients to trials reasonably accurately,
they apply decision policies inconsistently and produce rationales that are unfaithful
to their own decisions.

We introduce \name\footnote{Code, prompts, and evaluation artifacts: \url{https://github.com/stanford-oval/clinical-trial-matching}}, an LLM-based agent that translates a decision task, its constraints, and its policy into Satisfiability Modulo Theories (SMT), then derives the decision with SMT and \MaxSMT{} solvers — so policies are applied consistently and decisions are accountable by construction.

Across a SIGIR 2016-derived dataset and TREC 2021, \name achieves the strongest decision accuracy among LLM-only and neuro-symbolic baselines, applies policies with perfect consistency, and produces clinician-preferred rationales grounded in explicit assumptions and pivotal conditions, with improved counterfactual self-faithfulness.

\end{abstract}
\section{Introduction}\label{sec:intro}

\textbf{Accountability} is the property that a decision can be examined, justified, and contested after the fact: someone else can ask why, know what was assumed rather than known, trust that the decision was made consistently, and learn what would change the outcome~\cite{kroll2015accountable, binns2018algorithmic,diakopoulos2015algorithmic}.

Accountability is the essential operational requirement behind any consequential decision that must be understood, trusted, and acted upon. It means, for example, giving a patient the real reason they do not qualify for a clinical trial, clarifying what was assumed from missing lab results, explaining what would change the outcome, and knowing the decision would indeed be reversed if the patient followed the advice. Similarly, it involves providing a rejected loan applicant with the true reason for denial and clear guidance on what to improve before reapplying, and being confident that all applicants with similar qualifications are treated equally. It also means deciding that a driver gets a safe-driver discount, but warning that it would go away if he is in an at-fault accident.

LLMs produce fluent, confident output regardless of whether it is grounded, complete, or faithful to the process that generated it — hallucination. In a consequential-decision setting, this leads to four failure modes, each demanding a countermeasure.

\textbf{Grounding--Verified rationales.} A stated reason need not be the reason that actually produced a decision — explanations are generated to sound plausible, not to trace a derivation. We need verified rationales that are checked against, or derived from, the process that produced the decision.

\textbf{Faithfulness--Assumptions.} Provided evidence is often ambiguous or incomplete; however, an LLM tends to fill the gaps silently with unwarranted confidence. We need an explicit ledger of every assumption, turning silent judgment calls into auditable claims.

\textbf{Consistency--Policy adherence.} Because an LLM re-derives its reasoning in language, formally identical cases can be decided by different logic on different occasions. We need policy consistency to guarantee that the same policy is applied the same way whenever the same facts recur.

\textbf{Actionability--Pivotal conditions.} To be constructive, a decision should come with an explanation on how the decision can be reversed. We need to provide the pivotal conditions--the minimal changes that would alter the decision.

Together these four form the pillars of accountability — grounded, faithful, consistent, and actionable reasoning.
Note that none of these is addressed by improving ``trustworthiness'' in the conventional sense: better calibration or higher accuracy leaves all four gaps open, since accountability is a property of decisions and their derivation, not of the model's track record.

\paragraph{Clinical Trial Matching.}
We examine accountable AI through clinical trial matching, where clinicians want to specify policies for handling missing data, verify that a rationale is faithful to the decision, and understand the assumptions and pivotal conditions that could change it. We show that LLM-based matchers~\citep{jin2024matching,gupta2024prism,wornow2025zero} are reasonably accurate but fall short on all four accountability criteria.

An alternative is to combine language models with symbolic reasoning,
representing the matching problem as SMT (Satisfiability Modulo
Theories)~\citep{barrett2018satisfiability} formulas executed by solvers such as
Z3~\citep{de2008z3}~\citep{ye2023satlm,olausson2023linc,xu2026adaptive,pan2023logic}.
\textsc{SatIR}~\citep{zhou2026scalable} represents imputed missing data as SMT,
improving retrieval accuracy and recall, but uses an LLM to determine final
eligibility.

Building on this, we propose \name{} (VERified explanations with assumption 
Disclosure, Invariant Consistency, and Traceable pivots), a framework for
accountable decision making. Given a retrieved candidate, \name{} decides
whether the patient is eligible for the trial. We adopt \textsc{SatIR}'s trial
representation but extract patient information relative to each trial for better accuracy.
\name{} uses an SMT solver to determine eligibility; the solver trace yields
matching evidence and explicit assumptions, while a \MaxSMT{}
solver~\citep{de2008z3,bjorner2015nuz} computes pivotal conditions---the
minimal constraints that must change to reverse the decision.

\paragraph{Contributions}
\begin{itemize}[leftmargin=*, noitemsep, topsep=0pt]
\item \textbf{A formulation of clinical trial matching as accountable decision-making.} Given a decision, its constraints, and its policy, an accountable agent should apply the policy consistently and produce a decision along with (1) a formal derivation users can verify, (2) the assumptions behind it users can override, and (3) the pivotal conditions users can act on or monitor.

\item \textbf{\name{}: an accountable decision-making framework} that separates language understanding from decision making via SMT. Its inaccuracy is confined to language understanding, not decision logic, unlike monolithic LLM-based matchers, and root causes are directly observable from the representation. \name{}'s decisions come with a verifiably correct derivation, explicit assumptions, and pivotal conditions.

\item \textbf{A broad evaluation of accuracy, accountability, and actionability} across two public benchmarks, multiple model families, clinician review, policy-adherence tests, counterfactual evaluation, and failure analysis. \name{} outperforms strong LLM-only and neuro-symbolic baselines, produces clinician-preferred rationales, follows policies reliably, and has perfect rationale-decision causality by construction.
\end{itemize}

\begin{figure*}[t]
    \centering
    \includegraphics[width=0.95\textwidth]{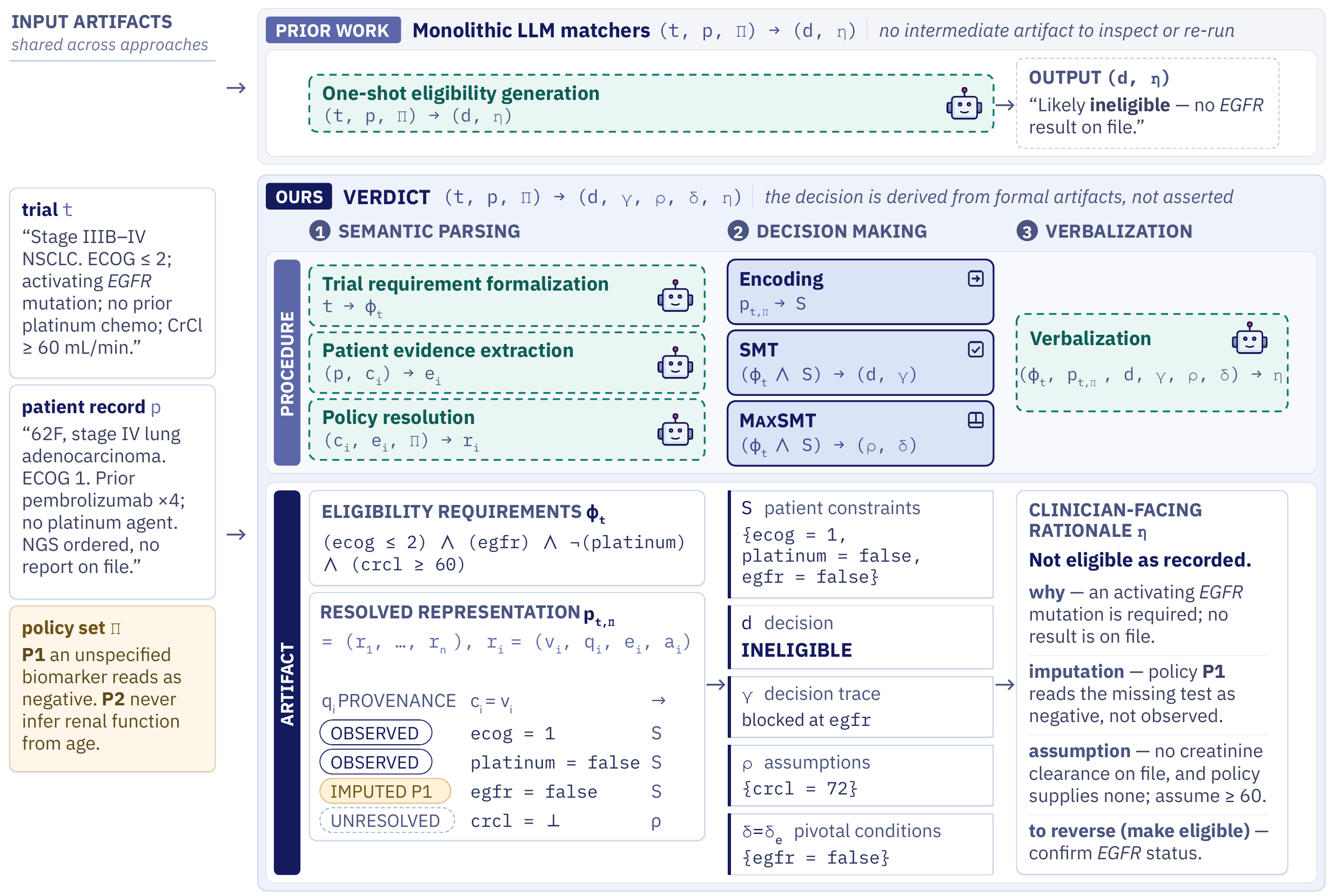}
    \vspace{-0.4em}
    \caption{\textbf{Overview of \name{}.}
    A monolithic LLM matcher (\emph{top}) asserts an eligibility decision in one
    step. \name{} (\emph{bottom}) splits the task: LLMs handle language,
    formalizing the trial into requirements $\phi_t$ and resolving patient
    evidence under policies $\Pi$ into $p_{t,\Pi}$, which marks each criterion as
    observed, imputed, or unresolved. Solvers handle the decision: SMT derives the
    decision $d$ and a trace $\gamma$; \MaxSMT{} derives the assumptions
    $\rho$ and pivotal conditions $\delta$. A final LLM step verbalizes these into a
    clinician-facing rationale $\eta$. Because the decision is derived rather than
    asserted, every artifact is inspectable, overridable, and re-runnable.}
    \vspace{-0.8em}
    \label{fig:architecture-overview}
\end{figure*}

\section{Accountable Decision Making}\label{sec:problem}

In this section, we define the general problem of accountable decision making and introduce the concept of {\em self-faithfulness} as a fundamental property of accountable decision makers. 


\subsection{Definition of Accountability}
An accountable decision-maker provides not just a decision, but also a rationale that explains how it was derived, the hidden assumptions on which it depends so they can be overridden if appropriate, and the pivotal conditions that could be changed to reverse the decision.

\begin{definition}
Given a constraint specification $y$, a case $x$, and a set of policies $\Pi$, an \textbf{accountable} decision maker is defined as 
\[
\textsc{Decide}(y,x,\Pi) = (d,\gamma,\rho,\delta),
\]
where \(d\in\{\eligible,\ineligible\}\) is the eligibility decision,
\begin{itemize}[leftmargin=*, noitemsep, topsep=0pt]
    \item the \emph{decision trace} \(\gamma\) records the logical derivation
    of \(d\) from the inputs; and
    \item the \emph{assumptions} \(\rho\) are the values the decision
    proceeds on that neither \(x\) nor \(\Pi\) establishes, indicating what
    information or actions are needed to determine eligibility;
    \item the \emph{pivotal conditions} \(\delta\) are conditions whose
    resolution or modification would flip the decision.
Note that by definition, $\delta \neq \emptyset$.
\end{itemize}
\end{definition}

\subsection{Accountable Decision Makers are Self-Faithful}
We introduce the concept of \textit{self-faithfulness} for evaluating decision-makers. If a decision-maker identifies conditions as pivotal, then changing those conditions should change the outcome. This expectation applies in both directions. In medicine, a patient may be ineligible for a clinical trial because the trial is restricted to stage IV cancer; if the patient progresses to stage IV, the decision should change from ineligible to eligible. Conversely, a patient may be eligible for a trial because their kidney function is within the required range; if a later test falls below the exclusion threshold, the decision should change from eligible to ineligible.

This notion applies to accountable decision-making more broadly, not just in medicine. For example, a driver may be ineligible for a safe-driver discount because of a recent accident, but become eligible again after enough accident-free years; conversely, a driver who currently qualifies for the discount should lose it after an at-fault accident. In each case, the stated pivotal condition is expected to determine the direction of the decision.

\begin{definition}
Given a case $x$, policies $\Pi$, and a set $\delta$ of condition values the
decision relies on, the
\textbf{counterfactual construction} $\CF(x,\Pi,\delta)$ denotes a minimally
modified case where $\delta$ no longer holds.
\end{definition}

\begin{theorem}
The \textsc{Decide} function is {\bf self-faithful} with respect to its pivotal conditions: if the pivotal conditions of a case are flipped, so will the decision.
Given a constraint specification $y$, a case $x$, and policies $\Pi$, if
\[
\begin{array}{rcl}
(d,\gamma,\rho,\delta) &=& \textsc{Decide}(y,x,\Pi),\\
(d',\gamma',\rho',\delta') &=& \textsc{Decide}(y,\CF(x,\Pi,\delta),\Pi),
\end{array}
\] then $
d \neq d'.
$
\label{thm:self-faithful}
\end{theorem}

\subsection{Evaluating Self-Faithfulness}

Theorem~\ref{thm:self-faithful} yields an automatic evaluation metric, requiring no external annotation, for measuring the accountability of decision-maker implementations.

\begin{definition}
Given a decision-maker implementation \textsc{DM}, policies $\Pi$, and a set of evaluation instances
$
M=\{(y_1,x_1),\dots,(y_n,x_n)\},
$
define
\[
\begin{aligned}
&\operatorname{PivotalFlipRate}(\textsc{DM}, M, \Pi)\\
&\qquad =
\frac{1}{|M_\delta|}\sum_{j \in M_\delta}
\mathbb{I}\!\left[d_j \neq d'_j\right].
\end{aligned}
\vspace{-1em}
\]

where
\[
\begin{array}{rcl}
(d_j,\gamma_j,\rho_j,\delta_j) &=& \textsc{DM}(y_j,x_j,\Pi),\\
(d'_j,\gamma'_j,\rho'_j,\delta'_j) &=& \textsc{DM}(y_j,\CF(x_j,\delta_j),\Pi).
\end{array}
\]
\end{definition}

\begin{corollary}
Given policies $\Pi$ and a set of evaluation instances
$
M=\{(y_1,x_1),\dots,(y_n,x_n)\},
$
\[
\operatorname{PivotalFlipRate}(\textsc{Decide}, M, \Pi)=1.
\]
\end{corollary}

\section{\name{} Clinical Trial Matcher}

Figure~\ref{fig:architecture-overview} overviews \name{} and its intermediate
artifacts. We decompose clinical trial matching into language understanding and
decision making: LLMs parse trial criteria, patient records, and policies
into a formal representation, while SMT and \MaxSMT{} solvers derive the
decision and accountable rationale. We next define the representation and
present the \name{} algorithm.

\subsection{Formal \SMT Representation}
\label{sec:formal-representation}
Given a trial, patient record, and policy, all written in natural language, the clinical trial matching problem is to produce a decision on whether the patient is eligible, along with an accountable rationale. 

Policies define how conditions should be resolved if they are not specified in a patient's record. They may encode common practices, institutional choices, or default user assumptions.

There are two kinds of policies: the first depends on the full context of the
trial and other rules in the policy set; the second is purely a function of
the kind of constraint. In principle, such policies may also depend on the
resolution state of the constraint (e.g., the strength or direction of
available patient evidence); in our current formulation, however, all
unresolved conditions are represented by a single unknown value \(\bot\).

The trial representation builds on the semantic formalization introduced in \textsc{SatIR}~\citep{zhou2026scalable}. \textsc{SatIR} decomposes natural-language eligibility criteria into atomic, typed conditions and their logical composition, yielding the trial-side constraint representation used here. \name{} extends this representation from retrieval to decision making by grounding conditions in patient evidence, resolving missing values under policies.

\begin{definition}
The \textbf{eligibility requirements} of a \textit{trial} $t$ in natural language are formalized as the logical expression $\phi_t=\phi(c_1,\dots,c_n)$
where\\
\indent $c = (\tau, V, u, \ell) \in C$ denotes the smallest checkable requirement,\\
\indent $\tau \in \mathcal{T}$ = \{diagnosis, medication, lab value, timing, $\ldots$\} denotes the kind of condition,\\
\indent $V$ denotes the domain of the condition (e.g., Boolean for a diagnosis, numeric for a lab value, a date for timing),\\
\indent $u$ is the textual description, and\\
\indent $\ell$ links to the source text for checkability.
\end{definition}

\begin{definition} 
Let $p$ be the natural-language record of a patient, and let
$\phi_t=\phi(c_1,\ldots,c_n)$ denote the formalized eligibility requirements
of trial $t$. For a policy set $\Pi$, the \textbf{patient's resolved representation
with respect to trial} $t$ is\\
\indent \(p_{t,\Pi}\in P=(r_1,\ldots,r_n)\), where\\
\indent \(r_i=(v_i,q_i,e_i,a_i)\) records the resolution of constraint \(c_i\),\\
\indent \(v_i\in V_{c_i}\cup\{\bot\}\) denotes the resolved value of \(c_i\),\\
\indent \(q_i\in\{\observed,\imputed,\unresolved\}\) indicates if \(v_i\)
is determined by patient evidence, supplied by policy resolution, or remains unknown,\\
\indent \(e_i\) denotes patient evidence supporting \(v_i\), with
\(e_i=\varnothing\) when no such evidence is available, and\\
\indent \(a_i\) denotes the imputation record produced by policy resolution,
describing or recording the basis for an imputed value, with
\(a_i=\varnothing\) when no imputation is made.
That is, 
\[
r_i=
\begin{cases}
(v_i,\observed,e_i,\varnothing),\\ 
 \quad \textrm{if } v_i \textrm{ is observed in evidence } e_i,\\
(v_i,\imputed,\varnothing,a_i),\\
  \quad  \textrm{if no value is observed in the record,}\\
  \quad \textrm{and } v_i \textrm{ is given by imputation } a_i  \\
  \quad \textrm{from }\textrm{applying policies } \Pi\\
(\bot,\unresolved,\varnothing,\varnothing),\\
    \quad \textrm{if no value is observed} \\
    \quad \textrm{or imputed with } \Pi
    \end{cases}
\]
\end{definition}

The representation \(p_{t,\Pi}\) is constructed by first extracting patient
evidence relevant to each condition \(c_i\). If the evidence determines the
condition value, it is marked \observed. Otherwise, the applicable policies
\(\pi_i\subseteq\Pi\) are resolved to a pair \((v_i,a_i)\): if they supply a
default value, the condition is marked \imputed{} and \(a_i\) records the
corresponding policy realization; if they do not, the condition is marked
\unresolved{} with \(v_i=\bot\) and \(a_i=\varnothing\).

\subsection{The \name Algorithm}
\label{sec:formal-matching}

\label{sec:decision-artifacts}

\noindent
\textbf{Algorithm} \name{} derives an eligibility decision and its formal accountability
artifacts using SMT and \MaxSMT{}, then verbalizes them for review.

\noindent
\textbf{Input:} Trial \(t\), patient record \(p\), and policies \(\Pi\).\\
\textbf{Output:} Eligibility decision \(d\), decision trace \(\gamma\),
assumptions \(\rho\), pivotal conditions \(\delta\), and
clinician-facing rationale \(\eta\).

\noindent
{\bf Step 1.}
Derive the formal representation with LLMs: eligibility requirements for trial $t$, $\phi_t=\phi(c_1,\dots,c_n)$, and the resolved representation for patient $p$ w.r.t. $\Pi$, $p_{t, \Pi}$, as discussed in Section~\ref{sec:formal-representation}.  
Let the patient constraints imposed by all the resolved patient values for the trial be
\[
\begin{array}{ll}
S = \bigwedge_i \{ c_i = v_i | &r_i = (v_i, \observed, \ldots ) \vee \\
&r_i = (v_i, \imputed, \ldots)\}
\end{array}
\]

\noindent

\noindent
{\bf Step 2.} Compute the decision $d$ and decision trace $\gamma$. Let
\[
(z,\gamma)=\textsc{SMT}(\phi_t \wedge S),
\]
where $z\in\{\textsc{sat},\textsc{unsat}\}$ and $\gamma$ is the trace of the
derivation; $d=\eligible$ iff $z=\textsc{sat}$, and $\ineligible$ otherwise.
The SMT solver is verifiably correct and, when possible, finds an assignment
to unresolved variables that makes the constraints \textsc{sat}.

\noindent
{\bf \MaxSMT.} To compute the assumptions and pivotal conditions, Steps 3--5 use \MaxSMT (weighted maximum satisfiability modulo theories).
Given a formula $F$ with non-negative soft-clause weights $W$,
$\MaxSMT(F,W)$ returns a model maximizing the total weight of satisfied soft
clauses. Let $\MaxSMT(F,W).x$ denote the value it assigns to clause $x$. 

All constraints in the trial constraints $\phi_t$ are considered hard, meaning that none of these constraints can be violated. Note that it is always possible to satisfy all the constraints in $\phi_t$ of a well-formed clinical trial. 
We define $W$ such that every clause of $S$ is considered soft and given unit weight. $\MaxSMT(\phi_t \wedge S, W)$ thus returns  a model that minimizes changes to the resolved conditions $S$. 

\noindent
{\bf Step 3.} Find $\delta_\textrm{E}$, the conditions that need to change to
render the decision eligible.
\[
\begin{aligned}
\delta_\textrm{E} = \{\,s \mid\; &s \textrm{ is a clause in } S, \\
& \textsc{MaxSMT}(\phi_t \wedge S, W).s = \textsc{false} \,\}.
\end{aligned}
\]
Note that $\delta_\textrm{E} = \emptyset$ iff $d=\eligible$.

\noindent
{\bf Step 4}. Compute the assumptions $\rho$ that maximize eligibility.
\[
\begin{array}{ll}
\rho=  \bigwedge_i \{\,c_i = & \textsc{MaxSMT}(\phi_t \wedge S, W).c_i \\
&\mid q_i=\unresolved\,\}
\end{array}
\]

\noindent
{\bf Step 5.} Find $\delta_{\textrm{I}}$, 
the conditions that need to change to render the decision \ineligible. We define $W'$ such that every clause of $S \wedge \rho$ is considered soft and given unit weight.
\[
\begin{aligned}
\delta_{\textrm{I}}
= \{\,x \mid\;&
x \textrm{ is a clause in } S \wedge \rho, \\
&\textsc{MaxSMT}(\neg\phi_t \wedge S \wedge \rho, W').x \\
&= \textsc{false}
\,\}.
\end{aligned}
\]
Symmetrically, $\delta_{\textrm{I}} = \emptyset$ iff 
$d=\ineligible$.

\noindent
{\bf Step 6.} Find the pivotal conditions $\delta$. 
\[
\delta=
\begin{cases}
\delta_{\textrm{I}}, & \textrm{if } d=\eligible\\
\delta_{\textrm{E}}, & \textrm{if } d=\ineligible.
\end{cases}
\]
Note that $\delta \neq \emptyset$.  

\noindent
\textbf{Step 7.} Present the rationale \(\eta\) in natural language.
The LLM verbalizes the decision, grounded in \(d\), \(\gamma\), the
condition-level evidence and imputations in \(p_{t,\Pi}\), \(\rho\), and
\(\delta\) (Appendix~\ref{app:verbalizer}). For numeric conditions in
\(\rho\) and \(\delta\), it reports the requirement \(\phi_t\) imposes
rather than the witness \(A\) assigns, since the witness is arbitrary
within the satisfying region.

\subsection{Accountability of \name}

\begin{theorem}
\label{thm:verdict-accountable}
\name is, by construction, \textbf{verifiably accountable} with respect to the \textit{formal representation} of the trial and patient records. The decision, trace, assumptions, and pivotal conditions are verifiably correct by virtue of the correctness of the \SMT and \MaxSMT solvers. 
\end{theorem}


\begin{corollary}
\label{cor:verdict-flip}
Let \(\Pi\) be a set of policies and
$
M=\{(y_1,x_1),\dots,(y_n,x_n)\}
$
a set of evaluation instances in \name's \textit{formal representation}, 
\[
\operatorname{PivotalFlipRate}(\name{}, M, \Pi)=1.
\]
\end{corollary}
With respect to constraints in its \textit{formal representation}, \name{} is
self-faithful: for every case $x_j$, applying the counterfactual construction
\(\CF(x_j,\delta_j)\) to the pivotal conditions \(\delta_j\) returned by
\name{} flips the eligibility decision.

\name{}, however, is not necessarily accountable with respect to the \textit{textual} trial
specification, patient record, and policies: the LLM may err in translating
them or applying policies. The representation remains inspectable, however, so
a reader who spots an issue in the rationale can trace it back to the source
text and revise the assumption, the translation, or the decision.

\section{Experimental Setup}
\label{sec:experimental_setup}

We describe the benchmarks, systems, and LLM backbones used in our evaluation.

\subsection{Benchmarks}
\label{sec:benchmarks}

We evaluate on two benchmarks with independently constructed eligibility
labels.

\paragraph{SIGIR 2016-derived benchmark.}
We use 552 patient--trial pairs from the SIGIR 2016 clinical-trial matching
benchmark~\citep{koopman2016test}. Because its judgments measure retrieval
relevance rather than eligibility, we derive binary labels using a five-judge
GPT-5 panel and validate a stratified subset with a clinician. These labels
were not used to develop \name{}. Appendix~\ref{app:reference_validation}
details the annotation and validation procedures.

\paragraph{TREC 2021.}
We use 363 patient–trial pairs sampled from the TREC 2021 Clinical Trials track~\citep{soboroff2021overview}, drawn from the pairs carrying an official eligibility judgment. All systems evaluate on the same pairs. This benchmark was not used for method development, prompt design, or policy construction.

\subsection{Systems and Backbones}
\label{sec:systems}

\paragraph{\name{}.}
We apply the \name{} algorithm, using the 
backbone LLM to formalize trial requirements and extract patient evidence (Appendix \ref{app:system_modules}).

\paragraph{\ourLLM{}.}
Our optimized natural-language matcher that directly predicts trial-level
eligibility using task-specific prompts, without constructing a formal decision
program or invoking a separate decision executor (Appendix \ref{app:baseline_ourllm}).

\paragraph{CoT LLM.}
Our chain-of-thought matcher that reasons over eligibility criteria before
producing a trial-level decision (Appendix~\ref{app:baseline_cot}).

\paragraph{ZSPM~\citep{wornow2025zero}.}
~\citet{wornow2025zero} do
not name their system; we write ZSPM after its
title, \emph{Zero-Shot clinical trial Patient Matching}. ZSPM is a zero-shot LLM-based matcher that predicts eligibility directly from the patient record
and trial criteria (Appendix \ref{app:baseline_zspm}).

\paragraph{TrialGPT~\citep{jin2024matching}.}
An end-to-end LLM framework for retrieving and matching clinical trials. We
use TrialGPT-Matching, its criterion-level eligibility component, and aggregate
its predictions into a trial-level decision (Appendix \ref{app:baseline_trialgpt}).

\paragraph{\DLSC{}~\citep{xu2026adaptive}.}
\citet{xu2026adaptive} do not name their system; we write \DLSC{} after its
title, \emph{dynamic logical solver composition}. It decomposes
natural-language problems, composes symbolic solvers, and autoformalizes each
instance into solver-specific code at inference time. We adapt its SMT
configuration to clinical-trial eligibility, executing the generated program to
produce the decision. Whereas \DLSC{} targets reasoning problems, \name{}
targets accountable decision making, deriving inspectable artifacts and
accepting an explicit policy set their method lacks.

For each benchmark, all systems use the same backbone and patient--trial
pairs. Due to budget constraints, we select complementary backbone suites to
test robustness across model capabilities and families. On the SIGIR
2016-derived benchmark, we evaluate GPT-4.1, GPT-4o, and
GPT-4o-mini~\cite{achiam2023gpt,hurst2024gpt}; on TREC 2021, we compare
proprietary and open-weight models using
GPT-5-mini~\cite{singh2025openai}, Claude Haiku
4.5~\cite{HaikuSystemCC}, and
Qwen2.5-7B-Instruct~\cite{Yang2024Qwen25TR}. The baseline suites differ because
the SIGIR evaluation predates \DLSC{}: TREC includes it as the closest neuro-symbolic system but omits TrialGPT, which ZSPM outperformed across SIGIR backbones.

\section{Evaluation}
\label{sec:evaluation}

We evaluate clinical-trial matching systems on decision accuracy (Tables~\ref{tab:trialelig_accuracy}
and~\ref{tab:trec_accuracy}), and on the four
accountability properties introduced in Section~\ref{sec:intro}:
grounding and actionability (Tables~\ref{tab:rationale_preference} and~\ref{tab:rationale_axes}), 
consistency (Table~\ref{tab:policy_adherence}), and 
self-faithfulness 
(Table~\ref{tab:counterfactual_raw}).

\subsection{Decision Accuracy}
\label{sec:eval-accuracy}

We evaluate whether the predicted eligibility decision agrees with the
benchmark annotation, reporting F1 and accuracy. This is an
end-to-end evaluation: systems receive the patient record and trial criteria
and must interpret both the available evidence and any missing information.
Note that the benchmark does not explicitly specify policies on how unreported
patient information should be handled. Thus, the evaluation includes how well
a decision maker matches the implicit policies.

We first ask whether an intermediate formal representation compromises predictive performance. It does not: \textbf{\name{} achieves the highest F1 across all backbones on both benchmarks} (Tables~\ref{tab:trialelig_accuracy} and~\ref{tab:trec_accuracy}). On the SIGIR 2016-derived benchmark, it achieves 0.900, 0.836, and 0.754 F1 with GPT-4.1, GPT-4o, and GPT-4o-mini.

\begin{table}[htbp]
\centering
\small
\begin{tabular}{llrr}
\toprule
Backbone & System & F1 & Accuracy \\
\midrule
GPT-4.1
& \name{}      & \textbf{0.900} & \textbf{0.902} \\
& \ourLLM{}    & 0.884 & 0.879 \\
& CoT LLM      & 0.721 & 0.775 \\
& TrialGPT     & 0.359 & 0.592 \\
& ZSPM         & 0.835 & 0.828 \\
\midrule
GPT-4o
& \name{}      & \textbf{0.836} & \textbf{0.826} \\
& \ourLLM{}    & 0.769 & 0.803 \\
& CoT LLM      & 0.773 & 0.804 \\
& TrialGPT     & 0.759 & 0.774 \\
& ZSPM         & 0.806 & 0.797 \\
\midrule
GPT-4o-mini
& \name{}      & \textbf{0.754} & \textbf{0.766} \\
& \ourLLM{}    & 0.476 & 0.649 \\
& CoT LLM      & 0.599 & 0.699 \\
& TrialGPT     & 0.588 & 0.685 \\
& ZSPM         & 0.691 & 0.572 \\
\bottomrule
\end{tabular}
\caption{Patient-level eligibility performance on the SIGIR 2016-derived
benchmark.}
\label{tab:trialelig_accuracy}
\vspace{-.5em}
\end{table}

On the independent TREC 2021 benchmark, \name{} again performs best across all
three backbones. Alongside the natural-language baselines, we include
\DLSC{}~\cite{xu2026adaptive}, the closest prior LLM--solver matcher, adapted to
clinical-trial eligibility (Section~\ref{sec:systems}). This comparison
distinguishes \name{} from another system that also uses a formal solver rather
than comparing solver-based and natural-language decision making alone.


\begin{table}[htbp]
\centering
\small
\setlength{\tabcolsep}{3pt}
\begin{adjustbox}{max width=\columnwidth}
\begin{tabular}{lccccc}
\toprule
Backbone & \name{} & \ourLLM{} & CoT LLM & ZSPM & \DLSC{} \\
\midrule
GPT-5-mini
& \textbf{0.828} & 0.776 & 0.711 & 0.759 & 0.700 \\
Claude Haiku 4.5
& \textbf{0.800} & 0.587 & 0.704 & 0.615 & 0.700 \\
Qwen2.5-7B
& \textbf{0.738} & 0.649 & 0.663 & 0.695 & 0.470 \\
\bottomrule
\end{tabular}
\end{adjustbox}
\caption{F1 on 363 TREC 2021 patient--trial pairs using official judgments.
Best F1 per backbone is bolded. Distilling the Qwen2.5-7B formalizer improves
\name{} from 0.738 to 0.829 F1, matching GPT-5-mini.}
\label{tab:trec_accuracy}
\vspace{-1em}
\end{table}

The corresponding TREC accuracies are 0.838, 0.815, and 0.697.
Across backbones, \name{} exceeds the strongest natural-language baseline by
0.04--0.10 F1 and \DLSC{} by 0.10--0.27 F1
(accuracy: 0.838/0.815/0.697 vs.\ 0.73/0.70/0.56).
Since \DLSC{} also uses a formal solver, the gain reflects the importance of \name{}'s
decomposition of trial formalization, evidence assignment, missingness
resolution, and decision execution. 
Distilling the Qwen2.5-7B formalizer raises F1 from 0.738 to 0.829, a performance matching the larger and proprietary GPT-5-mini model. 

\paragraph{Failure-case analysis.}
We manually traced all 59 TREC disagreements for \name{} with GPT-5-mini.
Most arise from missing-evidence imputations (31/59; 53\%), where criteria are
parsed correctly but unreported patient information is resolved differently
from the benchmark's implicit policy. Another 14 cases (24\%) reflect a mismatch
between strict eligibility and broader TREC relevance judgments. Only 10 cases
(17\%) stem from semantic parsing errors; three (5\%) arise from enforcing
non-chart-decidable operational criteria, and one (2\%) from an
incomplete trial input in which the inclusion-criteria section was missing. Thus, semantic parsing errors affect only 10 of the 363 final decisions (2.8\%);
the main remaining challenge is handling missing evidence and criteria that
cannot be resolved from the patient chart.

\subsection{Grounding and Actionability}
\label{sec:eval-rationale}
We next evaluate whether clinician-facing rationales are grounded in the
evidence and imputations supporting the decision and whether they expose
information useful for subsequent clinical review.

\paragraph{Clinician comparison.}
We evaluate grounding and actionability through blinded clinician comparisons of
\name{} against \textsc{ZSPM} and \textsc{LLMMatch}. Each uses a separate
stratified sample of 16 patient--trial pairs from the 552-pair SIGIR-derived
benchmark, four from each of four outcome strata: both systems match the
reference label, only \name{} does, only the comparator does, and neither does.
Absolute ratings are therefore comparable within, not across, the two samples.
The clinician sees anonymized rationales for each pair, selects the more useful
one or a tie, and rates both on criterion completeness, chart traceability,
coherence, actionability, and decision support. Appendix~\ref{app:reference_validation}
gives full sampling details.

\begin{table}[htbp]
\centering
\setlength{\tabcolsep}{4pt}
\renewcommand{\arraystretch}{1.08}

\resizebox{\columnwidth}{!}{%
\begin{tabular}{lcccc}
\toprule
Comparator
& \makecell{\name{}\\wins}
& Ties
& \makecell{Comparator\\wins}
& \makecell{Win\\rate} \\
\midrule
ZSPM      & 13 & 3 & 0 & 90.6\% \\
\ourLLM{} & 11 & 2 & 3 & 75.0\% \\
\bottomrule
\end{tabular}%
}

\caption{Clinician pairwise preferences for \name{} rationales. Ties count as half a win.}
\label{tab:rationale_preference}
\vspace{-0.7em}
\end{table}

The clinician prefers \name{}'s solver-grounded rationales over both 
systems compared. Against ZSPM, \name{} wins 13 of 16 comparisons and ties three,
yielding a 90.6\% tie-adjusted win rate. Against \ourLLM{}, it wins 11, ties
two, and loses three, for a 75.0\% win rate.









\paragraph{Clinician Ratings.}
Table~\ref{tab:rationale_axes} reports clinician ratings along five dimensions. 
\textit{Criterion completeness} measures whether the rationale addresses
decision-relevant trial criteria, while \textit{Chart traceability} measures
whether its claims trace to patient evidence or explicit policy-supplied
imputations. Together, they evaluate grounding in the decision
basis. 
\textit{Coherence} measures whether the claims in the rationale are logically consistent with each other;
\textit{Actionability} measures whether the rationale identifies
unresolved information and decision-changing conditions;
and overall \textit{Decision support} measures whether it helps the clinician understand,
verify, and act on the decision.

\begin{table}[htbp]
\centering
\small
\begin{tabular}{lrr}
\toprule
Dimension & \name{} & ZSPM \\
\midrule
Criterion completeness & 5.00 & 4.69 \\
Chart traceability     & 5.00 & 4.25 \\
Coherence    & 5.00 & 3.62 \\
Actionability          & 5.00 & 3.12 \\
Decision support                & 4.94 & 3.81 \\
\midrule
\toprule
Dimension & \name{} & \ourLLM \\
\midrule
Criterion completeness & 4.81 & 3.75 \\
Chart traceability     & 4.81 & 3.88 \\
Coherence   & 4.69 & 4.31 \\
Actionability          & 4.69 & 3.12 \\
Decision support                & 4.62 & 4.00 \\
\bottomrule
\end{tabular}
\caption{Mean clinician ratings of rationale quality (1--5, higher better)
vs.\ ZSPM and \ourLLM{}.}
\label{tab:rationale_axes}
\vspace{-1em}
\end{table}

\name{} receives higher mean ratings on all five dimensions against both
comparators. Higher criterion-completeness, chart-traceability, and coherence follow from the trace generated by the SMT solver. Higher actionability and decision-support
scores follow from the assumptions and pivotal conditions, exposing what
remains unresolved and what could change the decision. 

\subsection{Consistency}
\label{sec:eval-policy}

We next evaluate whether matchers consistently apply the specified policy set when resolving conditions not determined by patient evidence. \name{} adheres to these policies once the relevant condition types, evidence states, and policy-resolved values are correctly
determined.

Natural-language matchers are evaluated under four progressively explicit settings: \textbf{(1) Unassisted:} the matcher receives only the patient record and trial criteria; \textbf{(2) Rules provided:} the policy set is included in the prompt; \textbf{(3) Structured evidence:} the matcher additionally receives verified atomic conditions, their types, and evidence-based values, with missing values represented as $\bot$; and \textbf{(4) Applied per condition:} for each condition with a null evidence-based value, the policy is mechanically selected by type, and the LLM resolver used in \name{} generates condition-level values and assumption records before the trial-level decision.

Policy adherence generally improves with structure. Applying policies per
criterion yields the highest adherence, yet natural-language matchers still
follow $\Pi$ in only 71--81\% of cases (Table~\ref{tab:policy_adherence}).
This improvement over structured evidence alone is significant (McNemar's exact test, \(p<0.01\)).


\begin{table}[htbp]
\centering
\small
\setlength{\tabcolsep}{3pt}
\renewcommand{\arraystretch}{1.08}
\begin{adjustbox}{max width=\columnwidth}
\begin{tabular}{@{}llcccc@{}}
\toprule
Matcher & Model
& Unassisted
& \makecell{Rules\\provided}
& \makecell{Structured\\evidence}
& \makecell{Applied per\\criterion} \\
\midrule
\ourLLM{} & 5-mini & 67.5 & 64.8 & 68.9 & 71.9 \\
\ourLLM{} & 5      & 70.2 & 67.9 & 69.9 & 71.4 \\
ZSPM      & 5-mini & 66.3 & 67.3 & 71.9 & 80.1 \\
ZSPM      & 5      & 62.8 & 66.3 & 69.4 & 81.1 \\
\bottomrule
\end{tabular}
\end{adjustbox}
\caption{Agreement (\%) with the decision required by the explicit
missingness policy. GPT-5-mini and GPT-5 are abbreviated as 5-mini and 5.}
\label{tab:policy_adherence}
\vspace{-1em}
\end{table}

The structured-evidence settings separate policy execution from upstream
formalization and evidence extraction. Appendix~\ref{sec:parsing-audit} finds
99.6\% condition-level semantic preservation, 94.3\% fully correct criteria,
and parsing errors affecting only 10/363 decisions (2.8\%). 

\subsection{Self-Faithfulness}
\label{sec:eval-counterfactual}

We test whether the pivotal conditions reported by LLM-based matchers are self-faithful for \textsc{Ineligible} decisions. To compute the \textsc{PivotalFlipRate}, we modify the original record so that the pivotal conditions no longer hold. Because this process may introduce contradictions, a separate LLM verifier retains only valid counterfactuals, over which we compute the \textsc{PivotalFlipRate}.

For this evaluation, we rerun ZSPM with a counterfactual-specific prompt,
distinct from its accuracy prompt, that elicits a binary verdict and
evidence-status labels for identifying chart-supported blockers. We also
evaluate \textsc{LLMMatch-Pivotal}, explicitly prompted to enumerate changes
required for eligibility. Details are in Appendix~\ref{app:cf_details}.

Natural-language matchers frequently retain their original decision even after
their stated rejection reasons have been addressed. The strongest baseline
flips its decisions only on 65.0\% of validated cases, while \ourLLM{} flips on 48.9\%.
Explicitly prompting \ourLLM{} to enumerate the changes required for
eligibility, \ourLLM{}-Pivotal,
raises the rate only to 57.7\%.

The LLM-based verifier for counterfactual construction may err. A clinician
audits a stratified subsample and finds that roughly 80\% of modifications are valid
(Table~\ref{tab:cf_audit_only}). We report the clinician-adjusted rate in
Appendix~\ref{app:cf_details}.


\begin{table}[htbp]
\centering
\footnotesize
\begin{tabular}{llr}
\toprule
System & Backbone & \textsc{PivotalFlipRate} \\
\midrule
CoT LLM                & GPT-5   & 59.5\% \\
CoT LLM                & GPT-4.1 & 65.0\% \\
TrialGPT               & GPT-4.1 & 61.1\% \\
\ourLLM{}-Pivotal & GPT-4.1 & 57.7\% \\
\ourLLM{}              & GPT-4.1 & 48.9\% \\
ZSPM                   & GPT-4.1 & 26.6\% \\
\bottomrule
\end{tabular}
\caption{Raw ineligible-to-eligible \textsc{PivotalFlipRate} for
natural-language matchers.}
\label{tab:counterfactual_raw}
\end{table}
\vspace{-1em}

These results show that asking a natural-language matcher to enumerate rejection reasons does not ensure that they causally drive its decision, whereas in \name{} the decision and pivotal conditions are derived verifiably, so their relationship holds by construction, given formalized inputs.

\section{Related Work}
\paragraph{Clinical trial matching and retrieval.}
Patient--trial matching spans executable cohort identification and LLM-based
methods. Criteria2Query and
CriteriaMapper~\citep{yuan2019criteria2query,lee2024criteriamapper} map
eligibility criteria to executable EHR queries. TrialGPT~\citep{jin2024matching},
zero-shot matching~\citep{wornow2025zero}, and PRISM~\citep{gupta2024prism}
interpret patient records and trial criteria with LLMs.
\textsc{SatIR}~\citep{zhou2026scalable} formalizes patient and trial constraints
for high-recall retrieval. Unlike retrieval, this stage must resolve missing
evidence and produce an auditable eligibility decision. We derive eligibility
and its rationale for a candidate pair from an executable representation.

\paragraph{LLM--symbolic reasoning and faithful execution.}
Prior work uses LLM semantic parsing and symbolic execution to improve
reliability. SatLM~\citep{ye2023satlm}, Logic-LM~\citep{pan2023logic}, and
LINC~\citep{olausson2023linc} translate natural-language problems into logical
forms for symbolic solvers; Faithful CoT~\citep{lyu2023faithful}
deterministically executes explicit reasoning representations.
\DLSC{}~\citep{xu2026adaptive}, the closest neuro-symbolic method in our
evaluation, dynamically composes solvers and generates solver-specific
formalizations. These methods primarily use symbolic execution to improve
accuracy or reliability. We instead use executable representations to define
an authoritative decision state from which verdicts, policy applications,
assumptions, and pivotal conditions are mechanically derived.
Our experiments show that symbolic execution alone need not outperform strong
natural-language approaches; gains require explicit treatment of missing
evidence, assumptions, and decision policies.

\paragraph{Faithful explanations and accountable decisions.}
Interpretability distinguishes plausible from faithful
explanations~\citep{jacovi2020towards}, while LLM reasoning can fail causal
faithfulness tests~\citep{lanham2023measuring}. Algorithmic accountability
requires decisions to be inspectable, justifiable, and
contestable~\citep{kroll2015accountable,binns2018algorithmic,
diakopoulos2015algorithmic}, while counterfactual explanations and recourse
identify outcome-changing conditions~\citep{wachter2017counterfactual,
ustun2019actionable}. Our formulation unifies them: an
accountable decision provides its derivation, unresolved assumptions,
consistently applied policies, and pivotal conditions testably linked to the
decision.
\section{Conclusion}
This paper introduces accountability as a requirement for high-stakes AI decisions—encompassing grounding, faithfulness, consistency, and actionability. Even when LLM matchers decide eligibility correctly, they often lack accountability: inconsistent explanations, relevant conditions omitted, or reasoning misrepresented. This poses clinical risks because clinicians need reliable rationales to verify decisions and manage care.

\name{} addresses these challenges by separating reasoning from language understanding, constraining the LLM to produce explicit, auditable representations. Errors may remain, including misinterpretation of patient records or incorrect formalization, but it eliminates reasoning mistakes and makes the decision process transparent and reviewable. This allows users to trace each rationale step and identify, correct, or override decisions when necessary, supporting robust accountability.

Our results demonstrate that formal methods deliver, by construction, levels of accountability, transparency, and contestability unattainable by language-based models alone. By making accountability foundational to AI system design, we enable rigorous oversight and reliable deployment in critical domains such as healthcare.

\clearpage
\section*{Limitations}
\label{sec:limitations}

This paper has several limitations. First, although our system makes the final
matching decision explicit and rerunnable, it still relies on LLMs to formalize
trial criteria, extract patient evidence, and apply context-dependent policies.
If an LLM misses a fact, extracts the wrong evidence, or represents a requirement
incorrectly, the resulting formal representation---and therefore the final
decision---can still be wrong. The SMT and \MaxSMT{} steps make reasoning over the
encoded representation verifiable, but they do not guarantee that the
representation itself is clinically correct. Our semantic-parsing audit suggests
that such errors are relatively uncommon, but evidence extraction and
missing-information resolution remain important upstream sources of error.

Second, our counterfactual self-faithfulness evaluation depends on generated
counterfactual patient records. We use LLMs to modify the patient record according
to the matcher’s reported pivotal conditions, and a separate LLM validator to
determine whether the resulting edit correctly implements those changes without
introducing new eligibility violations and remains clinically realistic and
coherent. Because both stages can make errors, we additionally audit a stratified
subset with a clinician and report clinician-adjusted estimates. Nevertheless,
larger-scale review by multiple clinicians would provide stronger evidence about
counterfactual validity. For our system in particular, the end-to-end evaluation
reruns patient-side semantic parsing on the modified record, so failures may
reflect imperfect counterfactual rewriting or re-parsing rather than a mismatch
between the pivotal conditions and the formal decision boundary.

Third, although self-faithfulness is defined in both directions, our empirical
counterfactual evaluation focuses on initially ineligible cases and tests whether
modifying the reported pivotal conditions changes the decision to eligible. This
direction is the more immediately actionable one for trial screening, but it does
not test whether modifying pivotal conditions for an initially eligible case
would likewise change the decision to ineligible. Evaluating both directions,
including changes that arise as a patient's state evolves over time, would provide
a more complete empirical test of self-faithfulness.

Fourth, the available benchmarks are imperfect proxies for strict clinical-trial
eligibility. On the SIGIR 2016-derived benchmark, the original judgments concern
referral relevance rather than eligibility, so we construct binary eligibility
labels for 552 retrieved patient--trial pairs using a five-judge LLM panel and
validate a stratified subset with a clinician. These labels enable larger-scale
comparison but are not equivalent to adjudication by a multi-clinician panel, and
the candidate set is restricted to top-ranked retrieved trials rather than the
full trial corpus. TREC 2021 provides an independent benchmark with official
human judgments and was not used during method development, prompt design, or
policy construction. However, its judgments were also created for a retrieval
task and do not always coincide with strict chart-level eligibility; our failure
analysis identifies cases in which broader TREC relevance judgments conflict
with the eligibility interpretation. Neither benchmark explicitly specifies how
unreported patient information should be resolved, so decision accuracy also
partly measures agreement with the implicit missing-information policy reflected
in the reference labels.

Finally, the patient records in both benchmarks are synthetic rather than real longitudinal electronic health records. This avoids privacy barriers and enables controlled evaluation, but may underrepresent the noise, redundancy, conflicting evidence, missingness, and temporal complexity of real clinical documentation. The SIGIR-derived records are also unusually compact relative to hospital EHR notes. Testing the framework on real patient records under appropriate privacy safeguards and prospective clinical review remains important future work.
\section*{Ethical Considerations}
\label{sec:ethics}

Clinical trial matching is a high-stakes task: incorrect recommendations may
affect patient access to therapy, clinician workload, and trial recruitment.
Our system is intended as a decision-support tool for surfacing candidate
matches, imputations, and assumptions, not as an
autonomous enrollment or exclusion system. Final eligibility decisions should
remain under qualified clinical and trial-site review.

Our framework reduces some risks of end-to-end LLM matching by separating
clinical text interpretation from the final eligibility decision. The solver
makes the encoded decision explicit, rerunnable, and interpretable. However, the
system still relies on LLMs to parse trial criteria and patient records; missed
evidence, incorrect extraction, or faulty formalization can still lead to
wrong verdicts. Formal reasoning improves interpretability of the encoded policy,
but does not guarantee clinical correctness.

Our evaluation uses synthetic patient notes from the SIGIR benchmark~\citep{koopman2016test}, which
reduces privacy risks because no real patient records are used, but may not
capture the full complexity, incompleteness, and documentation bias of real
electronic health records. Deployment on real patient data would require
institutional review, privacy-preserving data handling, prospective validation,
and monitoring for subgroup-specific errors, especially for patients whose
records are incomplete or whose conditions are documented using nonstandard
terminology.

There are also risks of overreliance. Structured rationales and solver traces
may appear authoritative even when upstream extraction errors are subtle.
Deployed systems should expose supporting evidence, uncertainty flags, and
deferred criteria, and should not be used to deny patients clinician review
without human verification or appeal.
\section*{Acknowledgments}

We acknowledge support from the Verdant Foundation, the Hasso Plattner Institute, Itaú Unibanco, BMO Financial Group, and the Stanford Human-Centered Artificial Intelligence (HAI) Institute. We acknowledge the National Artificial Intelligence Research Resource (NAIRR) Pilot and Microsoft Azure for contributing to the results in this work. Zikai (Cyrus) Zhou is partially supported by the Stanford School of Engineering Fellowship.

We thank Dr. Bryant Lin for suggesting this research topic. We are grateful to Shicheng Liu, Dongwei Jiang, Jiacheng Sang, Harshit Joshi, Jiuding Sun, Yucheng Jiang, Sina Semnani, Tamara Czinczoll, Chris Hoenes, Sally Wang, Jungwoo Kim, and Stanford OVAL Lab members for their support, feedback, and discussions. We also thank Dr. James Ford, Dr. Bryant Lin, Dr. James Dickerson, Dr. Jimmy Lin, Dr. Michael Gensheimer, Professor Gill Bejerano, and Lisa S. Lowy for their expert guidance and helpful discussions. We thank Shengguang Wu, Yicheng Qian, Heng Yu, Juze Zhang, Yue Zhao, Hermann Kumbong, Devon Smith, Youngjoong Kwon, Sa Zhou, Yilong Zhao, Jiaming Tang, Alireza Haqi, Hanchen Li, Fangrui Huang, Haichuan Wang, Jiahao Lu, Ziqi Shu, Shreyas Agarwal, Jie Zhu, Wenyi Wang, Zhijie Huang, Grace Zhang, Jia Shan Zhao, Tianxin Wang, Qinghui Wang, Yunong Zhang, Haochen Pan, Yifei Zhang, and Shiqi Kuang for their support and discussions. We further thank Lefan Zhang, Shan Lu, Blase Ur, Nandish Shah, Ning Tang, Youwen Wu, and Qi Hu for their support and perspectives.

OpenAI ChatGPT and Anthropic Claude were used to polish author-written text and assist with software development, figure design, debugging, and experiment monitoring. The authors reviewed and verified all AI-assisted outputs and remain responsible for the paper's content and results.
\clearpage
\bibliography{smt-trial-matching-arr}

\clearpage
\appendix

\section*{Appendix Roadmap}
\addcontentsline{toc}{section}{Appendix Roadmap}

\begin{table}[ht]
\centering
\small
\setlength{\tabcolsep}{3pt}
\renewcommand{\arraystretch}{1.05}
\begin{tabularx}{\columnwidth}
{@{}c>{\raggedright\arraybackslash}X@{}}
\toprule
Appendix & Contents \\
\midrule
\multicolumn{2}{@{}l}{\emph{Part I --- System and methodology}} \\
\midrule
\ref{app:system_modules}
  & \name{} pipeline: parser, value miner, arbiter, verbalizer, and Z3 program. \\
\ref{app:baselines}
  & Baseline implementations and aggregation rules. \\
\ref{app:prompt_search}
  & Prompt-strategy search for \ourLLM{} and CoT LLM. \\
\ref{app:cf_details}
  & Counterfactual modifier, validator, and clinician audit. \\
\ref{app:example_rationales}
  & Native rationale form per system on two example pairs. \\
\midrule
\multicolumn{2}{@{}l}{\emph{Part II --- Data and reference labels}} \\
\midrule
\ref{app:dataset}
  & Dataset statistics for both benchmarks: patients, trials, pairs, and
    the TREC~2021 sampling frame. \\
\ref{app:descriptive}
  & Decision balance, strata, and audit composition. \\
\ref{app:gold_derivation}
  & Reference-label derivation from the five-judge LLM panel. \\
\ref{app:reference_validation}
  & Clinician validation of reference labels and agreement analysis. \\
\ref{app:qrels_correlation}
  & Comparison of our eligibility labels with SIGIR referral qrels. \\
\midrule
\multicolumn{2}{@{}l}{\emph{Part III --- Detailed results}} \\
\midrule
\ref{app:detailed_results}
  & Cross-backbone accuracy, clinician-adjusted F1, and pairwise pop-recon. \\
\ref{sec:parsing-audit}
  & Parsing audit: meaning, structure, and completeness of formalized criteria. \\
\midrule
\multicolumn{2}{@{}l}{\emph{Part IV --- Ethics and reproducibility}} \\
\midrule
\ref{app:risks}
  & Potential risks: asymmetric error costs, missingness and equity,
    overreliance, and dual use. \\
\ref{app:computational}
  & API-call budget, wall-clock time, and infrastructure. \\
\ref{app:parameters}
  & Package versions, sampling parameters, and seeds. \\
\ref{app:human_subjects}
  & Human-subjects status of the clinician audit. \\
\ref{app:participant_instructions}
  & Instructions given to clinician reviewers. \\
\ref{app:recruitment_payment}
  & Clinician recruitment, payment, and burden. \\
\ref{app:data_consent}
  & Data consent and corpus licensing. \\
\ref{sec:artifact_use}
  & Artifact use, licenses, and intended use. \\
\ref{app:ai_assistance}
  & Use of AI assistants in preparing the paper. \\
\bottomrule
\end{tabularx}
\caption*{Roadmap to the appendix.}
\end{table}

\section{System Modules and Prompts}
\label{app:system_modules}

This appendix documents the modules of \name{} and reproduces the prompts used
to produce the results in the paper. The appendix uses finer-grained module names than
\S\ref{sec:formal-matching}, which presents the pipeline as Steps~1--7.
The trial parser $\TParse$ and the patient parser $\PParse$ together
implement Step~1, producing the trial requirements $\phi_t$ and the resolved
patient representation $p_{t,\Pi}$ respectively. The SMT/\MaxSMT{} solver
$\Solve$ implements Steps~2--6, returning the decision $d$ and decision trace
$\gamma$ (Step~2), the assumptions $\rho$ (Step~4), and the pivotal
conditions $\delta$ (Steps~3, 5, and~6). The LLM verbalizer $\Verbalize$
implements Step~7, producing the clinician-facing rationale $\eta$. LLM-driven stages use \texttt{gpt-4.1} as the backbone unless
otherwise noted; the solver itself is deterministic. Each prompt uses
placeholder tokens, such as \verb|{{PATIENT_NOTES}}| and
\verb|{{VARIABLE_LIST}}|, that are substituted at call time.

\subsection{Pipeline and Prompt Inventory}
\label{app:pipeline_inventory}

For each patient--trial pair, \name{} first compiles the trial criteria into
formal constraints, then extracts patient facts only for the atoms required by
those constraints, evaluates the combined formula with an SMT/\MaxSMT{} solver,
and verbalizes the solver output into a clinician-readable rationale.
Table~\ref{tab:prompt-inventory} lists the pipeline stages and the prompts they
use. Stages that are deterministic use no prompt.

\begin{table}[ht]
\centering
\scriptsize
\setlength{\tabcolsep}{3pt}
\begin{tabularx}{\columnwidth}{@{}cl>{\raggedright\arraybackslash}Xl@{}}
\toprule
Stage & Alg. step & Module & Prompt \\
\midrule
1 & 1   & Leaf collection for trial criteria      & --- \\
2 & 1   & Trial scope mining in $\TParse$         & \S\ref{app:prompt_scope} \\
3 & 1   & Trial projection rewriting in $\TParse$ & \S\ref{app:prompt_projection} \\
4 & 1   & Patient value mining in $\PParse$       &
  \S\S\ref{app:prompt_vm_inc}, \ref{app:prompt_vm_exc} \\
5 & 1   & Alias remapping                         & --- \\
6 & 2--6 & SMT/\MaxSMT{} evaluation with $\Solve$   & --- \\
\midrule
-- & 1 & Extraction-resolution arbiter            & \S\ref{app:prompt_arbiter} \\
-- & 7 & Natural-language verbalizer $\Verbalize$ & \S\ref{app:prompt_verbalizer} \\
\bottomrule
\end{tabularx}
\caption{Pipeline stages and prompts. \emph{Alg. step} maps each stage to
the corresponding algorithm step in \S\ref{sec:formal-matching}.
Stages 1, 5, and 6 are deterministic. The arbiter is invoked only when
extracted patient facts conflict with auxiliary definitional constraints
rather than eligibility requirements.}
\label{tab:prompt-inventory}
\end{table}

\subsection{Trial Parser \TParse{}}
\label{app:tparse}

The trial parser $\TParse$ compiles the natural-language eligibility criteria
of a trial $t$ into the trial constraint formula $C_t=\TParse(t)$. We build on
the \textsc{SatIR} trial-side semantic parser of \citet{zhou2026scalable}, which
performs the underlying criterion-to-SMT compilation and canonicalizes clinical
terms using the SNOMED CT~\cite{snomedct}. In \name{}, we adapt this parser to the
matching setting by adding two patient-independent stages that specialize the
trial atoms for downstream patient-fact extraction. Because these stages depend
only on the trial, their outputs are cached per trial and reused across all
patients matched against that trial.

\paragraph{Scope miner.}
The scope miner classifies each atomic predicate in $\Atom(C_t)$ as either
\emph{in scope}, meaning that it describes a checkable current or historical
clinical state of the patient, or \emph{projected away}, meaning that its main
content belongs to a dimension we do not ask the patient parser to extract
directly, such as evidence strength or the criterion of determination. This
implements the state-projection restriction used by \name{}: $\PParse$ is only
asked about predicates that can be checked against the patient record.

\PromptBlock
{Scope-miner prompt}
{app:prompt_scope}
{https://github.com/verdict1234/verdict_prompts/blob/main/prompts/ours/trial_scope_miner.prompt}
  {prompts/ours/trial_scope_miner.prompt}

\paragraph{Projection rewriter.}
The projection rewriter turns each in-scope predicate into the version consumed
by $\PParse$: a typed signature together with a self-contained,
human-readable definition of what should be checked in the patient chart. This
stage ensures that every predicate passed to the patient parser has an
unambiguous chart-checkable meaning.

\PromptBlock
{Projection-rewriter prompt}
{app:prompt_projection}
{https://github.com/verdict1234/verdict_prompts/blob/main/prompts/ours/trial_projection_rewriter.prompt}
  {prompts/ours/trial_projection_rewriter.prompt}

\subsection{Patient Parser \PParse{}}
\label{app:pparse}

Given a patient record $p$ and the atoms of the trial formula, the patient
parser emits patient-side constraints
$C_p=\PParse(\Atom(C_t),p)$. Unlike systems that extract all possible clinical
facts from a record, $\PParse$ extracts only the facts relevant to the trial
being matched. Each predicate is presented to the LLM with its typed signature
and the self-contained definition produced by the projection rewriter.
Predicates not supported by the record are left unbound.

Patient value mining is run separately for inclusion-side and exclusion-side
predicates, so that each pass can apply side-appropriate handling of missing
evidence. For numerical atoms, the parser considers both the underlying value
and the trial-specific comparison. If the raw value is reported, it is extracted
directly; if the value is absent but the chart contains threshold-level
evidence, the parser may assess the comparison directly under the selected
missing-evidence policy.

\PromptBlock
{Inclusion-side value-miner prompt}
{app:prompt_vm_inc}
{https://github.com/verdict1234/verdict_prompts/blob/main/prompts/ours/patient_value_miner_inclusion.prompt}
  {prompts/ours/patient_value_miner_inclusion.prompt}

\PromptBlock
{Exclusion-side value-miner prompt}
{app:prompt_vm_exc}
{https://github.com/verdict1234/verdict_prompts/blob/main/prompts/ours/patient_value_miner_exclusion.prompt}
  {prompts/ours/patient_value_miner_exclusion.prompt}

\subsection{Solver and Resolution Module \Solve{}}
\label{app:solver}

The solver evaluates the combined patient--trial constraints
$C_{t,p}=C_t \wedge C_p$ after a deterministic alias-remapping pass. It uses
Z3~\citep{de2008z3} to determine satisfiability and its weighted optimization
interface~\citep{bjorner2015nuz} to compute the minimum-cost flip
set $\delta_{t,p}$ when the patient is ineligible. For a trial $t$ and a
patient record $p$, the solver returns
\[
\Solve(C_{t,p}) =
(d_{t,p},\, \gamma_{t,p},\, \rho_{t,p},\, \delta_{t,p},\, L_{t,p}),
\]
where $C_{t,p}=C_t \wedge C_p$ is the conjunction of the trial constraints
$C_t$ and the patient constraints $C_p$ defined above;
$d_{t,p} \in \{\eligible, \ineligible\}$ is the eligibility decision;
$\gamma_{t,p}$ is the decision trace, the logical derivation of $d_{t,p}$ from
$C_{t,p}$; $\rho_{t,p}$ is the set of assumptions, the values the solver
assigns to the conditions that neither patient evidence nor the policies
$\Pi$ resolve; $\delta_{t,p}$ is the set of pivotal conditions, the
minimum-cost set of resolved patient facts and assumptions whose change would reverse
$d_{t,p}$; and $L_{t,p}$ assigns each criterion of $t$ one of the labels
\satisfied{}, \violated{}, or \deferred{}. The artifacts $d_{t,p}$,
$\gamma_{t,p}$, $\rho_{t,p}$, and $\delta_{t,p}$ are the quantities written
$d$, $\gamma$, $\rho$, and $\delta$ in \S\ref{sec:formal-matching}, subscripted
here by the pair $(t,p)$; $L_{t,p}$ is an additional criterion-level summary
emitted by the implementation.

\paragraph{Resolution module.}
The SMT program also contains auxiliary definitional constraints linking
related predicates, such as continuity between current and historical states,
qualifier predicates implying their stems, and aggregate counts defined by
their constituents. When value-miner outputs violate one of these auxiliary
constraints, the resulting inconsistency reflects an extraction conflict rather
than a genuine eligibility conflict. In such cases, the arbiter re-reads the
chart against the conflicting atoms and relevant auxiliary constraints, then
outputs the minimal value changes needed to restore consistency while remaining
faithful to the chart. If the arbiter cannot determine a faithful correction, it
returns no override and the solver result is kept.

\subsection{Model Extraction and Determinism}
\label{app:model_extraction}

The assumptions $\rho_{t,p}$ of Step~4 are read off the \MaxSMT{} optimum
$A$ (\S\ref{sec:formal-matching}), so this subsection records how that
assignment is obtained and in what sense it is reproducible.

\paragraph{Where the values come from.}
Z3 returns an optimum over the atoms it retains after preprocessing. Atoms
that neither the chart nor the policies constrain still receive values, but
those values are a product of the search---decision order, phase saving, the
polarity chosen when a variable is first branched on, and how ties among
equal-cost optima are broken---rather than of the eligibility problem itself.
Three cases are therefore worth separating:

\begin{itemize}[leftmargin=*, noitemsep, topsep=2pt]
  \item \emph{Forced.} The atom occurs in $\phi_t$ in a way that admits only
  one value, as with a conjunctive exclusion. Every model agrees, and the
  entry in $\rho_{t,p}$ is canonical.
  \item \emph{Alternative.} Several completions satisfy $\phi_t$, as with a
  disjunctive requirement. The reported value is the one the search settled
  on; a different assignment would have served equally.
  \item \emph{Inert.} The atom sits in a disjunct already satisfied by
  observed evidence, so no value of it affects the verdict. Simplification
  may remove such an atom before the search begins; because \Solve{}
  evaluates with model completion, it is assigned a value regardless and
  appears in $\rho_{t,p}$ like any other. Inert entries are therefore the
  one component of $\rho_{t,p}$ that carries no decision-relevant content.
\end{itemize}

\paragraph{Reproducibility.}
\name{} calls Z3~4.12.5.0 with default parameters and no \texttt{set\_param}
overrides (\S\ref{app:parameters_z3}), and the alias-remapping pass that
precedes the solver call is deterministic. Re-running the same encoding under
the same solver version therefore reproduces the same model and hence the
same $\rho_{t,p}$. This is reproducibility rather than canonicity: the
forced entries are determined by $\phi_t$, while the alternative and inert
entries reflect the search and may differ under a different solver version
or a different clause ordering. The decision $d_{t,p}$ is unaffected in
either case, since satisfiability does not depend on which model is
returned.

\PromptBlock
{Arbiter prompt}
{app:prompt_arbiter}
{https://github.com/verdict1234/verdict_prompts/blob/main/prompts/ours/arbiter.prompt}
  {prompts/ours/arbiter.prompt}

\subsection{Verbalizer \Verbalize{}}
\label{app:verbalizer}

The verbalizer renders the structured solver output as the natural-language
rationale shown to clinicians. For a trial $t$ and a patient record $p$,
\[
\begin{aligned}
\Verbalize(&\phi_t,\, p_{t,\Pi},\, d_{t,p},\, \gamma_{t,p},\\[-0.2em]
&\rho_{t,p},\, \delta_{t,p},\, L_{t,p})
= (d_{t,p},\, \eta_{t,p}),
\end{aligned}
\]
where $\phi_t$ is the formalized eligibility requirements of $t$ and
$p_{t,\Pi}$ is the patient representation resolved under the policies $\Pi$,
both from \S\ref{sec:formal-representation}; $d_{t,p} \in
\{\eligible, \ineligible\}$ is the eligibility decision, $\gamma_{t,p}$ the
decision trace, $\rho_{t,p}$ the assumptions, $\delta_{t,p}$ the
pivotal conditions, and $L_{t,p}$ the per-criterion labels, all as returned by
$\Solve$ in \S\ref{app:solver}; and $\eta_{t,p}$ is the resulting
clinician-facing rationale in natural language.

The decision $d_{t,p}$ appears on both sides because the verbalizer passes it
through unchanged: it makes no new matching decision. It only describes the
solver-produced decision, the supporting patient facts in $p_{t,\Pi}$, the
criterion statuses in $L_{t,p}$, the assumptions $\rho_{t,p}$ for
\eligible{} decisions, and the pivotal conditions $\delta_{t,p}$ for
\ineligible{} decisions. Thus, the solver output remains the source of truth;
the rationale $\eta_{t,p}$ is only a readable rendering of it.

\PromptBlock
{Verbalizer prompt}
{app:prompt_verbalizer}
{https://github.com/verdict1234/verdict_prompts/blob/main/prompts/ours/verbalizer.prompt}
  {prompts/ours/verbalizer.prompt}
\section{Baseline implementations}
\label{app:baselines}

This appendix specifies the four LLM-based matching baselines used in
\S\ref{sec:systems}: \ourLLM{}, CoT LLM, TrialGPT-Matching, and ZSPM.
For each baseline, we describe the structured output it produces, how we
aggregate that output into a binary patient-level verdict, and the prompt used
to produce the reported results. All baselines use the same patient records,
trial criteria, and backbone LLMs as \name{}.

For baselines with multiple prompting or aggregation variants, we evaluate all
variants on a held-out development split and report only the best-performing
variant in the main results.

\subsection{\ourLLM}
\label{app:baseline_ourllm}

\paragraph{Pipeline.}
\ourLLM{} is a single-prompt LLM matcher. The model receives the patient chart
and the full trial listing, including both inclusion and exclusion criteria
with their original headers preserved. It is instructed to reason through the
patient's clinical state and the trial criteria, then return a binary
patient-level verdict and a short free-text rationale as a JSON object.

\paragraph{Variants tried.}
We evaluated the prompting variants of Appendix~\ref{app:prompt_search} on the development split; the strongest were:
\emph{think-then-aggregate}, \emph{single-step},
\emph{aggregation-first}, and \emph{blocker-then-verdict}. The
\emph{think-then-aggregate} variant performed best and is the one reported in
the main results. In this variant, the model first reasons through the
patient's clinical state and trial requirements before committing to a verdict.

\PromptBlock
{LlmMatch prompt}
{app:prompt_ourllm}
{https://github.com/verdict1234/verdict_prompts/blob/main/prompts/baselines/single-shot/two_step.prompt}
  {prompts/baselines/single-shot/two_step.prompt}

\subsection{CoT LLM}
\label{app:baseline_cot}

\paragraph{Pipeline.}
CoT LLM uses the same inputs as \ourLLM{} but requires an explicit
criterion-by-criterion analysis before producing the final verdict. For each
inclusion criterion, the model emits a status from
\{\textit{included}, \textit{not included}, \textit{unknown}\} and a short
reason. For each exclusion criterion, it emits a status from
\{\textit{excluded}, \textit{not excluded}, \textit{unknown}\} and a short
reason. The model then aggregates these per-criterion judgments into a binary
patient-level verdict and an aggregation rationale.

\paragraph{Variants tried.}
We evaluated three CoT variants on the development split:
(i) \texttt{V5\_COT}, which uses the shared label set
\{\textit{met}, \textit{not met}, \textit{unknown}\} for both inclusion and
exclusion criteria; (ii) \texttt{V5\_VERBOSE\_AGG}, which adds an explicit
aggregation step but retains the shared \textit{met}/\textit{not met} labels;
and (iii) \texttt{V5\_VERBOSE\_AGG\_V2}, which uses side-specific labels:
\textit{included}/\textit{not included} for inclusion criteria and
\textit{excluded}/\textit{not excluded} for exclusion criteria.
\texttt{V5\_VERBOSE\_AGG\_V2} performed best and is the variant reported in
the main results.

\PromptBlock
{CoT LLM prompt}
{app:prompt_cot_llm}
{https://github.com/verdict1234/verdict_prompts/blob/main/prompts/baselines/cot-llm/cot.prompt}
  {prompts/baselines/cot-llm/cot.prompt}

\subsection{TrialGPT-Matching}
\label{app:baseline_trialgpt}

\paragraph{Pipeline.}
TrialGPT-Matching~\citep{jin2024matching} classifies eligibility at the
criterion level using separate prompts for inclusion and exclusion criteria.
Each criterion is assigned one of five labels:
\{\textit{included}, \textit{not included}, \textit{not applicable},
\textit{excluded}, \textit{not excluded}\}, together with a short
per-criterion rationale. The original TrialGPT matching component does not
directly output a binary patient-level verdict.

\paragraph{Aggregation.}
To compare TrialGPT-Matching with the other matchers, we add a deterministic
binary aggregation rule on top of its per-criterion outputs. We evaluated four
aggregation rules on the development split: (i) all-inclusions-met, which marks
a pair eligible iff every inclusion criterion is labeled \textit{included};
(ii) majority-of-criteria; (iii) the published TrialGPT scoring heuristic with
a learned cutoff; and (iv) a permissive rule that marks a pair eligible iff no
inclusion criterion is labeled \textit{not included} and no exclusion criterion
is labeled \textit{excluded}. Rule (iv) performed best and is the aggregation
rule reported in the main results.

\PromptBlock
{TrialGPT-Matching inclusion prompt}
{app:prompt_trialgpt_inclusion}
{https://github.com/verdict1234/verdict_prompts/blob/main/prompts/baselines/trialgpt/trialgpt_inclusion.prompt}
  {prompts/baselines/trialgpt/trialgpt_inclusion.prompt}

\PromptBlock
{TrialGPT-Matching exclusion prompt}
{app:prompt_trialgpt_exclusion}
{https://github.com/verdict1234/verdict_prompts/blob/main/prompts/baselines/trialgpt/trialgpt_exclusion.prompt}
  {prompts/baselines/trialgpt/trialgpt_exclusion.prompt}

\subsection{ZSPM}
\label{app:baseline_zspm}

\paragraph{Pipeline.}
ZSPM~\citep{wornow2025zero} is the zero-shot patient matching baseline of
Wornow et al. The model receives the patient chart and the trial's inclusion
and exclusion criteria, parsed line by line. For each criterion, it emits an
\texttt{is\_met} judgment, a confidence label
(\textit{high}, \textit{medium}, or \textit{low}), and a short rationale. It
then aggregates the criterion-level judgments into a patient-level verdict.

We use two prompt variants of ZSPM, one for the main matching evaluation and
one for the counterfactual self-faithfulness analysis. The variants share the
same chart and criterion inputs but differ in the verdict and evidence-status
schema they ask the model to produce.

\paragraph{Matching-evaluation variant.}
For the main matching metrics in \S\ref{sec:experimental_setup}, we use the original
ternary-verdict prompt from Wornow et al., which emits
\texttt{global\_decision} $\in \{0,1,2\}$:
0 = \textit{not eligible}, 1 = \textit{might be eligible}, and
2 = \textit{eligible}. Because our reference evaluation is binary, we evaluated
both possible mappings of the middle bucket on the development split:
(i) map \textit{might be eligible} to \textsc{eligible}
($\texttt{global\_decision} \geq 1$), and (ii) map it to
\textsc{ineligible} ($\texttt{global\_decision}=2$ only). Mapping the middle
bucket to \textsc{eligible} performed better and is the mapping reported in the
main results.

\PromptBlock
{ZSPM matching-evaluation prompt}
{app:prompt_zspm_accuracy}
{https://github.com/verdict1234/verdict_prompts/blob/main/prompts/baselines/zspm/zspm.prompt}
  {prompts/baselines/zspm/zspm.prompt}

\paragraph{Counterfactual-analysis variant.}
For the counterfactual self-faithfulness analysis in \S\ref{sec:eval-counterfactual}, the
ternary verdict and binary \texttt{is\_met} flag are insufficient: we need to
distinguish criteria contradicted by chart evidence from criteria that are
unmet only because the chart is silent. We therefore rerun ZSPM with a
modified prompt that forces a binary patient-level decision
(\texttt{eligible: bool}) and assigns each criterion an
\texttt{evidence\_status} from
\{\texttt{deterministic\_met}, \texttt{deterministic\_not\_met},
\texttt{defer}\}. The first two labels mark criteria whose status is supported
by explicit chart evidence; \texttt{defer} marks criteria for which the chart
does not provide enough evidence to determine the criterion. This distinction
is needed for the minimum-flip-set analysis, because deferred criteria should
not be counted as evidence-grounded blockers. We rerun the matcher rather than
post-hoc relabeling the matching-evaluation output, because the original prompt
does not ask the model to expose this distinction.

\PromptBlock
{ZSPM{} counterfactual-analysis prompt}
{app:prompt_zspm_cf}
{https://github.com/verdict1234/verdict_prompts/blob/main/prompts/baselines/zspm/zspm_binary.prompt}
  {prompts/baselines/zspm/zspm_binary.prompt}
\section{Prompting Strategy Search for the \ourLLM{} Baseline}
\label{app:prompt_search}

The \ourLLM{} baseline reported in the main paper is the
strongest configuration recovered by an explicit search over prompting
strategies for a single-shot eligibility verdict. We did not pick a single
``obvious'' prompt and call it the baseline; instead, we wrote six variants
that exercise three orthogonal design axes and selected the one with the
highest dev-set macro-F1.

\subsection{Search axes}
\label{app:prompt_search_axes}

The six prompts cover the following three axes, which together span the
prompting choices most commonly described as ``state of the art'' in the
clinical-trial matching literature:

\begin{itemize}
  \item \textbf{Reasoning structure.} Single-pass verdict vs.\
        think-then-aggregate (the model first reasons criterion-by-criterion,
        then emits a global verdict) vs.\ aggregate-first verbose
        chain-of-thought.
  \item \textbf{Vocabulary specificity.} Generic include/exclude labels vs.\
        side-specific labels (\textsc{included}/\textsc{not-included} for
        inclusion criteria, \textsc{excluded}/\textsc{not-excluded} for
        exclusion criteria), and an explicit blocker channel.
  \item \textbf{Verbosity.} Minimal-rationale vs.\ verbose per-criterion
        rationales.
\end{itemize}

\subsection{Selection protocol}
\label{app:prompt_search_protocol}

Each variant was scored on the dev split using the same scoring harness as
the main experiments (gpt-4.1 backbone, identical chart and trial text,
identical post-processing). The search produces two separately reported
baselines:

\begin{itemize}
  \item \ourLLM{} --- the strongest variant over the full
        six-variant grid. The winner is the think-then-aggregate prompt
        (\Cref{app:prompt_two_step}).
  \item CoT LLM --- the strongest variant restricted to
        chain-of-thought-shape prompts (single-pass CoT vs.\
        aggregation-first verbose CoT, with generic vs.\ side-specific
        labels for the aggregation-first variant). The winner is the
        aggregation-first verbose CoT with side-specific labels
        (\Cref{app:prompt_verbose_v2}); the generic-label version
        (\Cref{app:prompt_verbose}) and the single-pass CoT
        (\Cref{app:prompt_cot}) both underperformed it on the dev split.
\end{itemize}

We report both baselines so that CoT LLM reflects a fair
chain-of-thought prompting effort rather than a single hand-written CoT
prompt. Variants are listed below in roughly descending dev-set strength.

\subsection{Variants searched}
\label{app:prompt_search_variants}

Two-pass prompt: the model first emits a structured per-criterion
assessment (criterion span, side, met/unmet/uncertain, brief justification),
then aggregates those into a single eligibility verdict. This is the prompt
used as \ourLLM{} in all main-paper tables.

\PromptBlock
  {Think-then-aggregate (selected)}
  {app:prompt_two_step}
  {https://github.com/verdict1234/verdict_prompts/blob/main/prompts/baselines/prompt-search/two_step.prompt}
  {prompts/baselines/prompt-search/two_step.prompt}

Variant of think-then-aggregate that adds an explicit \emph{blocker}
channel: the model is asked to first surface any criterion that, if unmet,
would single-handedly make the patient ineligible, and to commit to a
verdict consistent with that channel.

\PromptBlock
  {Blocker-then-verdict}
  {app:prompt_blockers}
  {https://github.com/verdict1234/verdict_prompts/blob/main/prompts/baselines/prompt-search/V5_TWO_STEP_BLOCKERS.prompt}
  {prompts/baselines/prompt-search/V5_TWO_STEP_BLOCKERS.prompt}

Aggregation-first variant: the model emits a verbose chain-of-thought over
all criteria and then commits to a verdict, using side-specific labels
(\textsc{included}/\textsc{not-included} vs.\
\textsc{excluded}/\textsc{not-excluded}) rather than generic include/exclude.
This is also the variant used to source the side-by-side rationales for the
CoT LLM system in the pairwise-preference table.

\PromptBlock
  {Aggregation-first verbose CoT (side-specific labels)}
  {app:prompt_verbose_v2}
  {https://github.com/verdict1234/verdict_prompts/blob/main/prompts/baselines/prompt-search/V5_VERBOSE_AGG_V2.prompt}
  {prompts/baselines/prompt-search/V5_VERBOSE_AGG_V2.prompt}

Earlier version of the aggregation-first verbose CoT prompt using generic
include/exclude vocabulary. Underperformed the side-specific variant on the
dev split.

\PromptBlock
  {Aggregation-first verbose CoT (generic labels)}
  {app:prompt_verbose}
  {https://github.com/verdict1234/verdict_prompts/blob/main/prompts/baselines/prompt-search/V5_VERBOSE_AGG.prompt}
  {prompts/baselines/prompt-search/V5_VERBOSE_AGG.prompt}

Single-pass CoT: the model is asked to think step-by-step about the chart
and the criteria in a single message and emit a verdict at the end, without
the explicit per-criterion table required by the think-then-aggregate
variant.

\PromptBlock
  {Chain-of-thought, single-pass verdict}
  {app:prompt_cot}
  {https://github.com/verdict1234/verdict_prompts/blob/main/prompts/baselines/prompt-search/V5_COT.prompt}
  {prompts/baselines/prompt-search/V5_COT.prompt}

Minimal baseline: chart plus criteria, verdict only, no per-criterion
reasoning and no enforced think-step. Included to confirm that the
prompting-strategy gains we report are not driven entirely by the model's
default behavior.

\PromptBlock
  {Plain single-step verdict}
  {app:prompt_plain}
  {https://github.com/verdict1234/verdict_prompts/blob/main/prompts/baselines/prompt-search/V5_PLAIN.prompt}
  {prompts/baselines/prompt-search/V5_PLAIN.prompt}

\subsection{Takeaway}
\label{app:prompt_search_takeaway}

Across the six variants, the think-then-aggregate prompt was the strongest
on the dev split, and we report it as \ourLLM{} in the main
paper; the strongest of the three chain-of-thought-shape prompts was the
aggregation-first verbose CoT with side-specific labels, which we report as
CoT LLM. This framing matters because a casual ``we wrote a single
prompt and called it the LLM baseline'' --- or a single hand-written CoT
prompt for CoT LLM --- would understate how much of the headroom
between off-the-shelf LLM matching and our system can be closed by prompt
engineering alone, and how much remains after that headroom is taken.
\section{Counterfactual Self-Faithfulness: Pipeline Details}
\label{app:cf_details}

This appendix specifies the counterfactual self-faithfulness pipeline
of \S\ref{sec:eval-counterfactual}: how flip targets are extracted from each system,
how counterfactual charts are constructed and validated, the three
modifier--validator configurations we report, how confidence intervals
are computed, and how clinician audit data is used to produce the
clinician-adjusted variant of the metric. We embed all CF-side prompts
verbatim in \S\ref{app:cf_prompts}.

\subsection{Per-pair workflow}
\label{app:cf_workflow}

For every patient--trial pair $(p,t)$ on which a matcher emits the
verdict \textsc{ineligible}, the pipeline runs the following five
steps.

\begin{enumerate}\itemsep 2pt
    \item \textbf{Extract flip targets.} For each matcher, the
    rejection reasons are extracted in the form most natural to the
    matcher (\S\ref{app:cf_blockers}). The result is a list of
    \emph{flip targets} --- either free-text criterion-and-fact pairs
    for LLM matchers, explicitly listed unmet requirements for
    \ourLLM{-Pivotal}, or symbolic atoms with target values for
    the SMT-based matcher.
    \item \textbf{Construct a counterfactual chart.} A modifier LLM is
    given the original chart, the trial, and the list of flip targets,
    and is instructed to make the minimal edits that address those
    targets while preserving everything else. The modifier is told
    explicitly not to introduce new conditions, treatments, or chart
    facts beyond what is needed to remove the cited rejection reasons.
    \item \textbf{Validate the counterfactual.} A separate validator
    LLM checks that the modified chart (i)~addresses the targeted
    blockers, (ii)~does not alter unrelated eligibility-relevant
    facts, (iii)~does not introduce new blockers, and (iv)~remains
    clinically coherent. Pairs whose modification fails any of these
    checks are excluded from the denominator of $\mathrm{SF}_{\mathrm{raw}}$.
    \item \textbf{Re-run the matcher.} The same matcher, with the same
    backbone and prompt, is run on the validated counterfactual chart.
    We record the new verdict $\hat{y}'(p)$.
    \item \textbf{Score.} The pair contributes to
    $\mathrm{SF}_{\mathrm{raw}}$ as a positive if the new verdict is
    \textsc{eligible}, and as a negative otherwise.
\end{enumerate}

The modifier and validator are run \emph{out of sample} with respect
to the matcher under test: they never see the matcher's verdict on
the original chart, so the modification is not biased toward
``producing a flip.''

\subsection{Per-system flip-target extraction}
\label{app:cf_blockers}

The matchers in our evaluation expose their rejection reasons in
different forms, so the flip-target extraction step is tailored to
each.

\paragraph{Single-shot, CoT, and TrialGPT LLM matchers.}
These matchers emit a natural-language rationale that names the
criteria that drove the rejection. We parse cited flip targets from
the rationale text with a side-LLM call using the prompt in
\S\ref{app:prompt_modifier_llm}. The output is a list of
\texttt{\{criterion, chart\_fact\}} pairs, where \texttt{criterion}
quotes the trial-side requirement the matcher cited and
\texttt{chart\_fact} quotes the chart evidence the matcher used.
These are the targets the modifier is asked to address.

\paragraph{\ourLLM{-Pivotal}.}
\ourLLM{-Pivotal} uses the same \ourLLM{} matcher, but asks the
model to explicitly list the unmet requirements whose change would
make the patient eligible. These listed flip targets are used directly
for counterfactual construction, rather than being recovered from a
post-hoc rationale parsing step.

\paragraph{ZSPM.}
ZSPM emits per-criterion \texttt{is\_met}/\texttt{not\_met}
assessments rather than a free-text rationale. We extract a
\emph{minimum flip set} from these assessments with the prompt in
\S\ref{app:prompt_shah_minflip}: the smallest set of per-criterion
verdicts that would need to flip for the global decision to change
from \textsc{ineligible} to \textsc{eligible}. This puts ZSPM's
flip targets on the same footing as the other matchers for the
modification step.

\paragraph{\name{}, SMT-based matcher.}
Our matcher's rejection reasons are already symbolic: the \MaxSMT{}
minimum-cost flip set $\delta_{t,p}$ (\S\ref{sec:formal-matching})
names the atomic patient predicates whose values must change to make
the program satisfiable. We extract these atoms with their target
values using the prompt in \S\ref{app:prompt_modifier_aegis} and pass
them to the modifier as a list of
\texttt{\{atom, current\_value, target\_value\}} tuples. Because
$\delta_{t,p}$ is minimal by \MaxSMT{}, the modification target is
unambiguous.

\subsection{Modifier--validator configurations}
\label{app:cf_configs}

We report results under the three modifier--validator configurations of
Table~\ref{tab:cf-configs}, which swap backbones to test the robustness of the
metric:

\begin{table}[ht]
\centering
\small
\setlength{\tabcolsep}{4pt}
\begin{tabularx}{\linewidth}{@{}llX@{}}
\toprule
Modifier & Validator & Purpose \\
\midrule
gpt-4.1 & gpt-4.1 strict
  & Baseline same-backbone setting \\
gpt-4.1 & gpt-5 simclin
  & Tests sensitivity to the validator \\
gpt-5 & gpt-5 simclin
  & Headline strongest modifier/validator setting \\
\bottomrule
\end{tabularx}
\caption{Modifier--validator configurations for counterfactual
self-faithfulness. The gpt-5 modifier / gpt-5 simclin validator
configuration is the headline setting; the other two configurations
test sensitivity to the modifier and validator choices.}
\label{tab:cf-configs}
\end{table}

The two validator prompts differ in framing. The gpt-4.1
\emph{strict} validator (\S\ref{app:prompt_validator_strict}) is a
checklist-style prompt that asks whether each modification leaves the
chart clinically coherent. The gpt-5 \emph{simclin} validator
(\S\ref{app:prompt_validator_simclin}) is a senior trial-coordinator
persona that re-screens the modified chart from scratch. Comparing
gpt-4.1 modifier / gpt-4.1 validator with gpt-4.1 modifier / gpt-5
validator isolates the effect of the validator; comparing gpt-4.1
modifier / gpt-5 validator with gpt-5 modifier / gpt-5 validator
isolates the effect of the modifier.

\subsection{Confidence intervals}
\label{app:cf_ci}

Because $\mathrm{SF}_{\mathrm{raw}}$ is a binomial proportion over
validator-accepted counterfactuals, we report Wilson 95\% confidence
intervals
\citep{wilson1927probable}:
\[
\mathrm{CI}_{95}
= \frac{\hat{p} + \frac{z^2}{2n}}{1 + \frac{z^2}{n}}
  \pm \frac{z}{1+\tfrac{z^2}{n}}
      \sqrt{\frac{\hat{p}(1-\hat{p})}{n} + \frac{z^2}{4n^2}},
\]
where $\hat{p}=\mathrm{SF}_{\mathrm{raw}}$ is the observed flip rate, $n$ is
the number of validator-accepted counterfactuals over which that rate is
computed for the system and configuration in question, and $z=1.96$ is the
standard normal quantile for a two-sided 95\% interval. Wilson intervals are
preferred over normal-approximation intervals at small $n$ and near-boundary
$\hat{p}$, both of which can occur for individual system/configuration
pairs.

\subsection{Clinician-adjusted self-faithfulness}
\label{app:cf_adjusted}

The validator is itself an LLM and can make two kinds of error: it
can accept counterfactuals that do not in fact address the cited flip
targets, or it can reject counterfactuals that actually do. The first
kind can inflate $\mathrm{SF}_{\mathrm{raw}}$ on systems whose cited
reasons are easy to remove cosmetically; the second kind can deflate
the denominator. To assess this effect, we run a stratified clinician
audit on the gpt-5 modifier / gpt-5 validator configuration.

\paragraph{Audit design.}
For each system, we stratify the set of validator-accepted
counterfactuals into two buckets: \emph{flipped} (the matcher's
re-run verdict became \textsc{eligible}) and \emph{not flipped} (the
verdict remained \textsc{ineligible}). We target $K=7$ counterfactuals
from each bucket, yielding 7--10 per bucket in practice, and present each to a senior trial-coordinator
clinician who marks the modification as clinically valid or not,
blind to the matcher's identity and the bucket label. The audit
produces, per system, two precision estimates:
$p_{\mathrm{flip}}$ = clinician-valid rate on the flipped bucket, and
$p_{\mathrm{not}}$ = clinician-valid rate on the not-flipped bucket.

\paragraph{Re-weighting.}
Let $N_F$ and $N_{NF}$ be the full-corpus counts of validator-accepted
flipped and not-flipped counterfactuals. The clinician-adjusted SF
re-weights the bucket-conditional clinician-valid rates back to the
full corpus:
\[
\mathrm{SF}_{\mathrm{adj}}
=
\frac{p_{\mathrm{flip}} \, N_F}
     {p_{\mathrm{flip}} \, N_F + p_{\mathrm{not}} \, N_{NF}}.
\]
This is the corpus-level fraction of \emph{clinician-valid}
counterfactuals on which the matcher's verdict flipped. The 95\%
confidence interval is computed by stratified pair-level bootstrap
(2{,}000 resamples) within each bucket, propagating uncertainty from
both the bucket-conditional rates and the bucket sizes.

\subsection{CF-pipeline prompts}
\label{app:cf_prompts}

\PromptBlock
  {Modifier prompt (LLM matchers)}
  {app:prompt_modifier_llm}
  {https://github.com/verdict1234/verdict_prompts/blob/main/prompts/cf/modifier_llm.prompt}
  {prompts/cf/modifier_llm.prompt}

\PromptBlock
  {Modifier prompt (\name{}, atom targets)}
  {app:prompt_modifier_aegis}
  {https://github.com/verdict1234/verdict_prompts/blob/main/prompts/cf/modifier_ours.prompt}
  {prompts/cf/modifier_ours.prompt}

\PromptBlock
  {ZSPM min-flip extractor}
  {app:prompt_shah_minflip}
  {https://github.com/verdict1234/verdict_prompts/blob/main/prompts/cf/shah_minflip.prompt}
  {prompts/cf/shah_minflip.prompt}

\PromptBlock
  {Strict validator (gpt-4.1)}
  {app:prompt_validator_strict}
  {https://github.com/verdict1234/verdict_prompts/blob/main/prompts/cf/validator_strict.prompt}
  {prompts/cf/validator_strict.prompt}

\PromptBlock
  {Simclin validator (gpt-5)}
  {app:prompt_validator_simclin}
  {https://github.com/verdict1234/verdict_prompts/blob/main/prompts/cf/validator_simclin.prompt}
  {prompts/cf/validator_simclin.prompt}

\subsection{User Interface for Clinician Counterfactual Audit}
\label{app:user_interface_for_clinician_cf_audit}

\begin{figure*}[tp]
  \centering
  \includegraphics[width=\linewidth]{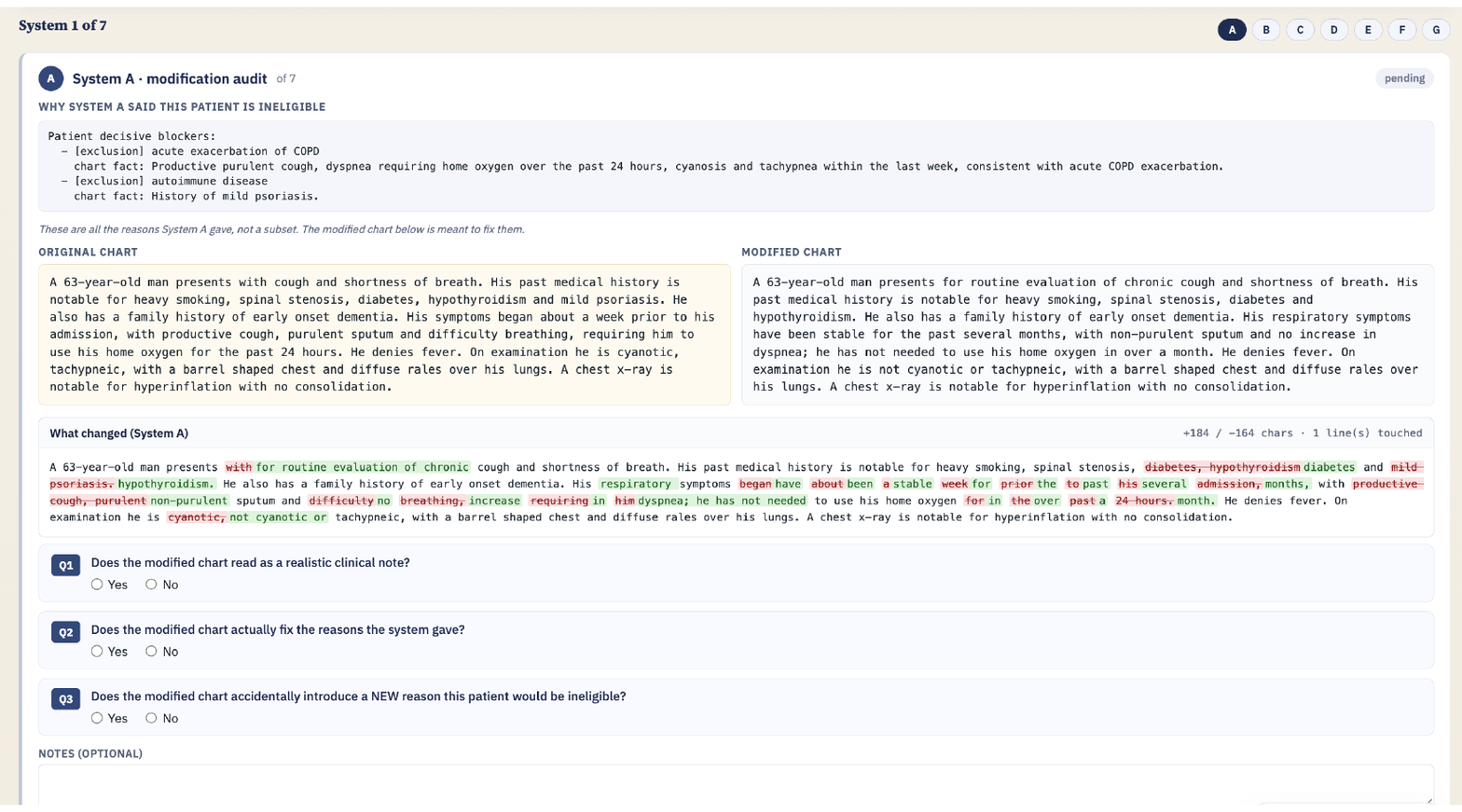}
  \caption{Annotation interface for the counterfactual clinical-validity
  audit.}
  \label{fig:clinician_counterfactual}
\end{figure*}

Figure~\ref{fig:clinician_counterfactual} shows a screenshot of the annotation
interface presented to the clinician during the
counterfactual audit described in Appendix~\ref{app:cf_adjusted}. The interface
implements a task type: Counterfactual clinical-validity judgment.

\paragraph{Counterfactual clinical-validity judgment
(Figure~\ref{fig:clinician_counterfactual}).}
For each sampled counterfactual, the clinician is shown the original patient
chart, the trial eligibility criteria, the modified chart with edits
highlighted, and the flip targets the modifier was instructed to address. The
clinician then marks the modification as \textbf{CLINICALLY VALID} --- the
edits address the cited blockers, do not introduce new ineligibility grounds,
and leave the chart clinically coherent --- or \textbf{NOT CLINICALLY VALID}
if one or more of these conditions fails, and provides a one-line written
reason for the decision.

\clearpage
\section{Example Rationales}
\label{app:example_rationales}

This appendix illustrates what each system's rationale looks like on
two patient--trial pairs, using each system's native output form ---
\name{}'s atom-grounded bullet structure, \ourLLM{}'s free-form
aggregation prose, ZSPM's per-criterion met / not-met list, CoT LLM's
strict-JSON per-criterion schema, and TrialGPT's two-pass criterion-
index JSON. All systems run on \texttt{gpt-4.1} with
\texttt{temperature}\,$=$\,$0$. Verdicts are each system's own; the
reference label is shown in each pair header.

\subsection{Pair A}
\label{app:ex_pair_a}

\noindent\textbf{Pair:} \texttt{sigir-20158\_\_NCT00393913}.
A 10-year-old boy with classic obstructive-sleep-apnea symptoms
(nighttime snoring, witnessed apneas, excessive daytime sleepiness).
\emph{Reference label: eligible.}

\paragraph{\name{} (verdict: eligible).}
\begin{quote}\small
A 10-year-old boy with prominent snoring, daytime sleepiness, and stable
medical history satisfies the trial's core inclusion criteria for
obstructive sleep apnea (OSA); no exclusion is triggered. Most exclusion
and some inclusion criteria are not addressed in the chart and will
require confirmation at the in-person visit.

The clinical picture is consistent with pediatric OSA: the patient
experiences snoring, excessive daytime sleepiness, and has a stable
medical history without documented medication changes. The trial's
inclusion criteria are directly met by these findings, and there is no
evidence of exclusionary conditions or comorbidities in the chart.

\textbf{Verdict:} Eligible

\textbf{Inclusions met:}
\begin{itemize}\setlength\itemsep{0pt}
  \item Inclusion 1 --- met: symptoms of OSA, including snoring and sleepiness (chart: ``nighttime snoring,'' ``excessive sleepiness during class'').
  \item Inclusion 2 --- met: stable medical history with no change in medications affecting sleepiness.
\end{itemize}

\textbf{No exclusions apply:}
\begin{itemize}\setlength\itemsep{0pt}
  \item Exclusion 1 --- not applicable: no suspected diagnosis of a sleep disorder other than OSA.
  \item Exclusion 5 --- not applicable: not pregnant (chart: ``10 yo boy'').
\end{itemize}

\textbf{To confirm at the in-person visit:}
\begin{itemize}\setlength\itemsep{0pt}
  \item \textbf{Medication history:} current medication use, change in medications affecting sleepiness, psychotropic medication use.
  \item \textbf{Comorbidities:} medically unstable conditions, recreational drug use, alcohol abuse history, cognitive impairment.
  \item \textbf{Communication and sensory:} inability to communicate verbally, visual or hearing impairment.
\end{itemize}
\end{quote}

\paragraph{\ourLLM{} (verdict: ineligible).}
\begin{quote}\small
The matcher determined that the patient meets the inclusion criterion of
experiencing symptoms of obstructive sleep apnea, as evidenced by
documented nighttime snoring, observed pauses in breathing, and reports
of excessive sleepiness during class. However, the matcher found that
the patient's stability of medical history and any recent changes in
medications that could affect sleepiness were not addressed in the
chart. On the exclusion side, the matcher noted that there was no
information in the chart regarding suspected diagnoses of other sleep
disorders, medically unstable health conditions, recent use of
psychotropic medications, history of recreational drug or alcohol use,
inability to communicate, or the presence of visual, hearing, or
cognitive impairment. The criterion regarding pregnancy was explicitly
addressed, with the matcher concluding that the patient, being a
10-year-old boy, is ineligible on this basis, as pregnancy is not
applicable. Ultimately, the matcher judged the patient ineligible for
the trial, citing the patient's age --- since the chart documents he is
10 years old and the trial does not specify inclusion of pediatric
patients --- as the decisive factor for ineligibility.
\end{quote}

\paragraph{CoT LLM (verdict: eligible).}
\begin{quote}\small
\{"inclusion\_criteria": [\{"text": "Experiences symptoms of OSA, including snoring and sleepiness", "status": "included", "reasoning": "The patient has nighttime snoring, pauses in breathing, and excessive sleepiness during class, which are symptoms of OSA."\}, \{"text": "Stable medical history with no change in medications that could affect sleepiness", "status": "unknown", "reasoning": "The chart does not mention any medical history or medication changes; this information is not provided."\}], "exclusion\_criteria": [\{"text": "Suspected diagnosis of a sleep disorder other than OSA (i.e., periodic leg movements, narcolepsy, insomnia, central sleep apnea, sleep hypoventilation syndrome)", "status": "not\_excluded", "reasoning": "There is no mention of other sleep disorders; symptoms described are consistent with OSA only."\}, \{"text": "Medically unstable health conditions (e.g., heart attack, congestive heart failure)", "status": "not\_excluded", "reasoning": "No evidence of medically unstable health conditions is provided in the chart."\}, \{"text": "Use of psychotropic medications that cause sedation in the 3 months prior to study entry", "status": "unknown", "reasoning": "The chart does not mention any medication use, including psychotropic medications."\}, \{"text": "Recent or confirmed history of recreational drug use or alcohol abuse", "status": "not\_excluded", "reasoning": "There is no mention of recreational drug use or alcohol abuse in the chart."\}, \{"text": "Pregnant", "status": "not\_excluded", "reasoning": "The patient is a 10-year-old boy, so pregnancy does not apply."\}, \{"text": "Inability to communicate verbally, write, or read", "status": "not\_excluded", "reasoning": "There is no evidence of communication impairment; the patient attends school and interacts with teachers."\}, \{"text": "Visual, hearing, or cognitive impairment", "status": "not\_excluded", "reasoning": "No visual, hearing, or cognitive impairment is mentioned; the patient attends school and is described as having declining grades, not impairment."\}], "aggregation\_reasoning": "Inclusion 1: included. Inclusion 2: unknown --- applied prescreen-doctrine, treated as included since it is not population-defining. Exclusion 1: not\_excluded. Exclusion 2: not\_excluded. Exclusion 3: unknown --- applied prescreen-doctrine, treated as not\_excluded since it is not population-defining. Exclusion 4: not\_excluded. Exclusion 5: not\_excluded. Exclusion 6: not\_excluded. Exclusion 7: not\_excluded. Therefore: ELIGIBLE.", "eligibility": "eligible"\}
\end{quote}

\paragraph{TrialGPT (no patient-level verdict; two passes).}
\begin{quote}\small
\emph{Inclusion pass:} \{"0": ["The patient experiences snoring, pauses in breathing, restlessness, and excessive sleepiness during class, which are symptoms of OSA.", [0, 2], "included"], "1": ["There is no information about the patient's medical history stability or medication changes.", [], "not enough information"]\}

\emph{Exclusion pass:} \{"0": ["There is no mention of a suspected diagnosis of a sleep disorder other than OSA; symptoms described are consistent with OSA.", [0], "not excluded"], "1": ["No evidence of medically unstable health conditions is provided.", [], "not enough information"], "2": ["No mention of psychotropic medication use.", [], "not enough information"], "3": ["No mention of recreational drug use or alcohol abuse.", [], "not enough information"], "4": ["The patient is a 10-year-old boy, so pregnancy is not applicable.", [], "not applicable"], "5": ["No mention of inability to communicate verbally, write, or read.", [], "not enough information"], "6": ["No mention of visual, hearing, or cognitive impairment.", [], "not enough information"]\}
\end{quote}

\subsection{Pair B}
\label{app:ex_pair_b}

\noindent\textbf{Pair:} \texttt{sigir-20153\_\_NCT02596555}.
A 65-year-old man with acute pulmonary embolism following hip-replacement
surgery; the trial enrolls intermediate-risk PE patients. \emph{Reference
label: eligible.}

\paragraph{\name{} (verdict: eligible).}
\begin{quote}\small
A 65-year-old man meets the trial's minimum age requirement of 18 years,
which is decisive; no exclusion is triggered. All other inclusion and
exclusion criteria are not documented in the chart and will require
confirmation at the in-person visit.

The chart picture is of a man presenting with acute onset shortness of
breath and chest pain following recent orthopedic surgery, consistent
with possible acute pulmonary embolism, but only age is confirmed. The
trial's diagnostic, risk stratification, and safety criteria are not
addressed in the chart and must be verified at screening.

\textbf{Verdict:} Eligible

\textbf{Inclusions met:}
\begin{itemize}\setlength\itemsep{0pt}
  \item Inclusion 1 --- met: age $\geq 18$ years (chart documents ``65 yo male'').
\end{itemize}

\textbf{No exclusions apply:}
\begin{itemize}\setlength\itemsep{0pt}
  \item Exclusion 1 --- not applicable: not pregnant or lactating (chart: male sex).
  \item Exclusion 2 --- not applicable: not a woman of childbearing potential (chart: male sex).
\end{itemize}

\textbf{To confirm at the in-person visit:} (12 items, condensed: imaging-based PE confirmation, hemodynamics, sPESI / NT-proBNP / troponin, procedural history, bleeding risk, comorbidities, hypersensitivity, anticoagulant washout, follow-up logistics).
\end{quote}

\paragraph{ZSPM (verdict: ineligible).}
\begin{quote}\small
\textbf{Inclusion criteria (per-criterion verdict):}
\begin{itemize}\setlength\itemsep{0pt}
  \item [met] The patient is 65 years old, which is $\geq 18$ years.
  \item [not met] There is no mention of an objectively confirmed diagnosis of acute PE by imaging. Symptoms are suggestive, but confirmation is not documented.
  \item [met] The patient presents with tachypnea and shortness of breath but there is no mention of hemodynamic collapse.
  \item [not met] sPESI score is not explicitly calculated\dots\ at least one point for age $>$65. However, the presence of additional required criteria is not documented.
  \item [not met] No imaging (CT or echo) findings of RV pressure overload/dysfunction are provided.
  \item [not met] No troponin levels or mention of myocardial injury are provided.
  \item [not met] No NT-proBNP levels or mention of RV failure are provided.
  \item [not met] No information is provided about the patient's ability to consent or understanding of the trial.
\end{itemize}

\textbf{Exclusion criteria (per-criterion verdict):}
\begin{itemize}\setlength\itemsep{0pt}
  \item [met] The patient is a 65-year-old male, so pregnancy and lactation do not apply.
  \item [met] (\dots\ seven additional [met] entries for criteria not contradicted by the chart.)
\end{itemize}
\end{quote}

\section{Dataset Statistics}
\label{app:dataset}

Our evaluation set comprises 552 patient--trial pairs built from the
SIGIR~2016 clinical-trial cohort: 59 patient vignettes, each paired
with its top-10 TrialGPT-Retrieval trial candidates (590 pairs), of
which 552 obtain complete five-judge gold coverage
(Appendix~\ref{app:gold_derivation}). Gold labels are near-balanced
(278 \textsc{eligible} / 274 \textsc{ineligible}).
Table~\ref{tab:dataset-stats} reports the full statistics.

\begin{figure}[htbp]
\centering
\includegraphics[width=\linewidth]{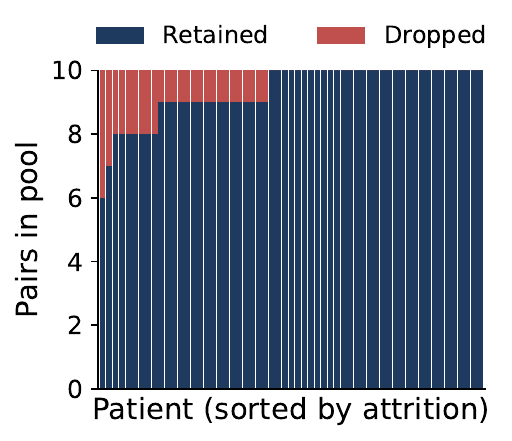}
\caption{Coverage attrition is evenly distributed across patients.
Each of the 59 patients contributes 10 candidate pairs to the
590-pair pool; 38 pairs are dropped for lack of complete five-judge
coverage. No patient loses more than 4 pairs, and 33 of 59 patients
lose none --- the 552-pair evaluation set is not biased toward any
patient subset.}
\label{fig:attrition}
\end{figure}

\begin{figure}[tp]
\centering
\includegraphics[width=0.7\linewidth]{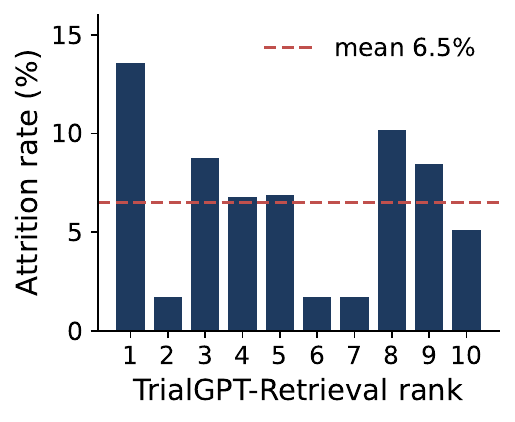}
\caption{Coverage attrition is uniform across retrieval rank. Each
TrialGPT-Retrieval rank contributes 57--59 candidate pairs to the
590-pair pool; the 38 dropped pairs show no monotonic trend with rank
(per-rank attrition 1.7--13.6\%, mean 6.5\%). The 552-pair evaluation
set is therefore not biased toward easily- or poorly-retrieved trials.}
\label{fig:attrition-rank}
\end{figure}

The 38 excluded pairs could in principle distort the benchmark if they were
concentrated on particular patients or retrieval ranks. Figures~\ref{fig:attrition}
and~\ref{fig:attrition-rank} therefore examine attrition by patient and by
TrialGPT-Retrieval rank. The exclusions are diffuse: most patients retain all
10 candidate pairs, no patient loses more than four, and the dropped pairs show
no monotonic pattern across retrieval rank. This suggests that the decisive
552-pair set preserves the structure of the original 590-pair candidate pool.

\begin{table}[htbp]
\centering
\small
\begin{tabular}{@{}lr@{}}
\toprule
Property & Value \\
\midrule
Patient--trial pairs        & 552 \\
\quad \textsc{eligible}     & 278 \\
\quad \textsc{ineligible}   & 274 \\
Unique patients             & 59 \\
Unique trials               & 479 \\
Pairs per patient           & 6--10 (mean 9.4) \\
Patient vignette length (words)   & 70 (median), 38--204 \\
Trial text length (words)         & 243 (median), 47--1{,}595 \\
\bottomrule
\end{tabular}
\caption{Evaluation dataset statistics.}
\label{tab:dataset-stats}
\end{table}

\paragraph{TREC 2021.}
Our second evaluation uses the TREC~2021 Clinical Trials
track~\citep{soboroff2021overview}. The track's relevance judgments are
three-valued; we keep the two that carry an eligibility decision, mapping
score~2 to \textsc{eligible} and score~1 to \textsc{ineligible}, and drop
score~0 (``not relevant''), which marks trials unrelated to the patient
rather than patients ruled out by eligibility criteria. From this judged
pool we draw a class-balanced sample of $1{,}000$ pairs ($500$ per
judgment class) and evaluate on the $363$ of them whose trial has a
compiled eligibility program. Because the sample is balanced by
construction, the evaluation set is near-parity ($52.3\%$
\textsc{eligible}) rather than following the pool's base rate; F1 in
Table~\ref{tab:trec_accuracy} should be read against that. Every system
is scored on the same $363$ pairs. Table~\ref{tab:trec-stats} summarizes the
pool, the sample, and the evaluation set.

\begin{table}[htbp]
\centering
\small
\begin{tabular}{@{}lr@{}}
\toprule
Property & Value \\
\midrule
\multicolumn{2}{@{}l}{\emph{Judged pool}} \\
Judged pairs (score 1 or 2)  & 11{,}589 \\
Unique patient topics        & 75 \\
Unique trials                & 10{,}138 \\
\midrule
\multicolumn{2}{@{}l}{\emph{Class-balanced sample}} \\
Pairs ($500$ per class)      & 1{,}000 \\
\midrule
\multicolumn{2}{@{}l}{\emph{Evaluation set}} \\
Patient--trial pairs         & 363 \\
\quad \textsc{eligible}      & 190 \\
\quad \textsc{ineligible}    & 173 \\
Unique patient topics        & 57 \\
Unique trials                & 263 \\
Pairs per topic              & 1--23 (mean 6.4) \\
Patient topic length (words) & 136 (median), 61--218 \\
Trial criteria length (words) & 92 (median), 10--1{,}173 \\
\bottomrule
\end{tabular}
\caption{TREC~2021 Clinical Trials statistics: the judged pool, the
class-balanced sample drawn from it, and the evaluation set used in
Table~\ref{tab:trec_accuracy}.}
\label{tab:trec-stats}
\end{table}

\section{Descriptive Statistics}
\label{app:descriptive}

Where Appendix~\ref{app:dataset} reports basic corpus properties (counts
of patients, trials, and retrieval-derived patient--trial pairs), this
appendix reports the \emph{output-side} statistics: how the five matchers'
verdicts distribute, what rationales they produce, how the systems
disagree with each other and with the reference panel, and the
composition of the clinician audit sample.

\subsection{Per-system verdict balance and rationale length}
\label{app:desc_verdicts}

Table~\ref{tab:desc_verdicts} shows, for each matcher, the fraction of
the $552$-pair comparison set labelled \emph{eligible}, alongside the
median and mean character length of the system's free-text rationale.
Two patterns stand out. First, eligibility-leaning systems
(\name{}\,$=$\,$55.4\%$, \textsc{ZSPM}\,$=$\,$54.0\%$) come close to the
reference base rate ($50.4\%$ eligible), while the prompted-LLM family
(\ourLLM{}\,$=$\,$35.0\%$, CoT LLM\,$=$\,$36.1\%$) is markedly
ineligibility-leaning. Second, rationale length varies by a factor of
$\approx 2$ across systems, from CoT LLM's compact per-criterion lists
($\approx 590$ chars median) to \textsc{ZSPM}'s exhaustive per-criterion
breakdown ($\approx 1340$ chars median), and these length differences
are not collinear with accuracy --- the longest rationale is not the
most-preferred in the clinician audit
(Table~\ref{tab:rationale_preference}).

\begin{table}[ht]
\centering
\small
\setlength{\tabcolsep}{4pt}
\begin{tabular}{l c c c}
\toprule
System & Elig. (\%) & Median chars & Mean chars \\
\midrule
\name{}    & 55.4 & 1021 & 1030 \\
\ourLLM{}  & 35.0 & 1195 & 1187 \\
CoT LLM    & 36.1 &  592 &  728 \\
TrialGPT   & 44.4 &  594 &  708 \\
\textsc{ZSPM} & 54.0 & 1342 & 1315 \\
\midrule
\emph{Reference panel} & \emph{50.4} & --- & --- \\
\bottomrule
\end{tabular}
\caption{Verdict balance and rationale length per system on the
552-pair comparison set.}
\label{tab:desc_verdicts}
\end{table}

\subsection{Disagreement strata}
\label{app:desc_strata}

For each \name{}-vs-baseline pair, every comparison-set pair falls into
one of four strata by joint agreement with the reference panel:
\emph{both\_right} (both match), \emph{\name{}\_right} (only \name{}
matches), \emph{comp\_right} (only the baseline matches),
\emph{both\_wrong} (neither matches). Table~\ref{tab:desc_strata} shows
the corpus-wide distribution per baseline at the GPT-4.1 backbone. Two
observations: (1) all four strata are populated for every baseline ---
no baseline is strictly dominated --- so the audit can sample
discriminating pairs in both directions; (2) disagreement strata are
$\approx 30$--$33\%$ of the corpus, large enough that the audit's
stratified $4/4/4/4$ design rests on real population mass, not on
hand-picked outliers.

\begin{table*}[ht]
\centering
\small
\setlength{\tabcolsep}{4pt}
\begin{tabular}{l c c c c}
\toprule
Baseline & both\_right & \name{}\_right & comp\_right & both\_wrong \\
\midrule
\ourLLM{}     & 67.4 & 15.2 & 12.7 & 4.7 \\
CoT LLM       & 68.7 & 13.9 & 12.0 & 5.4 \\
TrialGPT      & 67.5 & 15.1 &  9.8 & 7.6 \\
\textsc{ZSPM} & 68.3 & 14.3 & 11.4 & 6.0 \\
\bottomrule
\end{tabular}
\caption{Disagreement stratum distribution (\%) per
\name{}-vs-baseline at GPT-4.1, computed on the $552$-pair comparison
set. Rows sum to $100\%$ modulo rounding.}
\label{tab:desc_strata}
\end{table*}

\subsection{Reference-panel agreement}
\label{app:desc_panel_agreement}

Within the 5-judge GPT-5 panel, $446$ of $552$ pairs ($80.8\%$) receive
a unanimous $5$--$0$ verdict; the remaining $106$ split $4$--$1$
($69$ pairs, $12.5\%$) or $3$--$2$ ($37$ pairs, $6.7\%$). The split
distribution is roughly symmetric across the two verdict directions
($43$+$26$ eligible-leaning vs ineligible-leaning $4$--$1$;
$20$+$17$ at $3$--$2$). This intra-panel disagreement is why a single-call
GPT-5 matcher can diverge from the 5-judge consensus.

\subsection{Clinician audit sample composition}
\label{app:desc_audit}

The audit comprises $32$ patient--trial pairs: $16$ for the
\name{}-vs-\textsc{ZSPM} comparison and $16$ for \name{}-vs-
\ourLLM{}, each balanced $4/4/4/4$ across the four disagreement strata
above. All $32$ pairs receive a holistic A/B/tie preference and per-axis
Likert ratings on five axes (Sec.~\ref{sec:eval-rationale}); on the
$15$ pairs where the clinician's initial verdict disagreed with the
reference panel, the same clinician produced a second judgement on a
re-read with the rationales and the per-framing judge breakdown
visible. The re-review outcome distribution is shown in
Table~\ref{tab:desc_rereview}: the modal outcome was AMBIGUOUS, four
revisions all moved \emph{toward} the reference label, and only a
single case stood as a clear panel--clinician disagreement after
re-review.

\begin{table}[ht]
\centering
\small
\setlength{\tabcolsep}{4pt}
\begin{tabular}{l c}
\toprule
Re-review outcome & Count \\
\midrule
AMBIGUOUS                   & 10 \\
REVISE $\to$ eligible       &  2 \\
REVISE $\to$ ineligible     &  2 \\
UNCHANGED (against panel)   &  1 \\
\midrule
\textit{Total}              & 15 \\
\bottomrule
\end{tabular}
\caption{Outcome distribution of the $15$-pair clinician re-review,
applied to every pair where the initial clinician verdict disagreed
with the reference panel.}
\label{tab:desc_rereview}
\end{table}

\subsection{Chart and trial-criterion length}
\label{app:desc_lengths}

Patient charts in the SIGIR corpus are unusually compact compared to
hospital-EHR notes: median chart length is $440$ characters
(mean $494$, range $245$--$1{,}290$). Trial documents are larger:
median total text $\approx 1{,}535$ characters, of which median
inclusion section $313$ chars and median exclusion section $358$ chars.
Both distributions are right-skewed; the longest trial in the corpus
exceeds $5{,}000$ characters of criteria text. These length statistics
matter for prompting: a single chart+trial prompt fits comfortably in a
$4$--$8$\,K context, and we do not need any chunking or
retrieval-within-trial logic to handle the longest cases.

\subsection{TREC 2021 chart and criterion length}
\label{app:desc_trec_lengths}

Table~\ref{tab:desc_trec_lengths} reports the same length statistics for
the TREC~2021 evaluation set (Appendix~\ref{app:dataset}): $57$ patient
topics and the $263$ trials they are paired with. Two differences from
the SIGIR corpus matter for prompting. First, TREC patient topics are
roughly twice as long as SIGIR charts ($854$ vs.\ $440$ median
characters) and are narrative case descriptions rather than terse
vignettes, so more criteria are settled by stated chart facts and fewer
fall to the missingness policy. Second, criterion text is comparable in
size ($314{+}257$ vs.\ $313{+}358$ median characters for the inclusion
and exclusion sections), but its distribution is far more skewed: the
longest trial carries $65$ criteria against a median of $9$. Since
eligibility is a conjunction, this tail is where per-criterion errors
compound, and it is the population the failure analysis in
Appendix~\ref{sec:parsing-audit} draws from. Both corpora still fit a
single chart-plus-trial prompt without chunking.

\begin{table}[ht]
\centering
\small
\setlength{\tabcolsep}{4pt}
\begin{tabular}{@{}lr@{}}
\toprule
Property & Median (range) \\
\midrule
Patient topic length (chars)  & 854 (372--1{,}293) \\
Inclusion section (chars)     & 314 (0--2{,}165) \\
Exclusion section (chars)     & 257 (0--6{,}016) \\
Criteria text, both sections  & 655 (72--7{,}815) \\
\midrule
Inclusion criteria (count)    & 4 (0--18) \\
Exclusion criteria (count)    & 5 (0--55) \\
Criteria per trial (count)    & 9 (1--65) \\
\bottomrule
\end{tabular}
\caption{TREC~2021 chart and criterion length over the evaluation set
($57$ topics, $263$ trials). Criterion counts are source-text bullet
items, not decomposed atomic conditions.}
\label{tab:desc_trec_lengths}
\end{table}

\section{Reference-label derivation pipeline} \label{app:gold_derivation}

This appendix documents the pipeline used to derive the binary
\emph{eligible}/\emph{ineligible} reference labels used in
Section~\ref{sec:experimental_setup}. The pipeline takes a patient--trial pair as input,
queries five rubric-framing LLM judges in parallel, and assigns a reference
label by simple majority vote. These labels are used for relative system
comparison; they are not intended to replace clinician adjudication.

\subsection{Pipeline overview}

\begin{enumerate}
  \item \textbf{Pair pool construction.} We start from the 590-pair
  retrieval shortlist, consisting of 59 SIGIR patients and the top-10 retrieved
  trial candidates for each patient.

  \item \textbf{Candidate-rationale collection.} For each pair, we collect
  five candidate rationales from the matchers being compared. Before judge
  review, each rationale is rewritten into a shared prose format by a
  fidelity-preserving verbalizer. The verbalizer normalizes presentation only:
  it is instructed to narrate the matcher's existing decision artifact without
  introducing new clinical inferences.

  \item \textbf{Five-judge ensemble.} The patient chart, trial text, and five
  blinded candidate rationales are presented to five LLM judges. The judges use
  the same intended eligibility policy but receive different rubric framings:
  pragmatic screening, protocol adherence, schema-based review,
  checklist-driven scoring, and explanation-focused assessment. Candidate
  rationale slots are randomized independently for each pair and framing.

  \item \textbf{Independent verdict extraction.} Each judge first produces an
  independent verdict, \emph{eligible}, \emph{ineligible}, or
  \emph{inconclusive}, before rating any candidate rationale. Only the
  independent verdict is used for reference-label aggregation. Candidate
  rationale ratings from the same calls are used only for the rationale-quality analysis in Section~\ref{sec:eval-rationale} and Appendix~\ref{app:detailed_results}.

  \item \textbf{Majority aggregation.} A pair receives a reference label only
  when all five judges return decisive verdicts. The label is
  \emph{eligible} when at least three of the five judges vote eligible, and
  \emph{ineligible} otherwise.

  \item \textbf{Coverage attrition.} Of the 590-pair pool, 552 pairs receive
  complete five-judge coverage and constitute the final reference-evaluation
  set. The remaining 38 pairs are excluded because at least one required
  candidate artifact or judge output was unavailable.
\end{enumerate}

\subsection{Why a five-member LLM reference panel?}

Constructing eligibility judgments at this scale requires a practical
trade-off. Clinician review is the most clinically meaningful form of
validation, but it is difficult to apply redundantly to hundreds of
patient--trial pairs. A single LLM judge is scalable, but can be sensitive to
rubric wording on borderline cases even when the intended eligibility policy is
unchanged~\citep{zheng2023judging,panickssery2024llm}. We therefore use a
five-member LLM reference panel as a scalable comparison protocol, not as a
clinical gold standard. The panel estimates which verdicts are stable under
different presentations of the same eligibility policy.

The five framings vary the presentation and decision posture of the rubric
while keeping the intended policy fixed. They are not random paraphrases; they
cover common ways eligibility decisions are specified or requested in practice.

\textbf{Pragmatic screening.}
This framing presents the task as a prescreening decision: whether the patient
should remain a plausible candidate for the trial and which issues would
require follow-up.

\textbf{Protocol adherence.}
This framing presents the task as a protocol-compliance decision: whether the
available chart evidence establishes that the patient satisfies the inclusion
criteria and violates no exclusion criteria under the stated policy.

\textbf{Schema-based review.}
This framing expresses the same decision policy as an operational procedure:
identify inclusion requirements, identify exclusion requirements, map patient
evidence to each requirement, track missing or uncertain facts, and then
produce the final verdict.

\textbf{Checklist-driven scoring.}
This framing is deliberately terse and criterion-driven. It asks the model to
treat the trial as a checklist, helping expose cases where a narrative prompt
may skip less salient criteria, hidden conjuncts, temporal restrictions,
exception clauses, or exclusions.

\textbf{Explanation-focused assessment.}
This framing uses a naturalistic explanation-oriented style, reflecting how
eligibility systems are often queried in practice: users ask not only for a
verdict but also for a readable justification.

Together, the five framings span key prompt degrees of freedom that could
otherwise become hidden evaluation artifacts: screening versus
protocol-compliance posture, clinical versus procedural voice, narrative
versus checklist structure, and explanation-oriented versus canonical decision
form. A pair enters the reference-evaluation set only when all five framings
produce decisive verdicts; the majority vote is then used as an LLM-based
reference label for relative system comparison.

\subsection{Aggregation rule}

Let $J = \{j_1, \ldots, j_5\}$ denote the five judge framings and
$v_j(p) \in \{\textsc{elig}, \textsc{inelig}, \bot\}$ the verdict of
judge $j$ on pair $p$, where $\bot$ denotes a non-decisive output, parse
failure, or model error. Let
$n_{\mathrm{elig}}(p)=|\{j \in J : v_j(p)=\textsc{elig}\}|$.
The reference label $r(p)$ is defined as:

{\footnotesize
\begin{equation}
r(p) =
\begin{cases}
\textsc{elig} & 
  \begin{array}{l}
  n_{\mathrm{elig}}(p) \geq 3 \\
  \text{and } \bot \notin \{v_j(p):j\in J\},
  \end{array}
\\[0.8em]
\textsc{inelig} &
  \begin{array}{l}
  n_{\mathrm{elig}}(p) \leq 2 \\
  \text{and } \bot \notin \{v_j(p):j\in J\},
  \end{array}
\\[0.8em]
\textsc{undefined} & \text{otherwise.}
\end{cases}
\end{equation}%
}
The 552 evaluation pairs distribute as 278 eligible and 274 ineligible under this rule.

\subsection{Judge prompts}
\label{app:reference_judges}

The five prompts below differ in rhetorical framing but encode the same
eligibility policy. Placeholder tokens for the patient note, trial text, and
candidate rationales are substituted at call time. Each judge outputs strict
JSON; the independent verdict field is the vote used in the majority
aggregation.

\PromptBlock
{Pragmatic-screening judge prompt.}
{app:prompt_pragmatic_screening}
{https://github.com/verdict1234/verdict_prompts/blob/main/prompts/judges/accuracy_and_sharpness_clinician_v2.prompt}
  {prompts/judges/accuracy_and_sharpness_clinician_v2.prompt}

\PromptBlock
{Protocol-adherence judge prompt.}
{app:prompt_protocol_adherence}
{https://github.com/verdict1234/verdict_prompts/blob/main/prompts/judges/accuracy_and_sharpness_clinician_paraphrase.prompt}
  {prompts/judges/accuracy_and_sharpness_clinician_paraphrase.prompt}

\PromptBlock
{Schema-based-review judge prompt.}
{app:prompt_schema_review}
{https://github.com/verdict1234/verdict_prompts/blob/main/prompts/judges/accuracy_and_sharpness_engineering_canonical.prompt}
  {prompts/judges/accuracy_and_sharpness_engineering_canonical.prompt}

\PromptBlock
{Checklist-driven judge prompt.}
{app:prompt_checklist}
{https://github.com/verdict1234/verdict_prompts/blob/main/prompts/judges/accuracy_and_sharpness_mechanical.prompt}
  {prompts/judges/accuracy_and_sharpness_mechanical.prompt}

\PromptBlock
{Explanation-focused judge prompt.}
{app:prompt_explanation_focused}
{https://github.com/verdict1234/verdict_prompts/blob/main/prompts/judges/accuracy_and_sharpness_rhetorical.prompt}
  {prompts/judges/accuracy_and_sharpness_rhetorical.prompt}

\subsection{Candidate-rationale verbalizer prompt}
\label{app:reference_verbalizer}

Before judge review, each candidate rationale is normalized into a shared prose
format. The prompt below instructs the verbalizer to preserve the underlying
decision content while removing presentation differences across systems.

\PromptBlock
{Candidate-rationale verbalizer prompt.}
{app:prompt_candidate_verbalizer}
{https://github.com/verdict1234/verdict_prompts/blob/main/prompts/verbalizers/_freeform_rationale.prompt}
  {prompts/verbalizers/_freeform_rationale.prompt}
\section{LLM Reference Panel Validation}
\label{app:reference_validation}

This appendix validates the LLM reference panel against an expert clinician.
The audit is not intended as a full gold-standard benchmarking exercise: that
would require a multi-rater clinician panel at a larger scale. Instead, we use
a targeted stratified audit that concentrates clinician effort on the
patient--trial pairs where reference-panel errors would be most consequential
for system comparison.

\subsection{Stratified audit design}
\label{app:audit_design}

We sample 32 patient--trial pairs from the 552-pair comparison set. The sample
is stratified into four cells defined by the joint agreement of \name{}, the
strongest baseline matcher, and the reference verdict:

\begin{itemize}\itemsep 0pt
  \item \emph{Both agree with reference} (corpus frequency 82.1\%):
        both matchers agree with the reference verdict;
  \item \emph{Only \name{} agrees with reference} (8.2\%):
        \name{} agrees with the reference, while the baseline does not;
  \item \emph{Only baseline agrees with reference} (5.8\%):
        the baseline agrees with the reference, while \name{} does not;
  \item \emph{Neither agrees with reference} (4.0\%):
        both matchers disagree with the reference, making the pair especially
        likely to reveal a reference-panel error.
\end{itemize}

Uniform sampling across these four cells oversamples the disagreement strata
relative to their corpus frequencies. This is deliberate: the \emph{both agree}
cell is already well supported by the reference panel, while the off-diagonal
cells are where reference correctness most affects system-level conclusions.
The audit is therefore an information-concentration design, not an estimate of
unconditional corpus-level clinician agreement.

\subsection{Re-review protocol}
\label{app:rereview_protocol}

The clinician first issued an \emph{independent} eligibility verdict on each of
the 32 audited pairs, blind to the reference verdict. On the 15 pairs where the
clinician's initial verdict differed from the reference, we ran a structured
re-review. For each such pair:

\begin{enumerate}\itemsep 0pt
  \item The clinician was shown their original verdict, the reference verdict,
        the five reference judges' free-text reasoning, and 1--3 matcher
        rationales whose verdict aligned with the reference;
  \item The clinician was asked to either \textbf{REVISE} the verdict,
        \textbf{stand by} the original verdict (\textsc{unchanged}), or mark
        the pair as \textbf{AMBIGUOUS}, meaning that both interpretations were
        clinically defensible;
  \item The clinician provided a one-line written reason for the final
        re-review decision.
\end{enumerate}

This is an \emph{audit-and-resolve} design, not a blinded re-annotation. The
clinician was deliberately exposed to the reference panel's reasoning, analogous
to a second-read step in a clinical chart audit. The goal was to distinguish
true reference-panel errors from cases where the clinician and the reference
panel were applying different but clinically defensible interpretations of the
same chart. The one \textsc{unchanged} case in
Table~\ref{tab:desc_rereview} shows that the clinician retained the ability
to disagree with the reference after seeing its reasoning.

\subsection{Re-review outcomes}
\label{app:rereview_outcomes_sec}

Table~\ref{tab:desc_rereview} summarizes the clinician's final decision on
the 15 re-reviewed pairs.

The 10 ambiguous pairs are not refusals to decide. In each case, the
clinician's written reason explicitly endorsed the reference-panel reading as
one clinically defensible interpretation. Representative reasons include:

\begin{quote}
\emph{``acceptable if listed as a to-do at the screening visit''}\\
\emph{``the chart is incomplete for clinical evidence of X, acceptable if
[verified at screening]''}\\
\emph{``the description has no clear statement about established diagnosis, but
the clinical picture is consistent''}
\end{quote}

These cases reflect a literal-vs-prescreen-doctrine interpretation gap. Under a
strict literal reading, a chart-silent load-bearing inclusion criterion may make
the patient ineligible from the chart alone. Under the reference panel's
prescreening doctrine, the same missing evidence may be deferred to the
in-person screening visit when the chart contains no contradictory evidence.

\subsection{Population-reconstructed \texorpdfstring{$\kappa$}{kappa}}
\label{app:popkappa}

Because the audit deliberately samples non-uniformly, the raw audit sample does
not match the cell distribution of the 552-pair comparison set. Each of the four
audit cells contributes roughly one quarter of the audit sample, while the same
cells appear in the full comparison set at rates of 82.1\%, 8.2\%, 5.8\%, and
4.0\%. A simple Cohen's $\kappa$~\cite{doi:10.1177/001316446002000104} on the audit sample would therefore answer how
often the clinician agrees with the reference \emph{on this stratified sample},
not how often agreement would be expected under the corpus cell distribution.

To recover a corpus-level estimate, we use inverse-probability weighting (IPW).
Each audited pair $i$ in cell $c$ receives weight
\[
w_i =
\frac{n_c^{\mathrm{corpus}}/N^{\mathrm{corpus}}}
     {n_c^{\mathrm{audit}}/N^{\mathrm{audit}}},
\]
where $n_c^{\mathrm{corpus}}$ and $n_c^{\mathrm{audit}}$ are the number of
pairs in cell $c$ in the full comparison set and in the audit sample, and
$N^{\mathrm{corpus}}$ and $N^{\mathrm{audit}}$ are the corresponding totals
over all four cells. We then compute $\kappa$ using the weighted observed
agreement $p_o^w$ and the weighted expected (chance) agreement $p_e^w$:
\[
p_o^w =
\frac{\sum_i w_i \mathbf{1}[c_i=r_i]}{\sum_i w_i},
\qquad
\kappa^{\mathrm{pop}} =
\frac{p_o^w - p_e^w}{1-p_e^w},
\]
where $c_i$ and $r_i$ are the clinician and reference verdicts for pair $i$,
$\mathbf{1}[\cdot]$ is the indicator function, and $p_e^w$ is the chance
agreement obtained by combining the $w_i$-weighted marginal verdict
distributions of the clinician and of the reference in the usual Cohen
form.
This reweighting estimates the clinician--reference agreement we would expect
if the audit's per-cell agreement rates held across the full comparison set.

\subsection{Handling AMBIGUOUS responses}
\label{app:ambig_conventions}

Because AMBIGUOUS is not a binary verdict, it must be converted before
computing agreement and $\kappa$. We report two conventions:

\begin{itemize}\itemsep 0pt
  \item \textbf{Half-credit.} Each ambiguous pair is treated as 50/50
        agreement and disagreement. We compute $\kappa$ as the expected value of
        the binary $\kappa$ over assignments of the ambiguous pairs, estimated
        by 10{,}000 Monte Carlo draws. This is the neutral convention and our
        headline audit statistic.
  \item \textbf{Lenient.} Each ambiguous pair is treated as agreement with the
        reference, because the clinician's written reason endorsed the
        reference interpretation as clinically defensible. This provides an
        upper-bound convention.
\end{itemize}

We do not report a strict convention that treats AMBIGUOUS as disagreement,
because it would contradict the clinician's written reasons on every ambiguous
pair.

\subsection{Results}
\label{app:reference_validation_results}

After structured re-review, clinician--reference agreement is high under both
ambiguity conventions (Table~\ref{tab:stratum_agreement}). Under the half-credit
convention, raw agreement is
81.3\% and the population-reconstructed $\kappa$ is 0.67. Under the lenient
convention, raw agreement is 96.9\% and the population-reconstructed $\kappa$
is 0.97. These results support the use of the LLM reference panel for relative
system comparison, while preserving the distinction between a scalable reference
label and a clinician-adjudicated gold label.

\subsection{Limitations}
\label{app:reference_validation_limitations}

\paragraph{Single rater.}
Only one expert clinician was available within the timeframe of this work. We
therefore cannot estimate a clinician--clinician agreement ceiling for this
specific audit.

\paragraph{Audit-and-resolve, not blinded re-annotation.}
The clinician saw the reference panel's reasoning during re-review. This was a
deliberate design choice: the goal was to resolve apparent disagreements by
showing the evidence and reasoning that led to the reference verdict, not to
run an independent second blinded annotation. The one \textsc{unchanged} case
shows that exposure to the reference reasoning did not force agreement.

\paragraph{Scope of claim.}
We do not claim that the LLM reference panel is clinician-equivalent. We claim
that, on the audit cells where reference errors would most affect system
comparison, most apparent disagreements were attributable to a
literal-vs-prescreen-doctrine interpretation gap rather than clear
reference-panel errors.

\subsection{User Interface for Clinician Annotation}
\label{app:user_interface_for_clinician_annotation}

\begin{figure*}[t]
\centering
\begin{subfigure}{0.48\textwidth}
  \centering
  \includegraphics[width=\linewidth]{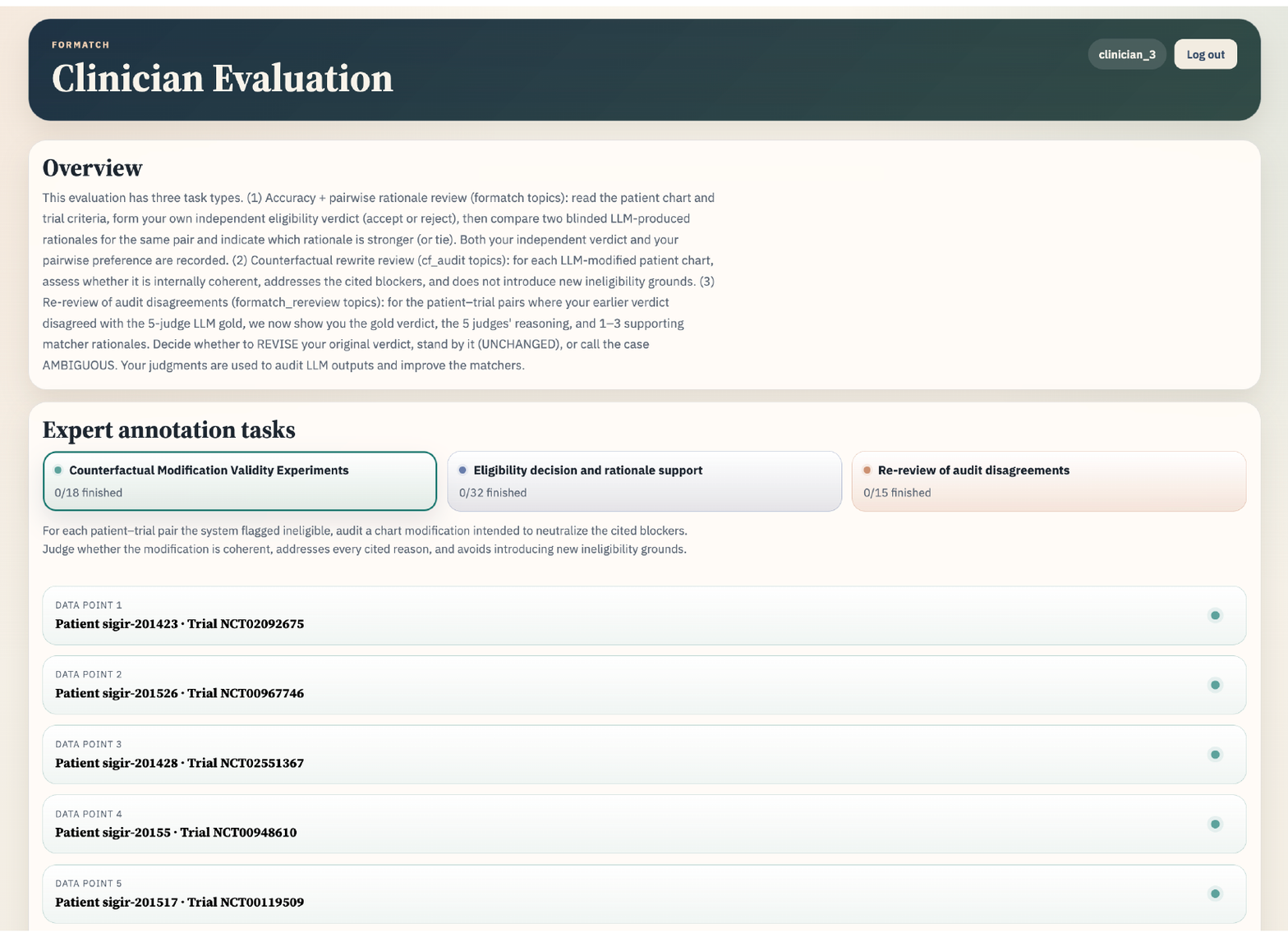}
  \caption{Overview screen.}
  \label{fig:clinician_judge_overview}
\end{subfigure}
\hfill
\begin{subfigure}{0.48\textwidth}
  \centering
  \includegraphics[width=\linewidth]{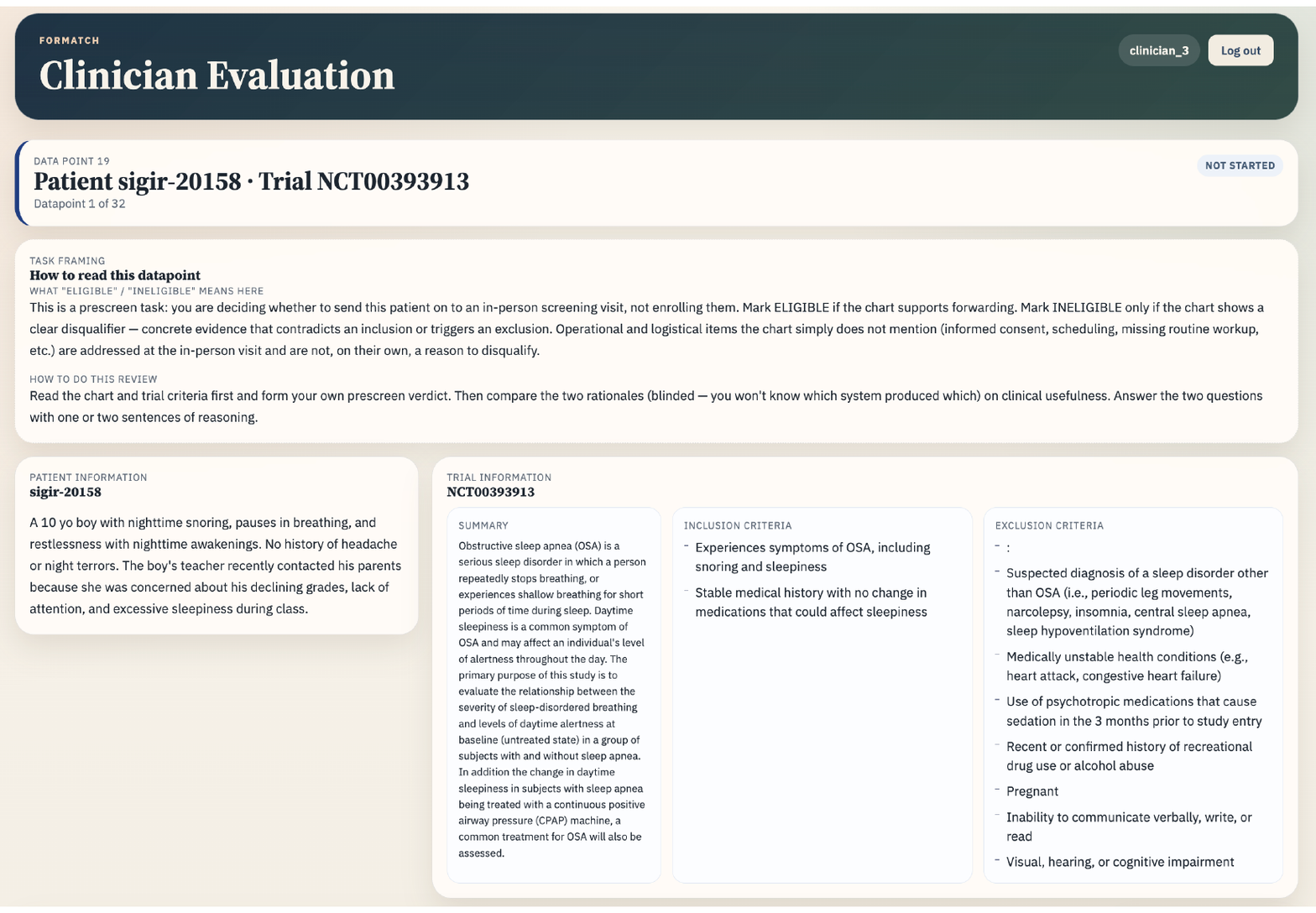}
  \caption{Eligibility task (1/2).}
  \label{fig:clinician_judge_eligibility1}
\end{subfigure}

\vspace{1em}

\begin{subfigure}{0.48\textwidth}
  \centering
  \includegraphics[width=\linewidth]{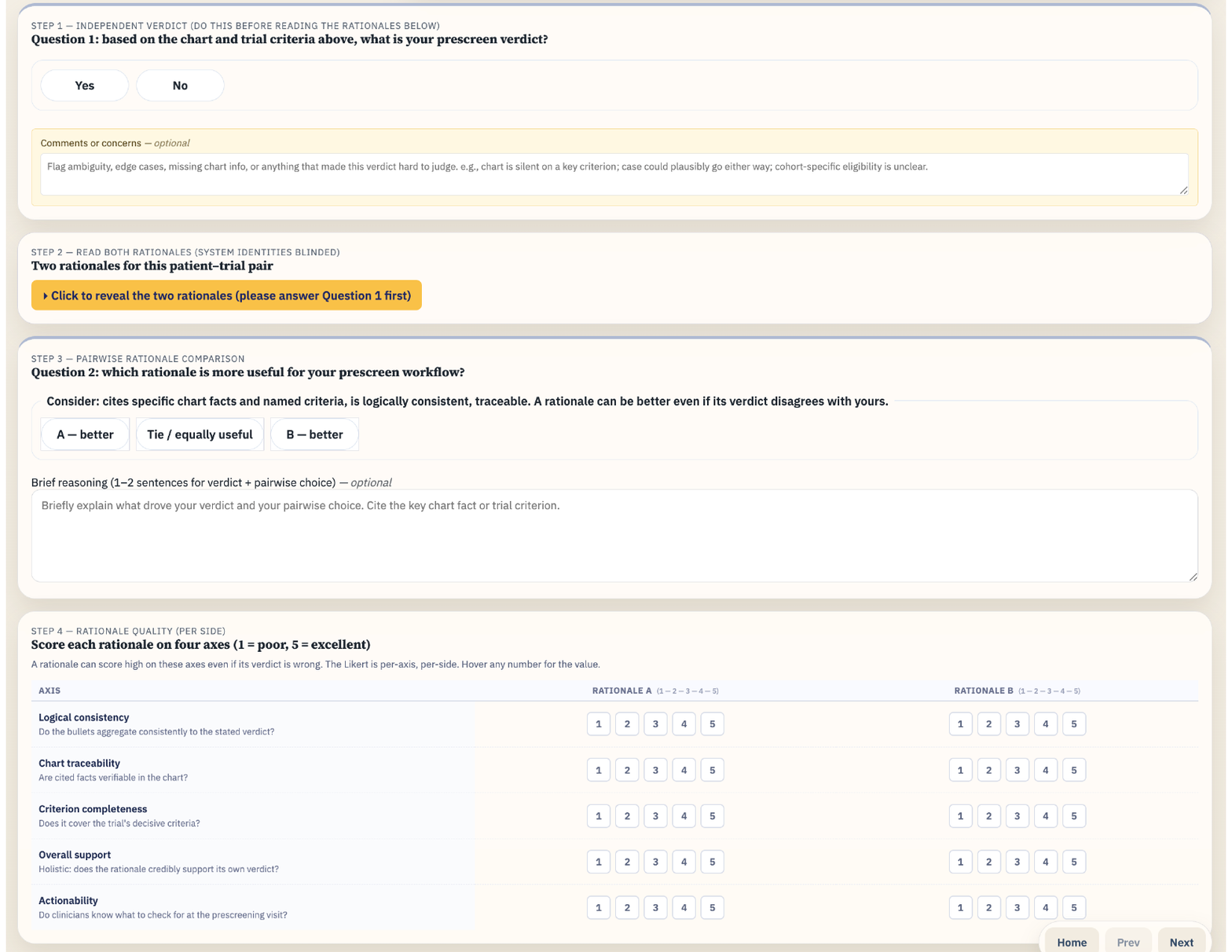}
  \caption{Eligibility task (2/2).}
  \label{fig:clinician_judge_eligibility2}
\end{subfigure}
\hfill
\begin{subfigure}{0.48\textwidth}
  \centering
  \includegraphics[width=\linewidth]{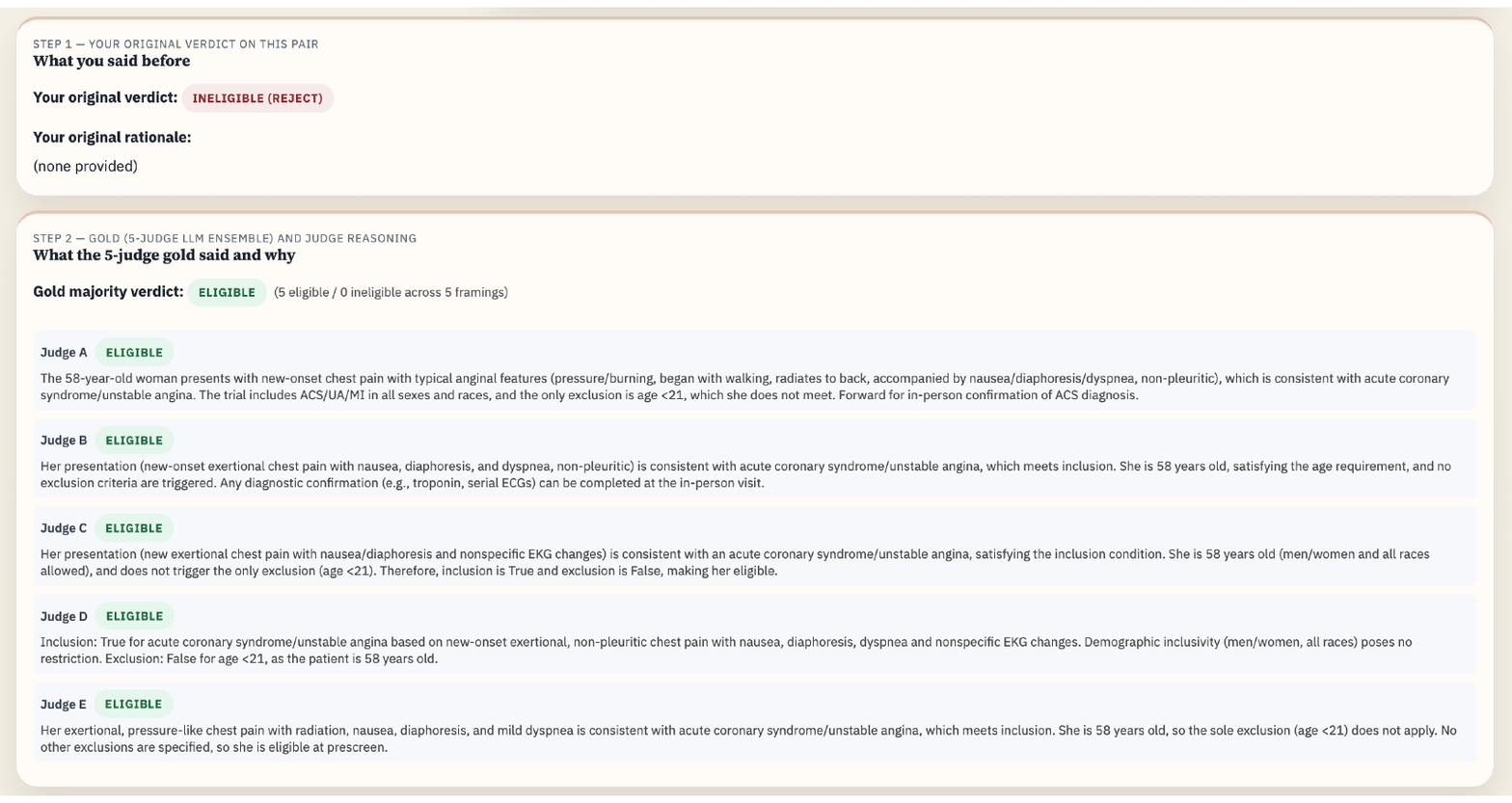}
  \caption{Re-review (1/2)}
  \label{fig:clinician_rereview1}
\end{subfigure}

\vspace{1em}

\begin{subfigure}{0.48\textwidth}
  \centering
  \includegraphics[width=\linewidth]{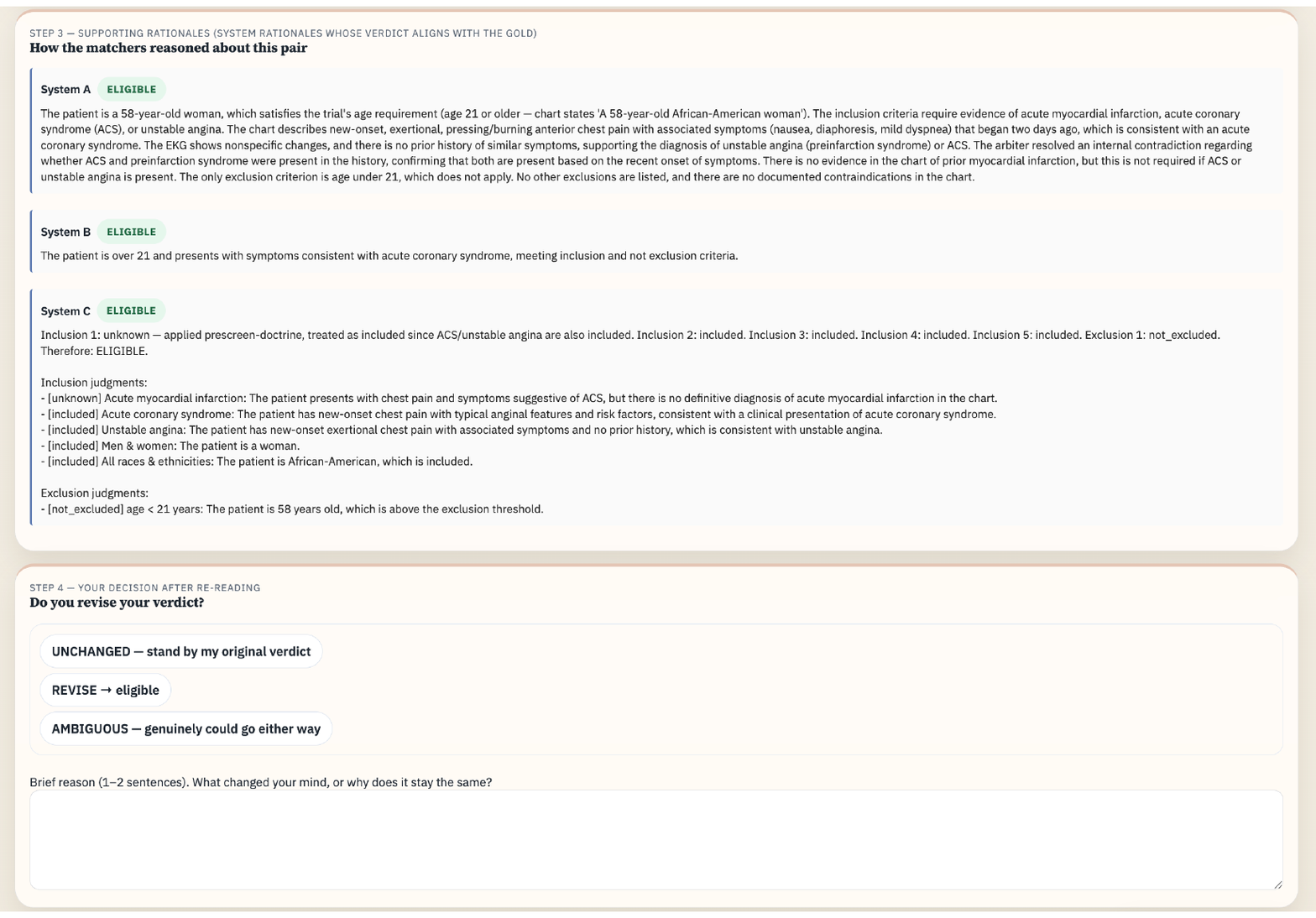}
  \caption{Re-review (2/2)}
  \label{fig:clinician_rereview2}
\end{subfigure}

\caption{Annotation interface presented to the clinician.
(a) overview; (b--c) independent verdict and pairwise rationale comparison;
(d--e) re-review of audit disagreements.}
\label{fig:clinician_ui}
\end{figure*}

Figures~\ref{fig:clinician_judge_overview},~\ref{fig:clinician_judge_eligibility1},~\ref{fig:clinician_judge_eligibility2},~\ref{fig:clinician_rereview1},~\ref{fig:clinician_rereview2} show screenshots of the
annotation interface presented to the clinician during the audit
described in Appendix~\ref{app:reference_validation}. The interface organizes
work into two task types, mirroring the two stages of the audit protocol.

\paragraph{Independent verdict and pairwise rationale comparison
(Figures~\ref{fig:clinician_judge_eligibility1}--\ref{fig:clinician_judge_eligibility2}).}
For each patient--trial pair, the clinician reads the patient chart and the
trial eligibility criteria and issues an independent eligibility verdict. After the verdict is submitted, the
interface reveals two blinded LLM-produced rationales for the same pair and the clinician
indicates which rationale is stronger or marks the pair as a tie. Both the
independent verdict and the pairwise preference are recorded. 

\paragraph{Re-review of audit disagreements
(Figure~\ref{fig:clinician_rereview1}--\ref{fig:clinician_rereview2}).}
For the subset of pairs where the clinician's independent verdict in the first task
differed from the reference panel verdict, the interface re-surfaces the pair
together with the reference verdict, the free-text reasoning from each of the
five reference judges, and one to three matcher rationales whose verdicts
aligned with the reference. The clinician then either \textbf{REVISES} the
original verdict toward the reference, marks the verdict as
\textbf{UNCHANGED} (stands by the original), or labels the pair
\textbf{AMBIGUOUS} (both interpretations clinically defensible), and provides
a one-line written reason for the decision. This implements the structured
re-review protocol of Appendix~\ref{app:rereview_protocol};
Figure~\ref{fig:clinician_ui} shows the full interface.

\clearpage
\section{Correlation with SIGIR Referral Judgments}
\label{app:qrels_correlation}

Our SIGIR 2016-derived benchmark reuses the patients and trials of the SIGIR
benchmark~\citep{koopman2016test} but derives new pairwise reference
eligibility labels (\S\ref{sec:benchmarks}). This appendix quantifies how these
reference eligibility labels relate to SIGIR's original referral relevance
judgments (\emph{qrels}). The two are correlated, as expected because referral
and eligibility are related, but they are far from identical. This difference
motivates deriving new eligibility labels rather than reusing SIGIR qrels as
eligibility labels.

\paragraph{Setup.}
SIGIR qrels assign each judged patient--trial pair a score in
$\{0, 1, 2\}$: $0$ = ``would not refer,'' $1$ = ``may refer,''
and $2$ = ``would refer.'' Of our 552 evaluation pairs, 357 carry a SIGIR
qrels score; the remaining 195 were never judged by SIGIR's original pooling
process and therefore have no referral label. This coverage gap already
motivates relabeling for eligibility evaluation.

\begin{center}
\begin{minipage}[t]{0.98\columnwidth}
\centering
\captionsetup{hypcap=false}
\small
\begin{tabular}{@{}lrrrr@{}}
\toprule
Reference label & qrels 0 & qrels 1 & qrels 2 & Total \\
\midrule
\textsc{eligible}   &  56 & 72 & 70 & 198 \\
\textsc{ineligible} & 121 & 31 &  7 & 159 \\
\midrule
Total               & 177 & 103 & 77 & 357 \\
\bottomrule
\end{tabular}
\captionof{table}{Reference eligibility labels vs.\ SIGIR referral qrels on the
357 pairs with both labels.}
\label{tab:reference-qrels}
\end{minipage}

\vspace{1em}

\begin{minipage}[t]{0.98\columnwidth}
\centering
\captionsetup{hypcap=false}
\includegraphics[width=\linewidth]{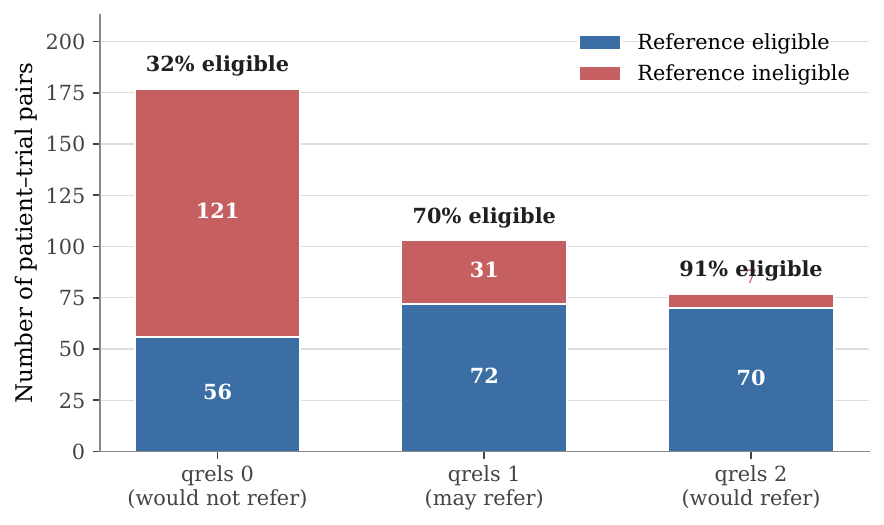}
\captionof{figure}{Reference-eligible fraction rises monotonically with SIGIR
qrels score (32\%\,\texorpdfstring{$\to$}{->}\,70\%\,\texorpdfstring{$\to$}{->}\,91\%), but the two labelings are not
interchangeable.}
\label{fig:reference-qrels}
\end{minipage}
\end{center}

\paragraph{Cross-tabulation.}
Table~\ref{tab:reference-qrels} cross-tabulates our reference eligibility
labels against the SIGIR qrels score on the 357 pairs with both labels, and
Figure~\ref{fig:reference-qrels} plots the same data.
\clearpage

\section{Detailed Results}\label{app:detailed_results}

\paragraph{Overview.}
This appendix reports the detailed results underlying the accuracy, rationale
preference, and counterfactual self-faithfulness analyses in the main paper.
The tables serve three purposes. First, they test whether the main accuracy
conclusions are stable after replacing audited reference labels with the
clinician's post-review verdicts. Second, they give the raw and
population-reweighted versions of the clinician rationale-preference study.
Third, they unpack the counterfactual self-faithfulness pipeline, separating
the validity of generated counterfactuals from whether each matcher actually
changes its decision after the cited blocker is removed.

\paragraph{Clinician-adjusted accuracy.}
Table~\ref{tab:accuracy_adjusted} shows that the main accuracy trends are
stable under clinician adjustment. For GPT-4.1, \name{} remains the strongest
system by lenient F1, with higher precision than \ourLLM{} and higher recall
than the high-precision CoT and TrialGPT baselines. Across GPT-4o and GPT-4o-mini,
\name{} again has the best lenient F1, suggesting that its advantage is not
limited to a single backbone. The lenient and half-credit conventions produce
nearly identical rankings, indicating that the results are not sensitive to how
ambiguous clinician-reviewed cases are counted.

\paragraph{Reference-panel validation.}
Table~\ref{tab:stratum_agreement} summarizes the clinician audit of the
LLM-based reference labels. The clinician initially disagreed with the reference
on a subset of cases, but most disagreements were resolved as clinically
ambiguous after re-review rather than clear reference errors. Under the lenient
convention, clinician--reference agreement is high, supporting the use of the
reference panel for large-scale accuracy evaluation. The half-credit convention
is more conservative, but still shows substantial agreement, especially after
population reconstruction.

\paragraph{Rationale preference.}
Tables~\ref{tab:pairwise_rationale_poprecon},
\ref{tab:pairwise_rationale_cells}, and
\ref{tab:pairwise_axes_poprecon} provide the detailed rationale-preference
results. Table~\ref{tab:pairwise_rationale_poprecon} gives the headline
pairwise preference rates, including population-reconstructed estimates that
correct for the stratified sampling design. Clinicians strongly prefer
\name{}'s rationales over ZSPM and also prefer them over \ourLLM{} overall.
The per-stratum breakdown in Table~\ref{tab:pairwise_rationale_cells} shows
that this preference is not merely an artifact of label accuracy: \name{} is
preferred or tied even in strata where the comparator matches the reference
label and \name{} does not. The per-axis results show that the advantage is
especially strong for actionability, support, chart traceability, and logical
consistency. This supports the main-paper claim that \name{} produces more
useful decision artifacts, not just more favorable labels.

\paragraph{Counterfactual self-faithfulness.}
Tables~\ref{tab:cf_cells12}, \ref{tab:cf_pipeline},
\ref{tab:cf_audit_only}, \ref{tab:cf_clinician_audit},
\ref{tab:cf_raw_breakdown}, and
\ref{tab:cf_self_faithfulness_poprecon} decompose the counterfactual
self-faithfulness analysis. For each rejected pair, we edit the patient chart
to address the matcher’s cited rejection reasons, validate that the edit is
clinically coherent and on-target, and then rerun the same matcher on the
modified chart. \name{-E2E} denotes \name{} evaluated end to end on this
protocol: rerunning the full \name{} pipeline rather than editing solver atoms
directly. Concretely,
the counterfactual chart is parsed again by the patient-side semantic parser,
the resulting constraints are solved again, and the new verdict is produced by
the same SMT-backed decision process. Thus, \name{-E2E} is not evaluated by
directly changing solver atoms or forcing the flip set; it tests whether the
end-to-end system remains self-faithful after counterfactual rewriting and
re-parsing.

The raw flip-rate tables show that \name{-E2E} has the highest flip rate across
modifier--validator settings: when the cited blocker is removed, \name{} is much
more likely than the LLM-based matchers to change its verdict. This is expected
because \name{}'s rejection reasons are generated from the flip set, the minimal
patient facts, imputations, or assumptions whose change makes the constraints satisfiable.
Failures for \name{-E2E} therefore mainly reflect counterfactual rewriting or
semantic-parsing errors, rather than a mismatch between the stated rationale and
the formal decision boundary.

The pipeline and audit tables further distinguish two issues: whether the
generated counterfactual is clinically valid, and whether the matcher actually
flips on that valid counterfactual. After clinician correction and population
reweighting, \name{} remains the strongest system, while several LLM-based
matchers have substantially lower self-faithfulness despite sometimes producing
plausible rationales. In particular, \ourLLM{-Pivotal} asks the same
\ourLLM{} matcher to explicitly list unmet requirements whose change would make
the patient eligible, and uses those listed targets for counterfactual
construction. Its improvement over \ourLLM{} is limited, suggesting that simply
asking an LLM to name flip targets does not reliably recover the model's actual
decision boundary. This supports the claim that free-text rationales often fail
to identify blockers that are sufficient to change the matcher’s own verdict.

\paragraph{Interpretation.}
Taken together, the appendix tables show that the main conclusions are robust
to clinician adjustment, ambiguity conventions, and population reweighting.
\name{} is not uniformly best on every aggregate accuracy metric, but it
consistently provides
strong accuracy together with more preferred rationales and substantially better
counterfactual self-faithfulness. These results support the paper's central
claim: formal matching is valuable not only because it can improve decisions,
but also because it exposes the assumptions and decision boundary behind those
decisions.

\begin{table*}[htbp]
\centering
\caption{Clinician-adjusted accuracy under two conventions for ambiguous audit cases.
Audited labels are replaced by the clinician's post-review verdict. In the lenient
setting, ambiguous cases are counted as agreement; in the half-credit setting,
they receive weight 0.5. Brackets give bootstrap 95\% CIs for F1. Best lenient
F1 per backbone is in \textbf{bold}.}
\label{tab:accuracy_adjusted}
\scriptsize
\begin{tabular}{l l c c c c | c c c c}
\toprule
 & & \multicolumn{4}{c|}{\emph{Lenient}} & \multicolumn{4}{c}{\emph{Half-credit}} \\
Backbone & System & F1 [95\% CI] & P & R & Acc & F1 [95\% CI] & P & R & Acc \\
\midrule
  \multirow{5}{*}{GPT-4.1} & \name{} & \textbf{0.902} [0.873,\,0.927] & 0.927 & 0.877 & 0.904 & 0.900 [0.872,\,0.925] & 0.926 & 0.875 & 0.902 \\
   & \ourLLM{} & 0.886 [0.857,\,0.912] & 0.850 & 0.924 & 0.880 & 0.882 [0.853,\,0.907] & 0.847 & 0.921 & 0.877 \\
   & CoT LLM & 0.718 [0.671,\,0.761] & 0.958 & 0.574 & 0.774 & 0.713 [0.667,\,0.757] & 0.952 & 0.570 & 0.770 \\
   & TrialGPT & 0.360 [0.294,\,0.426] & 0.863 & 0.227 & 0.594 & 0.360 [0.294,\,0.425] & 0.863 & 0.227 & 0.594 \\
   & ZSPM & 0.833 [0.800,\,0.865] & 0.803 & 0.866 & 0.826 & 0.833 [0.801,\,0.865] & 0.803 & 0.866 & 0.826 \\
\midrule
  \multirow{5}{*}{GPT-4o} & \name{} & \textbf{0.837} [0.806,\,0.868] & 0.797 & 0.881 & 0.828 & 0.835 [0.804,\,0.867] & 0.796 & 0.879 & 0.826 \\
   & \ourLLM{} & 0.770 [0.726,\,0.810] & 0.938 & 0.653 & 0.804 & 0.768 [0.725,\,0.808] & 0.935 & 0.652 & 0.803 \\
   & CoT LLM & 0.775 [0.729,\,0.815] & 0.929 & 0.664 & 0.806 & 0.771 [0.726,\,0.810] & 0.924 & 0.661 & 0.803 \\
   & TrialGPT & 0.757 [0.715,\,0.798] & 0.813 & 0.708 & 0.772 & 0.753 [0.712,\,0.792] & 0.809 & 0.704 & 0.768 \\
   & ZSPM & 0.803 [0.769,\,0.838] & 0.775 & 0.834 & 0.795 & 0.805 [0.771,\,0.838] & 0.777 & 0.836 & 0.797 \\
\midrule
  \multirow{5}{*}{GPT-4o-mini} & \name{} & \textbf{0.756} [0.715,\,0.796] & 0.802 & 0.715 & 0.768 & 0.756 [0.716,\,0.796] & 0.802 & 0.715 & 0.768 \\
   & \ourLLM{} & 0.477 [0.415,\,0.536] & 0.957 & 0.318 & 0.650 & 0.474 [0.412,\,0.533] & 0.951 & 0.316 & 0.649 \\
   & CoT LLM & 0.600 [0.541,\,0.654] & 0.912 & 0.448 & 0.701 & 0.600 [0.541,\,0.654] & 0.912 & 0.448 & 0.701 \\
   & TrialGPT & 0.589 [0.533,\,0.645] & 0.861 & 0.448 & 0.687 & 0.587 [0.530,\,0.641] & 0.858 & 0.446 & 0.685 \\
   & ZSPM & 0.689 [0.653,\,0.727] & 0.541 & 0.949 & 0.571 & 0.691 [0.655,\,0.727] & 0.542 & 0.951 & 0.572 \\
\bottomrule
\end{tabular}
\end{table*}

\begin{table*}[htbp]
\centering
\caption{Counterfactual self-faithfulness under alternative counterfactual-generation
settings. Rows vary only the LLMs used to modify patient charts and validate the
resulting counterfactuals. The raw flip rate is the fraction of validator-accepted
counterfactuals on which the matcher changes its verdict to \emph{eligible}.
Best per setting is in \textbf{bold}.}
\label{tab:cf_cells12}
\small
\begin{tabular}{l l c c}
\toprule
Modifier / validator & System & $n_{\text{flipped}}/n_{\text{valid}}$ & Raw flip rate [95\% CI] \\
\midrule
  \multirow{6}{*}{gpt-4.1 / gpt-4.1} & \name{-E2E} & 245/300 & \textbf{81.7\%} [76.9\%,\,85.6\%] \\
   & \ourLLM{} & 175/444 & 39.4\% [35.0\%,\,44.0\%] \\
   & CoT LLM & 469/838 & 56.0\% [52.6\%,\,59.3\%] \\
   & \ourLLM{-Pivotal} & 313/705 & 44.4\% [40.8\%,\,48.1\%] \\
   & TrialGPT & 298/550 & 54.2\% [50.0\%,\,58.3\%] \\
   & ZSPM & 66/395 & 16.7\% [13.4\%,\,20.7\%] \\
\midrule
  \multirow{6}{*}{gpt-4.1 / gpt-5} & \name{-E2E} & 282/348 & \textbf{81.0\%} [76.6\%,\,84.8\%] \\
   & \ourLLM{} & 170/417 & 40.8\% [36.2\%,\,45.5\%] \\
   & CoT LLM & 487/786 & 62.0\% [58.5\%,\,65.3\%] \\
   & \ourLLM{-Pivotal} & 344/707 & 48.7\% [45.0\%,\,52.3\%] \\
   & TrialGPT & 295/501 & 58.9\% [54.5\%,\,63.1\%] \\
   & ZSPM & 69/365 & 18.9\% [15.2\%,\,23.2\%] \\
\bottomrule
\end{tabular}
\end{table*}

\begin{table*}[htbp]
\centering
\caption{Counterfactual self-faithfulness pipeline under the GPT-5 modifier and
GPT-5 validator setting. The modifier attempts $N_{\text{total}}$ counterfactual
rewrites, the validator accepts $N_{\text{valid}}$, and the matcher flips on
$N_{\text{flipped}}$. Raw flip rate is $N_{\text{flipped}}/N_{\text{valid}}$.
Population-reconstructed self-faithfulness adjusts the raw rate using the
clinician audit of flipped and not-flipped buckets. Best values are in
\textbf{bold}.}
\label{tab:cf_pipeline}
\small
\setlength{\tabcolsep}{4pt}
\begin{tabular}{l rrrr l l}
\toprule
System & $N_{\text{total}}$ & $N_{\text{valid}}$ & $N_{\text{flipped}}$ & \% valid & Raw flip rate [95\% CI] & Population-recon.\ SF [95\% CI] \\
\midrule
\name{-E2E} & 580 & 388 & 327 & 67\% & \textbf{84.3\%} [80.3, 87.6] & \textbf{80.9\%} [56.0, 95.0] \\
\ourLLM{} & 488 & 464 & 227 & 95\% & 48.9\% [44.4, 53.5] & 51.4\% [39.8, 67.9] \\
CoT LLM (GPT-4.1) & 1026 & 866 & 563 & 84\% & 65.0\% [61.8, 68.1] & 72.7\% [65.0, 82.3] \\
CoT LLM (GPT-5) & 704 & 602 & 358 & 86\% & 59.5\% [55.5, 63.3] & 67.3\% [56.3, 83.7] \\
\ourLLM{-Pivotal} & 916 & 807 & 466 & 88\% & 57.7\% [54.3, 61.1] & 55.8\% [44.1, 66.9] \\
TrialGPT & 960 & 599 & 366 & 62\% & 61.1\% [57.1, 64.9] & 55.3\% [41.4, 61.4] \\
ZSPM & 496 & 417 & 111 & 84\% & 26.6\% [22.6, 31.1] & 24.1\% [17.4, 29.7] \\
\bottomrule
\end{tabular}
\end{table*}

\begin{table*}[htbp]
\centering
\caption{Counterfactual self-faithfulness on the clinician-audited sample.
The clinician reviewed validator-accepted counterfactuals from the flipped and
not-flipped buckets. Audit-only self-faithfulness is the fraction of
clinician-valid audited counterfactuals on which the matcher flips to
\emph{eligible}. The corpus-reweighted estimate is reported in
Table~\ref{tab:cf_pipeline}.}
\label{tab:cf_audit_only}
\small
\setlength{\tabcolsep}{4pt}
\begin{tabular}{l r rr rr r l}
\toprule
System & $K_{\text{aud}}$ & \multicolumn{2}{c}{Clin.\ valid @ \emph{flipped}} & \multicolumn{2}{c}{Clin.\ valid @ \emph{not flipped}} & Valid total & Audit-only SF [95\% CI] \\
\cmidrule(lr){3-4}\cmidrule(lr){5-6}
 &  & valid & audited & valid & audited &  &  \\
\midrule
\name{-E2E} & 15 & 4 & 8 & 3 & 7 & 7 & 57\% [25, 84] \\
\ourLLM{} & 18 & 8 & 10 & 6 & 8 & 14 & 57\% [33, 79] \\
CoT LLM (GPT-4.1) & 18 & 8 & 8 & 7 & 10 & 15 & 53\% [30, 75] \\
CoT LLM (GPT-5) & 17 & 8 & 9 & 5 & 8 & 13 & \textbf{62\%} [36, 82] \\
\ourLLM{-Pivotal}  & 16 & 7 & 8 & 7 & 8 & 14 & 50\% [27, 73] \\
TrialGPT & 16 & 7 & 9 & 7 & 7 & 14 & 50\% [27, 73] \\
ZSPM & 18 & 8 & 9 & 8 & 9 & 16 & 50\% [28, 72] \\
\bottomrule
\end{tabular}
\end{table*}

\begin{table*}[htbp]
\centering
\caption{Clinician audit used to correct counterfactual self-faithfulness.
$p_{\text{flip}}$ and $p_{\text{not}}$ are clinician-valid rates in the
matcher-flipped and not-flipped buckets; $N_F$ and $N_{NF}$ count the
re-judged counterfactuals in each bucket for this configuration. They are
filtered differently from the $N_{\text{valid}}$ and $N_{\text{flipped}}$ of
Table~\ref{tab:cf_pipeline} and are not expected to sum to $N_{\text{valid}}$.
Clinician-corrected self-faithfulness reweights these rates to the full
corpus, with pair-level bootstrap 95\% CIs. Best value is in
\textbf{bold}.}
\label{tab:cf_clinician_audit}
\small
\begin{tabular}{l c cc cc l}
\toprule
System & $n_{\text{aud}}$ & $p_{\text{flip}}$ & $p_{\text{not}}$ & $N_F$ & $N_{NF}$ & Clinician-corrected SF [95\% CI] \\
\midrule
  \name{-E2E} & 15 & 50.0\% & 42.9\% & 218 & 60 & \textbf{80.9\%} [56.0\%,\,95.0\%] \\
  \ourLLM{} & 18 & 80.0\% & 75.0\% & 113 & 114 & 51.4\% [39.8\%,\,67.9\%] \\
  CoT LLM & 18 & 100.0\% & 70.0\% & 281 & 151 & 72.7\% [65.0\%,\,82.3\%] \\
  CoT LLM & 17 & 88.9\% & 62.5\% & 100 & 69 & 67.3\% [56.3\%,\,83.7\%] \\
  \ourLLM{-Pivotal}  & 16 & 87.5\% & 87.5\% & 227 & 180 & 55.8\% [44.1\%,\,66.9\%] \\
  TrialGPT & 16 & 77.8\% & 100.0\% & 189 & 119 & 55.3\% [41.4\%,\,61.4\%] \\
  ZSPM & 18 & 88.9\% & 88.9\% & 57 & 180 & 24.1\% [17.4\%,\,29.7\%] \\
\bottomrule
\end{tabular}
\end{table*}

\begin{table*}[htbp]
\centering
\caption{Clinician spot-check of counterfactual chart rewrites under the GPT-5
modifier and GPT-5 validator setting. For each system and matcher bucket, the
clinician assessed whether the rewrite was coherent, removed the cited blocker,
and introduced no new blocker. Clinician-valid is the conjunction of all three
criteria.}
\label{tab:cf_raw_breakdown}
\small
\begin{tabular}{l l r cccc}
\toprule
System & Bucket & $n$ & Coherent & Removed cited & No new blocker & Clinician-valid \\
\midrule
\name{} & flipped & 8 & 5/8 (62\%) & 6/8 (75\%) & 8/8 (100\%) & 4/8 (50\%) \\
 & not\_flipped & 7 & 4/7 (57\%) & 4/7 (57\%) & 6/7 (86\%) & 3/7 (43\%) \\
\addlinespace[2pt]
\ourLLM{} & flipped & 10 & 8/10 (80\%) & 10/10 (100\%) & 10/10 (100\%) & 8/10 (80\%) \\
 & not\_flipped & 8 & 8/8 (100\%) & 6/8 (75\%) & 8/8 (100\%) & 6/8 (75\%) \\
\addlinespace[2pt]
CoT LLM & flipped & 8 & 8/8 (100\%) & 8/8 (100\%) & 8/8 (100\%) & 8/8 (100\%) \\
 & not\_flipped & 10 & 10/10 (100\%) & 7/10 (70\%) & 10/10 (100\%) & 7/10 (70\%) \\
\addlinespace[2pt]
CoT LLM (GPT-5) & flipped & 9 & 9/9 (100\%) & 9/9 (100\%) & 8/9 (89\%) & 8/9 (89\%) \\
 & not\_flipped & 8 & 6/8 (75\%) & 6/8 (75\%) & 8/8 (100\%) & 5/8 (62\%) \\
\addlinespace[2pt]
\ourLLM{-Pivotal}  & flipped & 8 & 8/8 (100\%) & 8/8 (100\%) & 7/8 (88\%) & 7/8 (88\%) \\
 & not\_flipped & 8 & 8/8 (100\%) & 7/8 (88\%) & 8/8 (100\%) & 7/8 (88\%) \\
\addlinespace[2pt]
TrialGPT & flipped & 9 & 9/9 (100\%) & 8/9 (89\%) & 8/9 (89\%) & 7/9 (78\%) \\
 & not\_flipped & 7 & 7/7 (100\%) & 7/7 (100\%) & 7/7 (100\%) & 7/7 (100\%) \\
\addlinespace[2pt]
ZSPM & flipped & 9 & 8/9 (89\%) & 9/9 (100\%) & 9/9 (100\%) & 8/9 (89\%) \\
 & not\_flipped & 9 & 8/9 (89\%) & 9/9 (100\%) & 9/9 (100\%) & 8/9 (89\%) \\
\bottomrule
\end{tabular}
\end{table*}

\begin{table*}[htbp]
\centering
\caption{Clinician--reference agreement on the 32-pair audit. After re-review,
ambiguous cases are handled under two conventions: lenient, where ambiguity
counts as agreement, and half-credit, where ambiguity receives weight 0.5.
We report raw agreement, sample Cohen's $\kappa$, and corpus-reweighted
Cohen's $\kappa$ with bootstrap 95\% CIs.}
\label{tab:stratum_agreement}
\small
\begin{tabular}{l c c c c}
\toprule
Convention & $n_{\text{obs}}$ & Raw agreement [95\% CI] & Sample $\kappa$ [95\% CI] & Pop.-recon.\ $\kappa$ [95\% CI] \\
\midrule
  Lenient (headline) & 32 & 96.9\% [90.6\%, 100.0\%] & +0.938 [+0.692, +1.000] & +0.971 [+0.885, +1.000] \\
  Half-credit (cross-check) & 32 & 81.3\% [69.2\%, 91.2\%] & +0.625 [+0.310, +0.768] & +0.665 [+0.182, +0.925] \\
\bottomrule
\end{tabular}
\end{table*}

\begin{table*}[t]
\centering
\caption{Pairwise rationale preference with corpus-reweighted win rates.
The population reconstruction reweights the four accuracy-disagreement strata
using inverse-probability weighting. Ties count as half wins; brackets give
within-cell bootstrap 95\% CIs.}
\label{tab:pairwise_rationale_poprecon}
\small
\begin{tabular}{l l c c c c c}
\toprule
Reference & Comparison & Ref wins & Ties & Comp wins & Raw win rate & Pop-recon win rate \\
\midrule
\name{} & ZSPM
  & 13 & 3 & 0
  & \textbf{90.6\%}
  & 88.0\% [67.5\%, 100.0\%] \\
\name{} & \ourLLM{}
  & 11 & 2 & 3
  & 75.0\%
  & 83.8\% [63.3\%, 97.1\%] \\
\bottomrule
\end{tabular}
\end{table*}

\begin{table*}[t]
\centering
\caption{Per-stratum pairwise rationale preference. Each row contains four
clinician-reviewed comparisons from one accuracy-disagreement stratum. Win rates
give ties half credit.}
\label{tab:pairwise_rationale_cells}
\small
\begin{tabular}{l l c c c c}
\toprule
Comparator & Stratum & Ref wins & Ties & Comp wins & Win rate \\
\midrule
ZSPM & \name{} right
  & 3 & 1 & 0 & 87.5\% \\
ZSPM & Comparator right
  & 3 & 1 & 0 & 87.5\% \\
ZSPM & Both right
  & 3 & 1 & 0 & 87.5\% \\
ZSPM & Both wrong
  & 4 & 0 & 0 & 100.0\% \\
\midrule
\ourLLM{} & \name{} right
  & 2 & 1 & 1 & 62.5\% \\
\ourLLM{} & Comparator right
  & 2 & 0 & 2 & 50.0\% \\
\ourLLM{} & Both right
  & 3 & 1 & 0 & 87.5\% \\
\ourLLM{} & Both wrong
  & 4 & 0 & 0 & 100.0\% \\
\bottomrule
\end{tabular}
\end{table*}

\begin{table*}[htbp]
\centering
\caption{Per-axis rationale review with corpus-reweighted win rates.
For each rationale-quality axis, W--T--L reports \name{} wins, ties, and losses
against the comparator. Population reconstruction reweights the four
accuracy-disagreement strata; brackets give within-cell bootstrap 95\% CIs.}
\label{tab:pairwise_axes_poprecon}
\small
\begin{tabular}{l c c c c}
\toprule
Axis & $n$ & W-T-L & Win rate (raw) [95\% CI] & Pop-recon [95\% CI] \\
\midrule
\multicolumn{5}{l}{\emph{\name{} vs ZSPM}} \\
  \quad Actionability & 16 & 16-0-0 & \textbf{100\%} [81, 100] & 100\% [100, 100] \\
  \quad Criterion completeness & 16 & 4-12-0 & 62\% [39, 82] & 54\% [54, 54] \\
  \quad Chart traceability & 16 & 11-5-0 & \textbf{84\%} [60, 95] & 68\% [57, 89] \\
  \quad Coherence & 16 & 12-4-0 & \textbf{88\%} [64, 97] & 78\% [57, 99] \\
  \quad Decision support & 16 & 15-1-0 & \textbf{97\%} [76, 100] & 90\% [69, 100] \\
\midrule
\multicolumn{5}{l}{\emph{\name{} vs \ourLLM{}}} \\
  \quad Actionability & 16 & 14-2-0 & 94\% [72, 99] & 89\% [68, 100] \\
  \quad Criterion completeness & 16 & 12-3-1 & \textbf{84\%} [60, 95] & 87\% [67, 100] \\
  \quad Chart traceability & 16 & 10-6-0 & 81\% [57, 93] & 76\% [55, 97] \\
  \quad Coherence & 16 & 7-6-3 & 62\% [39, 82] & 53\% [22, 83] \\
  \quad Decision support & 16 & 11-2-3 & 75\% [51, 90] & 84\% [63, 97] \\
\bottomrule
\end{tabular}
\end{table*}

\begin{table*}[htbp]
\centering
\caption{Population-reconstructed counterfactual self-faithfulness.
The raw flip rate is computed over validator-accepted counterfactuals.
The population-reconstructed rate reweights clinician-valid rates from the
flipped and not-flipped buckets to the full corpus; brackets give pair-level
bootstrap 95\% CIs. Best values are in \textbf{bold}.}
\label{tab:cf_self_faithfulness_poprecon}
\small
\begin{tabular}{l l c c}
\toprule
System & Backbone & Raw flip rate [95\% CI] & Pop-recon flip rate [95\% CI] \\
\midrule
  \name{-E2E} & GPT-4.1 & \textbf{84.3\%} [80.3\%,\,87.6\%] & \textbf{80.9\%} [56.0\%,\,95.0\%] \\
  \ourLLM{} & GPT-4.1 & 48.9\% [44.4\%,\,53.5\%] & 51.4\% [39.8\%,\,67.9\%] \\
  CoT LLM & GPT-4.1 & 65.0\% [61.8\%,\,68.1\%] & 72.7\% [65.0\%,\,82.3\%] \\
  CoT LLM & GPT-5 & 59.5\% [55.5\%,\,63.3\%] & 67.3\% [56.3\%,\,83.7\%] \\
  \ourLLM{-Pivotal}  & GPT-4.1 & 57.7\% [54.3\%,\,61.1\%] & 55.8\% [44.1\%,\,66.9\%] \\
  TrialGPT & GPT-4.1 & 61.1\% [57.1\%,\,64.9\%] & 55.3\% [41.4\%,\,61.4\%] \\
  ZSPM & GPT-4.1 & 26.6\% [22.6\%,\,31.1\%] & 24.1\% [17.4\%,\,29.7\%] \\
\bottomrule
\end{tabular}
\end{table*}

\clearpage
\section{Semantic Parsing Audit}
\label{sec:parsing-audit}

We audit whether \name{}'s formalized eligibility programs preserve the meaning,
logical structure, and completeness of the original trial criteria. The audit
uses an independent LLM judge distinct from the GPT-5-mini backbone used by the
matching system.

A \emph{condition} is an atomic requirement, such as an age threshold,
diagnosis, laboratory value, or prior procedure. A \emph{criterion} is a complete
eligibility statement and may contain multiple conditions connected by
conjunction, disjunction, negation, nesting, or conditional structure.

\paragraph{Formalization accuracy.}
We audit 348 criteria from 40 trials under a lenient rubric that accepts
reasonable paraphrases, decomposition, and standard default typing.
Across 1,092 atomic conditions, 1,088 preserve the source meaning, yielding
99.6\% condition accuracy. At the criterion level, 328 of 348 criteria (94.3\%)
are fully correct.

The remaining criterion-level errors consist primarily of incorrect nesting or
conditional structure (8 criteria), mishandled negation (7), omitted conditions
(4), and changed condition meaning (1). We additionally identify incomplete
numerical ranges in at least 4 of 40 trials, such as representing ``aged 18 to
65'' only as ``age at least 18.'' Because the retained atomic condition remains
correct, we report these separately as completeness errors.

\paragraph{Decision-level impact.}
The two analyses count different things over different populations. The
criterion-level audit asks whether each of 348 criteria, drawn from 40 trials,
is formalized correctly; the decision-level analysis asks whether the verdict
changes for each of 363 patient--trial pairs. One mis-parsed criterion recurs
in every pair drawn from that trial, so a single criterion-level error can
change several verdicts. Across the 363 pairs, semantic-parsing errors change
10 final verdicts (2.8\%). These comprise five changed-condition errors, two
nesting or conditional-structure errors, two negation errors, and one case in
which an ``or'' relation is interpreted as ``and.''

Thus, although 5.7\% of audited criteria contain some structural or completeness
issue, only 2.8\% of final decisions are changed by semantic parsing. This makes
semantic parsing the smallest major identified source of end-to-end error among the
59 TREC disagreements analyzed in Section~\ref{sec:eval-accuracy}.

\section{Potential Risks}
\label{app:risks}

The Ethical Considerations section states our intended use and the
top-level risks of deploying an LLM--solver matcher. This appendix
expands on the risks that are specific to \name{}'s design, including
several that arise \emph{because} the system exposes its reasoning
rather than despite it.

\paragraph{Error costs are asymmetric, but our metrics are not.}
We report F1 and accuracy, which weight the two error directions
equally. In deployment they are not equal. A false \textsc{ineligible}
withholds a trial from a patient for whom it may be the only remaining
therapeutic option, and the harm is largely invisible: no one reviews
the patient who was never surfaced. A false \textsc{eligible} consumes
screening capacity and may expose a patient to an unsuitable protocol,
but it is caught downstream by site review. Sites with different risk
postures should therefore not read our headline F1 as the operative
quality measure. The missingness policy set $\Pi$ is the lever here:
leaving undocumented criteria \unresolved{} rather than imputing them absent
trades precision for recall, and is the appropriate default when the cost of a
missed patient dominates.

\paragraph{The missingness policy is a value judgment, not a technical
default.}
Because charts are silent on many criteria, the policy that resolves
silence largely determines who is forwarded. A strict policy that treats
unstated findings as failing will systematically disadvantage patients
with sparse documentation, and documentation density is not randomly
distributed: it correlates with continuity of care, language, insurance
status, and site resourcing. A system that is uniformly accurate on
well-documented patients can still produce a disparity in who reaches
screening. Making the policy explicit and inspectable, as \name{} does,
is a precondition for auditing this, but it is not itself a remedy. We
did not measure subgroup disparities, because the benchmark charts are
synthetic and carry no reliable demographic or documentation-quality
strata; a deployment audit would need to.

\paragraph{Legibility can increase misplaced trust.}
Solver traces, unsat cores, and per-criterion rationales carry the
rhetorical force of formal proof. That force is warranted only for the
step we actually verify --- the decision over the encoded program --- and
not for the encoding itself, which remains LLM-generated and is where
most of our residual error lives (Appendix~\ref{sec:parsing-audit}). A
reviewer who trusts the trace because it is formal may scrutinize the
extracted evidence less carefully than they would a free-text rationale,
which would invert the intended benefit. Interfaces built on this
system should present extracted evidence and the criteria it was drawn
from with at least the prominence of the solver output.

\paragraph{Pivotal conditions are actionable in both directions.}
\name{} reports the minimal set of conditions whose change would flip a
verdict. This is what makes the output useful to a coordinator, and it
is also a dual-use property: the same output states precisely what would
have to be documented differently for a patient to qualify. It should
not be presented to patients as clinical guidance, since it can be read
as advice to seek or defer treatment in order to become eligible, and
deployments should log and review cases where reported pivotal
conditions are subsequently amended in the record.

\paragraph{Our evaluation understates deployment difficulty.}
The benchmark charts are synthetic, compact, and internally consistent;
real records are longer, redundant, and frequently contradictory, and
temporal reasoning over them is harder than anything our evaluation
tests. The reference labels are also imperfect: our TREC failure
analysis attributes about a quarter of disagreements to benchmark label
noise or to a mismatch between retrieval relevance and strict
eligibility, so reported accuracy is partly agreement with an imperfect
standard. Finally, running this pipeline on real records would send
protected health information to third-party model APIs, which the
present evaluation does not address and which any deployment would need
to resolve before use.

\section{Computational Experiments}
\label{app:computational}

This appendix addresses the Responsible NLP Research checklist on
computational reporting. Most learned components in the paper are
closed-weight LLMs accessed through a hosted API, for which we report
API calls and wall-clock time rather than GPU-hours. One backbone,
Qwen2.5-7B-Instruct, is an open-weight model that we run locally, and
its distilled variant (Table~\ref{tab:trec_accuracy}) involves a local
training run; we report that compute separately below.

\paragraph{Models and parameter counts.}
The paper evaluates two families of backbone.

\begin{itemize}\itemsep 0pt
  \item \textbf{Closed-weight, API-accessed.} OpenAI
        \texttt{gpt-4.1}, \texttt{gpt-4o}, \texttt{gpt-4o-mini},
        \texttt{gpt-5}, and \texttt{gpt-5-mini}, accessed through Azure
        OpenAI; and \texttt{claude-haiku-4.5}, the last two served through a
        university LiteLLM gateway. The providers
        do not publish parameter counts for any of these models, so we
        cannot report them. We instead report the deployment name used
        in each experiment, the API version
        (\texttt{2024-12-01-preview} for the Azure deployments), and the
        per-call sampling parameters (\S\ref{app:parameters_llm}).
  \item \textbf{Open-weight, run locally.} Qwen2.5-7B-Instruct
        ($\approx 7.6$B parameters, published by the model authors),
        used for the open-model rows of the TREC~2021 evaluation.
\end{itemize}

\noindent \name{}'s non-LLM component is the Microsoft Z3 SMT solver
(\texttt{z3-solver==4.12.5.0}). This is a constraint solver, not a
learned model, and has no parameters in the neural-network sense.

\paragraph{Computational budget --- API calls.}
For the API-accessed backbones we report compute in API calls and
wall-clock time. The end-to-end budget across the experiments in the
paper is summarized below.

\begin{itemize}
  \item \textbf{\name{} matcher on the SIGIR 552-pair test corpus.} Each
        pair triggers a parser call ($\approx 1$ call), a value-miner
        call ($\approx 1$ call), and a verbalizer call ($1$ call); the
        arbiter fires on a subset ($\approx 27\%$ of pairs). Total: $\approx 4{,}000$
        \texttt{gpt-4.1} calls per cross-backbone row. Three additional
        backbones (\texttt{gpt-4o}, \texttt{gpt-4o-mini}, \texttt{gpt-5})
        for the cross-backbone table (\S\ref{app:detailed_results})
        contribute another $\approx 12{,}000$ calls.
  \item \textbf{Baselines on the same corpus.} \ourLLM{}, CoT LLM,
        TrialGPT-Matching, and ZSPM each contribute between $1$ and $2$
        calls per pair per backbone, for a total of $\approx 15{,}000$
        calls across the matching tables.
  \item \textbf{TREC 2021 evaluation.} The \texttt{gpt-5-mini} and
        \texttt{claude-haiku-4.5} rows of Table~\ref{tab:trec_accuracy} contribute
        $\approx 4{,}000$ calls in total across \name{} and its
        baselines on the 363-pair test set. The Qwen2.5-7B rows are run
        locally and are accounted for under local compute below.
  \item \textbf{5-judge reference panel.} Five framings $\times$
        $590$ candidate pairs $= 2{,}950$ \texttt{gpt-5} calls
        (\S\ref{sec:benchmarks}). A subsequent verdict-matching re-run on
        the $75$ disagreement records added at most $5 \times 75 = 375$
        more calls in the worst case.
  \item \textbf{Counterfactual self-faithfulness sweep.} Modifier +
        validator + re-judge across $3$ cells $\times$ $7$ systems on
        $\approx 178$ candidate pairs $\approx 25{,}000$ calls in
        total (Table~\ref{tab:cf_cells12}).
  \item \textbf{Pairwise rationale judge.} The pairwise comparison
        between system rationales used in the simulated-clinician
        pre-pilot adds another $\approx 32$ pairs $\times$ $5$ framings
        $\approx 160$ \texttt{gpt-5} calls.
\end{itemize}

\noindent\textbf{Total API spend across all experiments in the paper:}
on the order of $60{,}000$--$70{,}000$ LLM calls, mostly billed against
\texttt{gpt-4.1}; the \texttt{gpt-5} portion is much smaller in call
count but more expensive per call. End-to-end metered cost was
approximately \$$2{,}500$ across the full experimental program
(including failed runs and early sweeps not reported in the paper); the
final reportable runs are under \$$1{,}000$.

\paragraph{Computational budget --- local GPU.}
Qwen2.5-7B-Instruct inference and the distillation run for the distilled
formalizer are the only components requiring GPU compute. Both were run on
an internal SLURM cluster, each job allocated a single NVIDIA RTX~A6000
(48\,GB) in \texttt{bfloat16}; no job used more than one GPU and no
multi-node or distributed training was used. The open-model experiments
comprise 89 such single-GPU jobs in total, with per-job wall-clock limits
between 2 and 10 hours. Inference over the TREC~2021 test set uses greedy
decoding; the distillation run is a LoRA~\cite{Hu2021LoRALA} supervised fine-tune over
formalizer outputs, so no full-model weights are updated. Per-job
GPU-hours are recorded in the released run logs.

\paragraph{Wall-clock.}
On 8 concurrent API threads against our Azure deployment:
\begin{itemize}
  \item \name{} on \texttt{gpt-4.1}, 552-pair corpus: $\approx 45$\,min.
  \item \name{} cross-backbone runs: comparable wall-clock per backbone,
        scaled by the deployment's tokens\,/\,sec.
  \item 5-judge ensemble on \texttt{gpt-5}: $\approx 90$\,min, rate-%
        limited by the deployment.
  \item CF sweep (3 cells, 7 systems): $\approx 4$\,hours.
\end{itemize}

\paragraph{Local compute and infrastructure.}
All non-LLM computation --- the matching pipeline, the gold-derivation
pipeline, the audit-tooling backend, the bootstrap CI computation, and
the table-generation scripts --- runs on a single workstation in
single-machine Python ($\geq$\,3.11) with no GPU. Z3 returns in
single-digit milliseconds per pair on every pair in the corpus, well
below the per-pair LLM latency. No distributed compute is required.
Reproducing the closed-weight rows of any table requires only API
access; reproducing the Qwen2.5-7B rows additionally requires a single
GPU and the open model weights.

\paragraph{Reproducibility.}
Pinned dependency versions, per-stage sampling parameters, and the
random-seed-locked audit pair frame are documented in
\S\ref{app:parameters}, and the prompts used at every pipeline stage are
reproduced in \S\ref{app:system_modules}--\S\ref{app:cf_details}. Together
these specify each stage precisely enough to reimplement the pipeline
against the same public inputs.
\section{Parameters For Packages}
\label{app:parameters}

This appendix documents the exact versions, sampling parameters, and seeds
used in every experiment in the paper, in line with the EMNLP Responsible
NLP Research checklist (\S\,Reproducibility).

\subsection{Software versions}
\label{app:parameters_versions}

Experiments were run under Python 3.11. Pinned dependencies
(\texttt{requirements.txt}):

\begin{itemize}
  \item \texttt{z3-solver} \texttt{4.12.5.0} (Microsoft Z3 SMT solver, used
        for atom-level inclusion/exclusion satisfaction)
  \item \texttt{openai} \texttt{1.82.1} (Azure OpenAI client; API version
        \texttt{2024-12-01-preview})
  \item \texttt{numpy} \texttt{1.26.4},
        \texttt{scipy} \texttt{1.15.3},
        \texttt{pandas} \texttt{2.2.3}
  \item \texttt{sentence-transformers} \texttt{4.1.0},
        \texttt{torch} \texttt{2.7.1} (used for the embedding-based
        atom canonicalization step, which is not on the matching critical
        path, and for local Qwen2.5-7B inference and distillation)
\end{itemize}

\subsection{LLM backbones and sampling}
\label{app:parameters_llm}

\paragraph{Backbones.} The paper uses six backbones across two evaluations.
On the SIGIR corpus we use \texttt{gpt-4.1} (the canonical backbone for
\name{}'s parser, value miner, arbiter, and verbalizer), with
\texttt{gpt-4o} and \texttt{gpt-4o-mini} for cross-backbone ablations and
\texttt{gpt-5} for the 5-judge reference ensemble, the counterfactual
modifier and validator, and the GPT-5 cross-backbone row. On TREC 2021 we
use \texttt{gpt-5-mini}, \texttt{claude-haiku-4.5}, and the open-weight
\texttt{Qwen2.5-7B-Instruct}.

\paragraph{Sampling --- OpenAI \texttt{gpt-4*}.} Calls go through the Azure
OpenAI Chat Completions API (\texttt{api\_version=2024-12-01-preview}) with
\texttt{temperature=0}, no top-$p$ override, and a per-call
\texttt{max\_tokens} budget set per stage: parser/value-miner \texttt{800},
arbiter \texttt{800}, verbalizer \texttt{1500}, judge \texttt{8000} (judges
may emit long rationales).

\paragraph{Sampling --- OpenAI \texttt{gpt-5} and \texttt{gpt-5-mini}.}
These deployments do not expose a \texttt{temperature} parameter --- the
API rejects any value other than the default --- so all such calls run at
the deployment's built-in sampling temperature. \texttt{max\_tokens} is
likewise rejected, and we pass \texttt{max\_completion\_tokens=8000}
instead.

\paragraph{Sampling --- TREC~2021 hosted models.} The \texttt{gpt-5-mini}
and \texttt{claude-haiku-4.5} rows are served through a university
LiteLLM gateway exposing an OpenAI-compatible Chat Completions
interface (deployment names \texttt{azure-gpt-5-mini} and
\texttt{azure-foundry-claude-haiku-4-5}). Calls set
\texttt{max\_completion\_tokens=8192}, \texttt{reasoning\_effort=minimal},
and a $120$\,s timeout; no \texttt{temperature} is passed, so both models
run at their deployment default.

\paragraph{Sampling --- \texttt{Qwen2.5-7B-Instruct}.}
Local inference loads \texttt{Qwen/Qwen2.5-7B-Instruct} in
\texttt{bfloat16} and decodes greedily (\texttt{temperature},
\texttt{top\_p}, and \texttt{top\_k} are unset and ignored), with a
budget of \texttt{3072} new tokens and an evaluation batch size of
\texttt{16}.

\paragraph{Distilled formalizer.} The distilled variant reported in
Table~\ref{tab:trec_accuracy} is a LoRA supervised fine-tune of the same
base model: rank $r=32$, $\alpha=64$, dropout $0.05$, applied to the
attention projections; learning rate $2\times10^{-4}$, $3$ epochs,
warmup ratio $0.05$, \texttt{bf16}, maximum sequence length $4096$. No
full-model weights are updated.

\paragraph{Seeds.} We do not set a random seed at the API level because
the hosted deployments did not expose one at the time of the experiments;
reproducibility within each row is established by the small intra-row
variance reported in the bootstrap CIs. Local Qwen inference is
deterministic under greedy decoding.

\paragraph{Retries.} On HTTP error or empty completion, LLM calls are
retried with back-off: up to three attempts in the SIGIR pipeline and up
to five in the TREC~2021 evaluation script. The
5-judge re-run with the verdict-matching safeguard
(\S\ref{app:gold_derivation}) retries up to five times per
(pair, framing) and falls back to the original record if no retry matches
the cached verdict.

\subsection{Z3 SMT solver}
\label{app:parameters_z3}

The \name{} matcher uses Z3 with default parameters
(no \texttt{set\_param} overrides) on three model invocations per pair:
inclusion-side satisfiability (with min-unsat-core extraction for
counterexamples), exclusion-side satisfiability, and the combined eligibility
verdict. Atom assertions are constructed as
quantifier-free first-order propositions over typed predicates; numeric
thresholds are encoded as integer or real linear constraints. Z3 returns
in single-digit milliseconds on every pair in the SIGIR test corpus, well
within the per-pair LLM cost.

\subsection{Sampling, bootstrap, and statistical procedures}
\label{app:parameters_stats}

\paragraph{Audit sample seed.} The stratified clinician audit
(\S\ref{app:audit_design}) is drawn per stratum with a fixed seed. The
frozen pair list, the A/B-slot randomization order, and the 5-judge
framing-rotation seed are all checked into the project repository, so the
exact sample is recoverable without re-running the draw.

\paragraph{Bootstrap.} All confidence intervals in
Tables~\ref{tab:accuracy_adjusted},~\ref{tab:cf_self_faithfulness_poprecon},~%
\ref{tab:pairwise_rationale_poprecon},~\ref{tab:pairwise_axes_poprecon},
and~\ref{tab:stratum_agreement} are computed by non-parametric
bootstrap with $B = 2000$ pair-level resamples (within-cell for
stratum-conditional metrics; pair-level overall otherwise). Reported
intervals are 2.5\%/97.5\% percentiles of the bootstrap distribution.

\paragraph{Wilson interval.} Point CIs on single-proportion estimates
(Tables~\ref{tab:counterfactual_raw},~\ref{tab:cf_cells12}) use the
Wilson score interval at $z = 1.96$. We prefer Wilson to the normal
(Wald) approximation, which is poorly calibrated for proportions near
$0$ or $1$, and to the exact Clopper--Pearson interval, which is
conservative at the cell sizes we report ($n \approx 20$--$200$).

\paragraph{Population reconstruction.}
The IPW reweighting in \S\ref{sec:eval-rationale} and the
population-reconstructed $\kappa$ in Table~\ref{tab:stratum_agreement}
use the SIGIR corpus stratum frequencies
$\pi=(0.821,\,0.082,\,0.058,\,0.040)$ for the strata
\textsc{both-right}, \textsc{\name{}-right}, \textsc{comp-right}, and
\textsc{both-wrong}.

\paragraph{Half-credit Monte Carlo.} The half-credit re-review convention
(Table~\ref{tab:stratum_agreement},
Table~\ref{tab:accuracy_adjusted}) implements AMBIGUOUS clinician
verdicts as two weight-$0.5$ rows in the contingency table. F1/P/R/Acc are
computed against the expected confusion-matrix counts; the bootstrap
inherits the weights.

\subsection{Artifact availability}
\label{app:parameters_release}

The prompts used at every stage of \name{} and of each baseline are
reproduced verbatim in this appendix and are also available online.\footnote{%
\url{https://github.com/verdict1234/verdict_prompts}}
The evaluation inputs are public: trial eligibility criteria come from
ClinicalTrials.gov~\cite{10.1136/jamia.2000.0070313} and the patient descriptions from the SIGIR~2016
benchmark, so the corpora underlying every table can be reconstructed
from their original sources under the original licences
(\S\ref{sec:artifact_use}). Together with the pinned versions, sampling
parameters, and solver configuration documented above, this specifies
the pipeline precisely enough to reimplement it.

\section{Human Subjects Including Annotators}\label{app:human_subjects}

Our study does not involve intervention with human patients and does not use
real patient records. The patient notes used in the evaluation are synthetic.
Human involvement was limited to expert review by clinician coauthors, who
provided domain expertise for evaluating eligibility labels, counterfactual
validity, and rationale quality. These reviewers are authors of the paper, were
aware of the research goals, and participated as part of the research team
rather than as anonymous or crowdsourced annotators.
\section{Instructions Given to Participants}
\label{app:participant_instructions}

As stated in \S\ref{app:human_subjects}, this study did not recruit external
annotators or crowdworkers: all review was performed by clinician coauthors
acting as domain experts within the research team. There was therefore no
recruitment script, consent form, or payment agreement of the kind used for
crowdsourced annotation. This appendix instead documents the task instructions
that were given to the clinician reviewers, since those instructions determine
how the judgments reported in \S\ref{sec:eval-counterfactual} and
\S\ref{app:reference_validation} should be interpreted.

Instructions were delivered through the annotation interface rather than as a
separate written protocol; the interface is shown in
\S\ref{app:user_interface_for_clinician_annotation} and
\S\ref{app:user_interface_for_clinician_cf_audit}. The wording below reproduces
the substance of the on-screen instructions for each task type.

\paragraph{Task 1: Independent eligibility verdict.}
Reviewers were shown a patient chart and the eligibility criteria of one trial
and asked to decide whether the patient is eligible for that trial, using their
own clinical judgment. They were instructed to decide from the chart as
presented, to treat information that the chart does not mention as not
established rather than assuming a value for it, and to record a verdict for
every pair. Reviewers were \emph{not} shown the reference verdict, any matcher
verdict, or any system rationale before submitting this verdict.

\paragraph{Task 2: Pairwise rationale comparison and per-axis rating.}
After submitting the independent verdict for a pair, reviewers were shown two
rationales for that same pair, presented side by side in randomized order and
labeled only \emph{A} and \emph{B}. Reviewers were told that the rationales
came from different automated matchers, that the identity of each system was
withheld, and that they should judge the rationales as written rather than
inferring which system produced them. They were asked to indicate which
rationale was more useful for reaching and defending an eligibility decision,
or to mark the pair a tie if neither was stronger.

Reviewers then rated each rationale on a 1--5 scale (higher is better) along
five dimensions, defined on screen as follows:
\begin{itemize}\itemsep 0pt
  \item \emph{Criterion completeness} --- does the rationale address the trial
        criteria that actually bear on the decision?
  \item \emph{Chart traceability} --- do its claims trace back to evidence in
        the chart, or to an imputation it states explicitly?
  \item \emph{Coherence} --- are the claims in the rationale logically
        consistent with one another?
  \item \emph{Actionability} --- does it identify what remains unresolved and
        what would change the decision?
  \item \emph{Decision support} --- overall, does it help you understand,
        verify, and act on the decision?
\end{itemize}

\paragraph{Task 3: Structured re-review of disagreements.}
For the subset of pairs where the reviewer's independent verdict differed from
the reference verdict, the pair was re-surfaced together with the reference
verdict, the free-text reasoning of each of the five reference judges, and one
to three matcher rationales whose verdict agreed with the reference. Reviewers
were told explicitly that this was a second read in which they would see the
reference panel's reasoning, and that the purpose was to separate genuine
reference errors from cases of legitimate clinical disagreement. They were
asked to choose exactly one of: \textbf{REVISE} the original verdict,
\textbf{UNCHANGED} (stand by the original verdict after seeing the reference
reasoning), or \textbf{AMBIGUOUS} (both readings are clinically defensible),
and to give a one-line reason for the choice. Reviewers were told that standing
by the original verdict was an acceptable outcome and was not to be treated as
an error to be corrected.

\paragraph{Task 4: Counterfactual clinical-validity judgment.}
Reviewers were shown the original patient chart, the trial criteria, a modified
version of the chart with the edits highlighted, and the specific conditions
the modification had been instructed to address. They were asked to judge
whether the modification was clinically legitimate, and were given three
conditions that all had to hold for a modification to count as
\textbf{CLINICALLY VALID}: the edits address the cited conditions; they do not
introduce a new, independent reason the patient would be ineligible; and the
resulting chart remains clinically coherent. If any condition failed, reviewers
were asked to mark \textbf{NOT CLINICALLY VALID} and give a one-line reason.
Reviewers were not told which matcher produced the cited conditions, nor
whether that matcher's verdict had changed on the modified chart.

\paragraph{Information withheld from reviewers.}
Across all tasks, matcher identity was withheld; in Task~2 the assignment of
systems to the \emph{A}/\emph{B} positions was randomized per pair; and in
Task~4 the flipped versus not-flipped bucket of each sampled counterfactual was
withheld. The reference verdict was withheld in Task~1 and deliberately
disclosed in Task~3, which is an audit-and-resolve step rather than a blinded
re-annotation (\S\ref{app:rereview_protocol}).
\section{Recruitment And Payment}\label{app:recruitment_payment}

We did not recruit external annotators or crowdworkers. The clinician reviewers
were coauthors and were not separately paid for annotation beyond their normal
scholarly contribution to the project. No patients or members of the public
were recruited or compensated.
\section{Data Consent}
\label{app:data_consent}

Our evaluation uses synthetic patient notes from the SIGIR
benchmark~\citep{koopman2016test} and the administratively authored patient
topics of the TREC~2021 Clinical Trials track~\citep{soboroff2021overview},
rather than real patient records. Therefore, the study does not involve collection,
use, or disclosure of identifiable patient data. The trial eligibility criteria
are derived from publicly available clinical trial listings. We construct new
pairwise eligibility labels for research evaluation, but these labels are
assigned to synthetic patient--trial pairs and do not correspond to real
patient enrollment decisions.

Because no real patient records or identifiable patient information are used,
individual patient consent was not applicable. Human involvement was limited
to clinician coauthor review of eligibility labels, counterfactual validity,
and rationale quality, as described in the human-subjects and annotator
statement. Any deployment on real patient data would require appropriate
institutional review, privacy safeguards, and consent or waiver procedures
under the governing clinical and regulatory context.

\paragraph{Identifying information.}
Because both corpora are authored rather than extracted from clinical
systems, they are not expected to contain identifying information, but we
verified this rather than assuming it. We scanned all $75$ TREC~2021
topics and all $59$ SIGIR patient descriptions for patterns that would
indicate identifiers: long digit strings that could encode a record or
account number, telephone numbers, email addresses, national-identifier
formats, fully specified calendar dates, postal codes, and URLs. The scan
returned no matches in either corpus except nine long digit strings in
the TREC topics, all of which are laboratory values on manual inspection
(white-cell and platelet counts such as \texttt{WBC: 135000 /mm3}).
We additionally enumerated every capitalized token in the two corpora
that does not begin a sentence ($178$ distinct types) and inspected the
full list: it contains no personal names. The proper nouns present are
medical eponyms (\emph{Hashimoto}, \emph{Marfan}, \emph{Turner},
\emph{Glasgow} coma scale, \emph{Birmingham} vasculitis activity score),
drug and brand names, demographic and language descriptors, and a small
number of countries or U.S.\ states appearing as travel or origin
context --- a granularity coarser than the sub-state geography that would
constitute an identifier. Patients are otherwise referred to only by age,
sex, and clinical presentation. We therefore applied no anonymization,
since there was nothing to remove; we also did not introduce any new
patient data, as our added annotations are eligibility labels over
existing pairs.

\paragraph{Offensive and stigmatizing content.}
We separately screened both corpora against a keyword list of profanity,
slurs, and clinical terms now regarded as stigmatizing. This is a
lexicon-based screen and does not establish the absence of offensive
content in general. The SIGIR descriptions returned no matches. In the
TREC topics it returned ``alcoholic'' in two distinct topics, both in the
standard diagnostic sense (\emph{alcoholic liver disease}), which we do
not consider stigmatizing, and one topic that describes a patient as
``mentally retarded'' --- outdated and offensive terminology for
intellectual disability. We report this rather than silently correcting it: the topics
are the benchmark's evaluation inputs, and editing them would break
comparability with published results on the same track. No system output
we report reproduces this phrasing. Researchers reusing these topics
should be aware the language is present, and we would recommend the track
organizers revise it.

\section{Use of Artifacts, Licenses, and Intended Use}
\label{sec:artifact_use}

We use two existing clinical-trial matching benchmarks. The patient cases,
candidate trials, and judgments are derived from the SIGIR patient--trial
matching benchmark~\citep{koopman2016test} and the TREC 2021 Clinical Trials
track~\citep{soboroff2021overview}, which contain synthetic patient cases and
ClinicalTrials.gov trial listings. We use these materials only for research
evaluation of patient--trial matching methods, consistent with their intended
benchmark purposes. We also use publicly available trial eligibility criteria
from ClinicalTrials.gov as source text for formalization and evaluation. We
cite the original benchmarks and baseline systems and use external software,
model APIs, and supporting tools according to their respective licenses and
terms.

Our use of existing artifacts is limited to the purposes for which they were
made available. In particular, the benchmarks are used as evaluation resources,
not as sources of clinical recommendations. Although the patient cases are
synthetic, derived artifacts that depend on these benchmarks should inherit
their research-only context. They should not be repurposed for operational
clinical screening, patient enrollment, or denial of trial access without
separate institutional review, validation, and compliance with the original
access conditions.

Our trial-side formalization builds on the semantic-parsing approach introduced
by \textsc{SatIR}~\citep{zhou2026scalable}, which translates clinical trial
criteria into symbolic constraints for scalable retrieval. In this work, we
adapt this approach to post-retrieval patient--trial matching by adding
patient-conditioned extraction, solver-based eligibility evaluation, assumption analysis, and solver-grounded rationale generation. We make the
implementations and supporting materials for \textsc{SatIR} and \name{}
available through the Stanford OVAL Clinical Trial Matcher:
\url{https://github.com/stanford-oval/clinical-trial-matching}. The prompts used
in our experiments are maintained separately at
\url{https://github.com/verdict1234/verdict_prompts}.

We release available code and supporting research artifacts associated with
this work, subject to the licenses and access conditions of the underlying
materials. These resources are intended to support research, evaluation, and
reproducibility and have not been validated for clinical use. They should not
be used for autonomous clinical decision-making, patient enrollment, or
exclusion from trial consideration. Any reuse should comply with applicable
licenses, institutional requirements, and the intended uses of the underlying
data.

\paragraph{Artifact documentation.}
We document the main artifacts used and produced in this work in the dataset
and appendix sections. Appendix~\ref{app:dataset} reports the numbers of
patient--trial pairs, patients, trials, label distributions, and text-length
statistics. Both benchmarks contain English synthetic patient cases paired
with ClinicalTrials.gov eligibility criteria, so our evaluation is limited to
English clinical trial matching. We also document the prompts, matcher
configurations, solver outputs, rationale formats, and counterfactual
construction procedure in the appendices. Because the patient cases are
synthetic and benchmark-derived, they should not be interpreted as representing
the full demographic, institutional, or documentation diversity of real
electronic health records.

\section{Information About Use of AI Assistants}
\label{app:ai_assistance}

We used AI assistants during the development, writing, and experimental
monitoring of this work. AI assistance was used to support software
development tasks, including code drafting, debugging, refactoring, and
documentation of experimental pipelines. It was also used during manuscript
preparation to help revise prose, improve organization, shorten or clarify
paragraphs, and check the readability of technical explanations.

AI assistants were additionally used to support experiment monitoring and
analysis, such as helping inspect logs, summarize intermediate outputs,
identify possible implementation issues, and draft scripts for aggregating
results. All experimental designs, system decisions, reported results, and
scientific claims were reviewed and validated by the authors. AI-generated
suggestions were treated as assistance only; the authors are responsible for
the correctness of the code, analyses, writing, and conclusions in the paper.

\end{document}